\pdfoutput=1

\documentclass[sigconf, nonacm]{acmart}

\AtBeginDocument{%
	\providecommand\BibTeX{{%
			\normalfont B\kern-0.5em{\scshape i\kern-0.25em b}\kern-0.8em\TeX}}}

\setcopyright{acmcopyright}
\copyrightyear{2021}
\acmYear{2021}
\acmDOI{10.1145/1122445.1122456}

\usepackage{graphicx} 
\usepackage{caption} 
\usepackage{subcaption}
\usepackage{comment}
\usepackage{amsmath}
\usepackage{amsfonts}
\usepackage{algorithm}
\usepackage{bm}
\usepackage{multirow}
\usepackage{algpseudocode}
\usepackage{siunitx}

\acmConference[ACM MM '21]{ACM Multimedia 2021}{Oct 20--24, 2021}{Chengdu, China}
\acmBooktitle{ACM MM '21: ACM Multimedia 2021,
	Oct 20--24, 2021, Chengdu, China}
\acmPrice{15.00}
\acmISBN{978-1-4503-XXXX-X/18/06}

\renewcommand{\algorithmicrequire}{\textbf{Input:}}  

\newcommand{\nn}{\num[group-separator={,},group-minimum-digits=3]}
\newcommand{\B}{\bfseries}

\acmSubmissionID{1718}

\author{Chen Ma}
\email{mac16@mails.tsinghua.edu.cn}
\orcid{0000-0001-6876-3117}
\affiliation{%
	\institution{School of Software, BNRist, Tsinghua University}
	\city{Beijing}
	\country{China}
	\postcode{100084}
}

\author{Shuyu Cheng}
\email{chengsy18@mails.tsinghua.edu.cn}
\affiliation{%
	\institution{Dept. of Comp. Sci. and Tech., BNRist Center, Institute for AI, THBI Lab, Tsinghua University}
	\city{Beijing}
	\country{China}
	\postcode{100101}
}

\author{Li Chen}
\email{chenlee@tsinghua.edu.cn}
\authornote{Corresponding author.}
\affiliation{%
	\institution{School of Software, BNRist, Tsinghua University}
	\city{Beijing}
	\country{China}
	\postcode{100084}
}

\author{Jun Zhu}
\email{dcszj@mail.tsinghua.edu.cn}
\affiliation{%
	\institution{Dept. of Comp. Sci. and Tech., BNRist Center, Institute for AI, THBI Lab, Tsinghua University}
	\city{Beijing}
	\country{China}
	\postcode{100101}
}

\author{Junhai Yong}
\email{yongjh@tsinghua.edu.cn}
\affiliation{%
	\institution{School of Software, BNRist, Tsinghua University}
	\city{Beijing}
	\country{China}
	\postcode{100084}
}

\begin{document}
	
	\title{Switching Transferable Gradient Directions for Query-Efficient Black-Box Adversarial Attacks}
	


	\begin{abstract}
		We propose a simple and highly query-efficient black-box adversarial attack named SWITCH, which has a state-of-the-art performance in the score-based setting. SWITCH features a highly efficient and effective utilization of the gradient of a surrogate model $\hat{\mathbf{g}}$ w.r.t. the input image, \textit{i.e.,} the transferable gradient. In each iteration, SWITCH first tries to update the current sample along the direction of $\hat{\mathbf{g}}$, but considers switching to its opposite direction $-\hat{\mathbf{g}}$ if our algorithm detects that it does not increase the value of the attack objective function. We justify the choice of switching to the opposite direction by a local approximate linearity assumption. In SWITCH, only one or two queries are needed per iteration, but it is still effective due to the rich information provided by the transferable gradient, thereby resulting in unprecedented query efficiency. To improve the robustness of SWITCH, we further propose SWITCH$_\text{RGF}$ in which the update follows the direction of a random gradient-free (RGF) estimate when neither $\hat{\mathbf{g}}$ nor its opposite direction can increase the objective, while maintaining the advantage of SWITCH in terms of query efficiency. Experimental results conducted on CIFAR-10, CIFAR-100 and TinyImageNet show that compared with other methods, SWITCH achieves a satisfactory attack success rate using much fewer queries, and SWITCH$_\text{RGF}$ achieves the state-of-the-art attack success rate with fewer queries overall. Our approach can serve as a strong baseline for future black-box attacks because of its simplicity. The PyTorch source code is released on \url{https://github.com/machanic/SWITCH}.
	\end{abstract}



	
	\settopmatter{printfolios=true}
	\maketitle
	\section{Introduction}
	Adversarial attacks present a major security threat to deep neural networks (DNNs) by adding human-imperceptible perturbations to benign images for the misclassification of DNNs. Adversarial attacks can be divided into two main categories based on whether the internal information of the target model is exposed to the adversary, \textit{i.e.,} white-box and black-box attacks. Compared to white-box, black-box attacks are more useful because they do not require the parameters or gradients of the target model. The black-box attacks can be divided into transfer- and query-based attacks. 
	
	Transfer-based attacks generate adversarial examples by attacking a pre-trained surrogate model to fool the target model \cite{liu2017delving,Ambra2019Why,huang2019enhancing}. Such attacks have practical value because they do not require querying the target model. However, the attack success rate is typically low in the following situations: (1) targeted attack, and (2) the network structures of the surrogate model and target model have a large difference. The low success rate is mainly because the surrogate model's gradient deviates too much from the true gradient. The former could frequently point to a non-adversarial region of target model, so following this direction consecutively leads to a failed attack. Hence, how to switch the gradient when it deviates from the correct direction is worthy of further exploration.
	
	Query-based attacks require an oracle access to the target model, and we focus on the score-based attack setting which requires accessing to the loss function value. Some score-based attacks~\cite{chen2017zoo,tu2019autozoom,ilyas2018blackbox,ilyas2018prior} estimate an approximate gradient by querying the target model at multiple points. However, the high query complexity is inevitable in estimating the approximate gradient with high precision. To reduce the query complexity, another type of query-based attacks, \textit{i.e.,} random-search-based attacks \cite{guo2019simple,ACFH2020square}, eliminates the gradient estimation and turns to sampling a random perturbation at each iteration, which is either added to or subtracted from the target image. Then, the perturbed image is fed to the target model to compute a loss value. If the loss value peaks a historical high, the adversary accepts the perturbation; otherwise, it is rejected. The main issue of this strategy is that the sampled perturbation does not incorporate the gradient information, thereby leading to a high rejection rate and inefficient optimization. In addition, nothing is done if the perturbation is rejected, resulting in many wasted queries. Thus, how to incorporate the gradient information to enhance the optimization's efficiency should be investigated.
	
	In this study, we propose a novel black-box adversarial attack named SWITCH that exploits and revises the surrogate model's gradient (called surrogate gradient) to improve the query efficiency. The surrogate gradient is a promising exploration direction for \textit{maximizing} the objective loss function\footnote{In this study, we maximize the attack objective loss function to update the adversarial example.}, but it does not always point the target model's adversarial region. To detect such situation and reduce incorrect updates, SWITCH proposes to query a loss value $\mathcal{L}(\mathbf{x}_{\mathrm{adv}}+\eta\cdot \overline{\hat{\bm{\mathrm{g}}}})$ based on a temporarily generated image $\mathbf{x}_{\mathrm{adv}}+\eta\cdot \overline{\hat{\bm{\mathrm{g}}}}$ updated along the $\ell_p$ normalized surrogate gradient $\overline{\hat{\bm{\mathrm{g}}}}$, where $\eta$ is the step size. If this loss is not increased from the last iteration ($\mathcal{L}(\mathbf{x}_{\mathrm{adv}}+\eta\cdot \overline{\hat{\bm{\mathrm{g}}}}) < \mathcal{L}(\mathbf{x}_{\mathrm{adv}})$), SWITCH considers switching to a new direction that is obtained by taking a negative sign for the surrogate gradient (namely $-\overline{\hat{\bm{\mathrm{g}}}}$). The effectiveness of this opposite direction is based on the (local) approximate linearity assumption of DNNs \cite{goodfellow6572explaining}. However, the approximate linearity assumption may not always hold and then the $-\overline{\hat{\bm{\mathrm{g}}}}$ may point to a worse direction. To prevent such situation, our algorithm queries the loss $\mathcal{L}(\mathbf{x}_{\mathrm{adv}}-\eta\cdot \overline{\hat{\bm{\mathrm{g}}}})$ and then selects a relatively better direction from $\{-\overline{\hat{\bm{\mathrm{g}}}},  \overline{\hat{\bm{\mathrm{g}}}}\}$. Specifically, we compare $\mathcal{L}(\mathbf{x}_{\mathrm{adv}}-\eta\cdot \overline{\hat{\bm{\mathrm{g}}}})$ with $\mathcal{L}(\mathbf{x}_{\mathrm{adv}}+\eta\cdot \overline{\hat{\bm{\mathrm{g}}}})$, and the gradient direction ($-\overline{\hat{\bm{\mathrm{g}}}}$ or $\overline{\hat{\bm{\mathrm{g}}}}$) that corresponds the the larger loss value will be taken as the final direction for the update. Note that this comparison cannot exclude the situation in which both $\mathcal{L}(\mathbf{x}_{\mathrm{adv}}+\eta\cdot \overline{\hat{\bm{\mathrm{g}}}}) < \mathcal{L}(\mathbf{x}_{\mathrm{adv}})$ and $\mathcal{L}(\mathbf{x}_{\mathrm{adv}}-\eta\cdot \overline{\hat{\bm{\mathrm{g}}}}) < \mathcal{L}(\mathbf{x}_{\mathrm{adv}})$ happen, and thus SWITCH cannot always increase the loss during the attack. To deal with this problem, we further propose an extended version of SWITCH, \textit{i.e.,} SWITCH$_\text{RGF}$.
	
	SWITCH$_\text{RGF}$ makes up for the shortcoming of SWITCH. It adopts a more effective direction ($\hat{\bm{\mathrm{g}}}_\text{RGF}$) by using random gradient-free (RGF). $\hat{\bm{\mathrm{g}}}_\text{RGF}$ is an approximate gradient direction that is obtained by using a number of queries, but it can increase the loss value with a great probability if both $\overline{\hat{\bm{\mathrm{g}}}}$ and $-\overline{\hat{\bm{\mathrm{g}}}}$ fail. Therefore, SWITCH$_\text{RGF}$ improves the success rate over SWITCH.
	
	Finally, the adversarial image is updated by using the selected direction. Thus, in each iteration, the algorithm aims to keep the loss function increasing by switching the gradient. Both the attack success rate and query efficiency are improved because the switched direction could avoid following the wrong direction and consequently bypass the potential obstacle in optimization. 
	
	We evaluate the proposed method on the CIFAR-10 \cite{krizhevsky2009learning}, CIFAR-100 \cite{krizhevsky2009learning} and TinyImageNet \cite{russakovsky2015imagenet} datasets and compare it with 7 state-of-the-art query-based black-box attacks. The results show that the proposed approach requires the fewest queries to achieve a high attack success rate, which significantly outperforms other methods. Therefore, SWITCH achieves new state-of-the-art performance. 
	
	The main contributions of this study are as follows.
	
	(1) We propose a simple and highly query-efficient black-box attack named SWITCH that exploits and revises the surrogate gradient $\overline{\hat{\bm{\mathrm{g}}}}$ during optimization. SWITCH tries to select a better direction in each iteration to avoid following the wrong direction, which keeps the objective loss function rising as much as possible.
	
	(2) We also propose an extended version of our algorithm \textit{i.e.,} SWITCH$_\text{RGF}$, which further improves the success rate over SWITCH.
	
	(3) Despite its simplicity, SWITCH and SWITCH$_\text{RGF}$ spend fewer queries than 7 state-of-the-art methods and significantly improves the attack success rate over the baseline on the CIFAR-10, CIFAR-100 and TinyImageNet datasets. We consider the proposed method to be a strong baseline for future attacks.

	\section{Background}

	\noindent\textbf{Black-box attacks.} It is unrealistic to obtain internal information of target models in many real-world systems, thus many studies focus on performing attacks in the black-box setting. Black-box attacks can be divided into transfer- and query-based attacks. In transfer-based attacks \cite{liu2017delving,papernot2016transferability,Ambra2019Why,huang2019enhancing}, adversarial examples generated by attacking a surrogate model might remain adversarial for the target model. Given that targeted attack is difficult for transfer-based attacks, certain methods use the ensemble of surrogate models \cite{liu2017delving} to improve the success rate. However, the attack success rate remains unsatisfactory and new types of attacks can be detected easily by using a meta-learning-based detector \cite{MetaAdvDet}.

	\begin{figure*}[t]
		\begin{center}
			\includegraphics[width=1\linewidth]{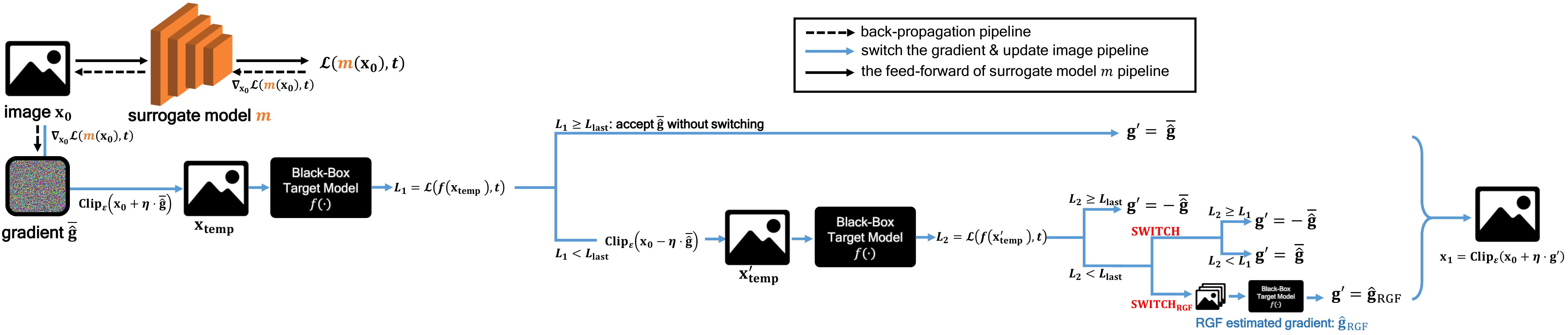}
		\end{center}
		\caption{Illustration of an attack iteration (detailed in Algorithm \ref{alg:attack}), where $L_{\text{last}}$ indicates the loss value of the last iteration, and $t$ indicates the target class in targeted attack and true class in untargeted attack. In each iteration, the algorithm makes a temporary image $\mathbf{x}_{\text{temp}}$ with the normalized surrogate gradient $\overline{\hat{\mathbf{g}}}$, namely $\mathbf{x}_{\text{temp}} = \text{Clip}_\epsilon (\mathbf{x}_\mathrm{adv} + \eta \cdot \overline{\hat{\bm{\mathrm{g}}}})$. Then, the loss $L_1$ is computed by feeding $\mathbf{x}_{\text{temp}}$ to the target model. If $L_1 \ge L_{\text{last}}$, the algorithm accepts $\overline{\hat{\mathbf{g}}}$. Otherwise, the gradient $\overline{\hat{\mathbf{g}}}$ deviates too much from the true gradient such that $L_1 < L_{\text{last}}$. Then we switch to the direction of $-\overline{\hat{\mathbf{g}}}$ and query a loss value $L_2$ based on another temporary image $\text{Clip}_\epsilon (\mathbf{x}_\mathrm{adv} - \eta \cdot \overline{\hat{\bm{\mathrm{g}}}})$. After the feedback is obtained, we consider two options, namely, SWITCH and its extended version SWITCH$_\text{RGF}$. SWITCH selects a direction from $\overline{\hat{\mathbf{g}}}$ and $-\overline{\hat{\mathbf{g}}}$ that corresponds to the larger loss value between $L_1$ and $L_2$ as the final direction $\mathbf{g}^\prime$. SWITCH$_\text{RGF}$ turns to using RGF to estimate an approximate gradient as $\mathbf{g}^\prime$ if both $L_1$ and $L_2$ are less than $L_{\text{last}}$. Finally, the sample $x_1$ is generated by using $\mathbf{g}^\prime$.}
		\label{fig:attack}
	\end{figure*}
	
	Query-based attacks can be further divided into decision- and score-based attacks based on the type of information returned from the target model to the adversary.
	In decision-based attacks \cite{dong2019efficient,cheng2019sign}, the adversary only knows the output label of the target model. In this study, we focus on the score-based attack, in which the adversary can obtain the output scores of the target model. Although the true gradient cannot be directly obtained in black-box attacks, we can still optimize the adversarial examples by using the approximate gradients. Early studies \cite{chen2017zoo,bhagoji2018practical} estimate the approximate gradient by sampling from a noise distribution around the pixels, which is expensive because each pixel needs two queries. To reduce the query complexity, methods are improved by incorporating data and time prior information \cite{ilyas2018prior}, using meta-learning to learn a simulator \cite{ma2021simulator}, searching the solution among the vertices of the $\ell_\infty$ ball \cite{moonICML19}, and searching the adversarial perturbation on a low-dimensional embedding space \cite{tu2019autozoom}. Different from the methods that estimate gradients, random-search-based attacks \cite{guo2019simple,ACFH2020square} sample a random perturbation with the values filled with the maximum allowed perturbation $\epsilon$ or $-\epsilon$, which is either added to or subtracted from the target image. Then, the modified image is fed to the target model to determine whether to accept the perturbation. This type of methods significantly reduces the query complexity because they directly find the solution among the extreme points in $\ell_p$ ball and do not require the high-cost gradient estimation. However, the sampled perturbation does not encode the gradient information, resulting in inefficient optimization. In contrast, the proposed method exploits the gradient of the surrogate model and then switches the direction when necessary to keep the loss increasing as much as possible, thereby achieving high success rate and query efficiency.
	
	\noindent\textbf{Adversary Setup.} Given the target model $f$ and input-label pair $(\mathbf{x},y)$, which is correctly classified by $f$. The adversarial example $\mathbf{x}_\mathrm{adv}$ is produced by $\mathbf{x}_\mathrm{adv} = \mathbf{x} + \delta$ that satisfies $ \vert \vert \mathbf{x}_\mathrm{adv} - \mathbf{x} \vert\vert_p \leq \epsilon$, where $\epsilon$ is a predefined maximum allowed distance, and $\delta$ is a small perturbation. The goal is to find $\mathbf{x}_\mathrm{adv}$ that maximizes the attack objective loss function $\mathcal{L}(\cdot,\cdot)$, which can be the cross-entropy loss or the max-margin logit loss. The projected gradient descent (PGD) attack \cite{madry2018towards} iteratively updates adversarial examples as $\mathbf{x}_{\mathrm{adv}} \leftarrow \mathrm{Clip}_\epsilon(\mathbf{x}_{\mathrm{adv}} + \eta \cdot \overline{\bm{\mathrm{g}}})$, where $\mathrm{Clip}_\epsilon$ denotes the clipping operation to restrict images within a $\ell_p$ ball centered at $\mathbf{x}$ with the radius of $\epsilon$; $\eta$ is the learning rate; $\bm{\mathrm{g}}=\nabla_{\mathbf{x}_{\mathrm{adv}}} \mathcal{L}(f(\mathbf{x}_{\mathrm{adv}}), y)$ is the gradient of the objective loss function at $\mathbf{x}_{\mathrm{adv}}$; and $\overline{\bm{\mathrm{g}}}$ denotes a normalized version of $\bm{\mathrm{g}}$ under $\ell_p$ norm, \textit{i.e.,} $\overline{\bm{\mathrm{g}}} = \frac{\bm{\mathrm{g}}}{\vert\vert\bm{\mathrm{g}}\vert \vert_2}$ under $\ell_2$ norm and $\overline{\bm{\mathrm{g}}} = \text{sign} (\bm{\mathrm{g}})$ under $\ell_\infty$ norm. In this study, we also follow above steps to update the adversarial image, except that we do not have access to the true gradient $\bm{\mathrm{g}}$ in the black-box setting and replace it with a suitable direction $\hat{\bm{\mathrm{g}}}$.

	\section{Method}

	The proposed algorithm is illustrated in Fig.~\ref{fig:attack} and Algorithm \ref{alg:attack}. In each iteration, the algorithm first feeds current image $\mathbf{x}_{\mathrm{adv}}$ and its label $t$ to a pre-trained surrogate model $m$, where $\mathbf{x}_{\mathrm{adv}}$ is initialized as the benign image and updated subsequently. To optimize $\mathbf{x}_{\mathrm{adv}}$, we should \textit{maximize} the attack objective loss, which is max-margin logit loss \cite{Carlini2017TowardsET} in untargeted attacks and negative cross-entropy loss in targeted ones. The max-margin logit loss is shown in Eq. \eqref{eqn:cw_loss}:

	\begin{equation}
	\label{eqn:cw_loss}
	\mathcal{L}(\hat{y},t) = \max_{j\neq t} \hat{y}_j - \hat{y}_t
	\end{equation}
	where $\hat{y}$ is the model's logits output of $\mathbf{x}_{\mathrm{adv}}$, $t$ is the true class label $y$, and $j$ indexes the other classes. The negative cross-entropy loss, which is used in the targeted attack, is shown in Eq. \eqref{eqn:xent_loss}.
	
	\begin{equation}
	\label{eqn:xent_loss}
	\mathcal{L}(\hat{y},t) = \log (\frac{\exp(\hat{y}_t)}{\sum_{j=1}^K \exp(\hat{y}_j)})
	\end{equation}
	where $\hat{y}$ is the model's logits output of $\mathbf{x}_{\mathrm{adv}}$, $t$ is the target class label $y_\mathrm{adv}$ of the targeted attack, and $j$ enumerates all $K$ classes. Then, the gradient of the surrogate model $m$'s loss w.r.t. $\mathbf{x}_{\mathrm{adv}}$ is computed as $\hat{\bm{\mathrm{g}}} \gets \nabla_{\mathbf{x}_\mathrm{adv}} \mathcal{L} (m(\mathbf{x}_\mathrm{adv}), t)$ (called surrogate gradient). For convenience, we let $\mathcal{L}(\cdot)$ denote $\mathcal{L}(f(\cdot), t)$ and $\overline{\hat{\bm{\mathrm{g}}}}$ denote the suitable normalization of $\hat{\bm{\mathrm{g}}}$ under $\ell_p$ norm constraint as mentioned before. A direct application of PGD with the true gradient $\bm{\mathrm{g}}$ replaced with $\hat{\bm{\mathrm{g}}}$ yields the update $\mathbf{x}_{\mathrm{adv}}+\eta\cdot \overline{\hat{\bm{\mathrm{g}}}}$. When $\mathcal{L}(\mathbf{x}_{\mathrm{adv}}+\eta\cdot \overline{\hat{\bm{\mathrm{g}}}})\geq \mathcal{L}(\mathbf{x}_{\mathrm{adv}})$, we directly accept the update. However, sometimes directly following $\overline{\hat{\bm{\mathrm{g}}}}$ may fail to increase the loss because $\overline{\hat{\bm{\mathrm{g}}}}$ deviates too much from the true gradient $\overline{\bm{\mathrm{g}}}$. SWITCH proposes to query the loss value to detect such situations in which $\mathcal{L}(\mathbf{x}_{\mathrm{adv}}+\eta\cdot \overline{\hat{\bm{\mathrm{g}}}})< \mathcal{L}(\mathbf{x}_{\mathrm{adv}})$.
	
	When following $\overline{\hat{\bm{\mathrm{g}}}}$ fails to increase the loss, SWITCH algorithm considers following its opposite direction, namely $-\overline{\hat{\bm{\mathrm{g}}}}$. Its effectiveness is based on the approximate linearity assumption of DNNs \cite{goodfellow6572explaining}. Assuming the target model $f$ is differentiable at the current point $\mathbf{x}_{\mathrm{adv}}$, and $\mathcal{L}(\cdot, t)$ is differentiable at $f(\mathbf{x}_{\mathrm{adv}})$, then $\mathcal{L}(f(\mathbf{x}_{\mathrm{adv}}), t)$ is differentiable at $\mathbf{x}_{\mathrm{adv}}$. According to Taylor expansion, we have
	\begin{equation}
	\mathcal{L}(\mathbf{x}_{\mathrm{adv}} + \eta \cdot \overline{\hat{\bm{\mathrm{g}}}}) = \mathcal{L}(\mathbf{x}_{\mathrm{adv}}) + \eta\cdot\overline{\hat{\bm{\mathrm{g}}}}^\top \nabla \mathcal{L}(\mathbf{x}_{\mathrm{adv}}) + o(\eta),
	\end{equation}
	where $\lim_{\eta\to 0} \frac{o(\eta)}{\eta}=0$. Therefore, when $\eta$ is very small, we have
	\begin{equation}
	\mathcal{L}(\mathbf{x}_{\mathrm{adv}} + \eta \cdot \overline{\hat{\bm{\mathrm{g}}}}) \approx \mathcal{L}(\mathbf{x}_{\mathrm{adv}}) + \eta\cdot\overline{\hat{\bm{\mathrm{g}}}}^\top \nabla \mathcal{L}(\mathbf{x}_{\mathrm{adv}}).
	\label{eqn:linear}
	\end{equation}
	According to the approximate linearity assumption, when we set $\eta$ to be a moderate value, this approximation still holds.
	\begin{figure}[t]
		\begin{center}
			\includegraphics[width=0.8\linewidth]{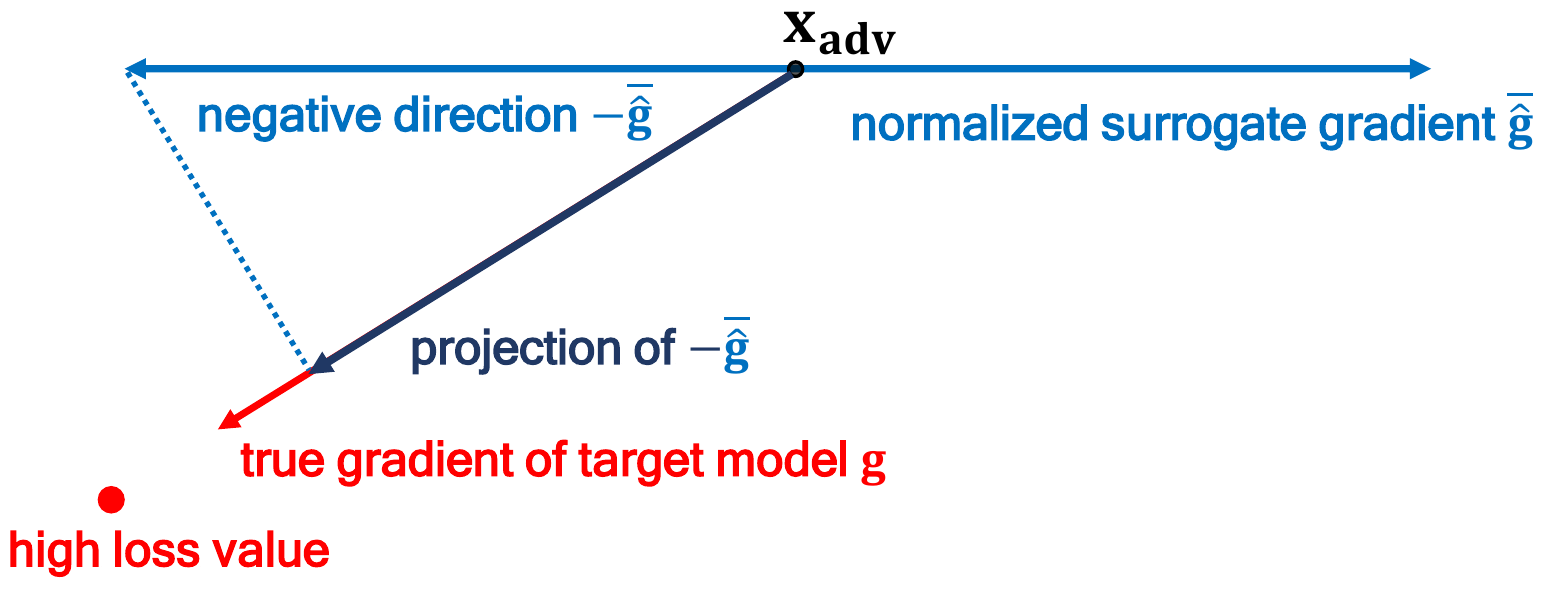}
		\end{center}
		\caption{Illustration of SWITCH, where $\mathbf{x}_\text{adv}$ is the current image to be updated. When the target model $f$ is locally approximately linear, the loss should increase along $-\overline{\hat{\mathbf{g}}}$ direction if it decreases along $\overline{\hat{\mathbf{g}}}$ direction. } 	
		\label{fig:surrogate_gradient}
	\end{figure}
	
	When \eqref{eqn:linear} holds, $\mathcal{L}(\mathbf{x}_{\mathrm{adv}}+\eta\cdot \overline{\hat{\bm{\mathrm{g}}}})< \mathcal{L}(\mathbf{x}_{\mathrm{adv}})$ corresponds to the case in which $\eta\cdot\overline{\hat{\bm{\mathrm{g}}}}^\top \nabla \mathcal{L}(\mathbf{x}_{\mathrm{adv}})<0$. By switching the update direction to the $-\overline{\hat{\bm{\mathrm{g}}}}$, then we have $\mathcal{L}(\mathbf{x}_{\mathrm{adv}} - \eta \cdot \overline{\hat{\bm{\mathrm{g}}}}) \approx \mathcal{L}(\mathbf{x}_{\mathrm{adv}}) - \eta\cdot\overline{\hat{\bm{\mathrm{g}}}}^\top \nabla \mathcal{L}(\mathbf{x}_{\mathrm{adv}})>\mathcal{L}(\mathbf{x}_{\mathrm{adv}})$, which explains the effectiveness of the switch operation. Intuitively,
	assuming the target model $f$ is locally linear at the current point $\mathbf{x}_{\mathrm{adv}}$, then the loss value increases along the $-\overline{\hat{\mathbf{g}}}$ direction if it decreases along the $\overline{\hat{\mathbf{g}}}$ direction. Fig.~\ref{fig:surrogate_gradient} illustrates the reason, since in such cases $-\overline{\hat{\mathbf{g}}}$ would have a positive component along the true gradient $\mathbf{g}$ because the inner product of $-\overline{\hat{\mathbf{g}}}$ and $\mathbf{g}$ is positive.
	
	When $\mathcal{L}(\mathbf{x}_{\mathrm{adv}}+\eta\cdot \overline{\hat{\bm{\mathrm{g}}}})< \mathcal{L}(\mathbf{x}_{\mathrm{adv}})$, since $\eta$ is not infinitesimal or $\left|\overline{\hat{\bm{\mathrm{g}}}}^\top \nabla \mathcal{L}(\mathbf{x}_{\mathrm{adv}})\right|$ is very small, the approximate linearity assumption may not hold. It is possible that $\mathcal{L}(\mathbf{x}_{\mathrm{adv}}-\eta\cdot \overline{\hat{\bm{\mathrm{g}}}})< \mathcal{L}(\mathbf{x}_{\mathrm{adv}})$ as well. Therefore, SWITCH queries the value of $\mathcal{L}(\mathbf{x}_{\mathrm{adv}}-\eta\cdot \overline{\hat{\bm{\mathrm{g}}}})$ to decide whether to follow the proposed opposite direction $-\overline{\hat{\bm{\mathrm{g}}}}$. If both $\mathcal{L}(\mathbf{x}_{\mathrm{adv}}+\eta\cdot \overline{\hat{\bm{\mathrm{g}}}})<\mathcal{L}(\mathbf{x}_{\mathrm{adv}})$ and $\mathcal{L}(\mathbf{x}_{\mathrm{adv}}-\eta\cdot \overline{\hat{\bm{\mathrm{g}}}})<\mathcal{L}(\mathbf{x}_{\mathrm{adv}})$ happen, it means that both $\overline{\hat{\bm{\mathrm{g}}}}$ and $-\overline{\hat{\bm{\mathrm{g}}}}$ cannot lead to a increased loss value. In such tricky situation, our switch strategy is still applicable. We propose two strategies to deal with such situation, namely, SWITCH and SWITCH$_\text{RGF}$.
	SWITCH does not seek other directions but still chooses a relatively better direction from $\overline{\hat{\bm{\mathrm{g}}}}$ and $-\overline{\hat{\bm{\mathrm{g}}}}$. Specifically, SWITCH compares $\mathcal{L}(\mathbf{x}_{\mathrm{adv}}+\eta\cdot \overline{\hat{\bm{\mathrm{g}}}})$ and $\mathcal{L}(\mathbf{x}_{\mathrm{adv}}-\eta\cdot \overline{\hat{\bm{\mathrm{g}}}})$ to identify a better direction between $\overline{\hat{\bm{\mathrm{g}}}}$ and $-\overline{\hat{\bm{\mathrm{g}}}}$. If $\mathcal{L}(\mathbf{x}_{\mathrm{adv}}-\eta\cdot \overline{\hat{\bm{\mathrm{g}}}}) < \mathcal{L}(\mathbf{x}_{\mathrm{adv}}+\eta\cdot \overline{\hat{\bm{\mathrm{g}}}})$, it means that to switch is even worse than not to switch, then SWITCH still follows the original surrogate gradient $\overline{\hat{\bm{\mathrm{g}}}}$. Otherwise, it switches to $-\overline{\hat{\bm{\mathrm{g}}}}$ in current iteration. 
	
	Although SWITCH selects a relatively better update direction from $\{\overline{\hat{\bm{\mathrm{g}}}}, -\overline{\hat{\bm{\mathrm{g}}}}\}$ when $\mathcal{L}(\mathbf{x}_{\mathrm{adv}}+\eta\cdot \overline{\hat{\bm{\mathrm{g}}}})< \mathcal{L}(\mathbf{x}_{\mathrm{adv}})$ and $\mathcal{L}(\mathbf{x}_{\mathrm{adv}}-\eta\cdot \overline{\hat{\bm{\mathrm{g}}}})< \mathcal{L}(\mathbf{x}_{\mathrm{adv}})$, it still leads to a decrease of the loss. Alternatively, in such cases we can try other reasonable updates using more queries such that the loss increases. Thus, we develop an extension of SWITCH called SWITCH$_\text{RGF}$, which uses RGF to obtain an approximate gradient estimate by using $q$ queries (\textit{e.g.,} $q=50$), and then follows such estimated gradient when both $\overline{\hat{\bm{\mathrm{g}}}}$ and $-\overline{\hat{\bm{\mathrm{g}}}}$ are not helpful. Therefore, we can update the current point so that the surrogate gradient may become useful again while keeping the value of the loss function rising. SWITCH$_\text{RGF}$'s query efficiency might be lower than that of SWITCH but its attack success rate is much higher. We can understand the advantage of SWITCH$_\text{RGF}$ from another perspective: SWITCH$_\text{RGF}$ can achieve the similar success rate to RGF but with an unprecedented low number of queries, because it saves the amount of queries used to estimate the gradient in those attack iterations when the surrogate gradient is useful. For example, it can reduce the query number from 50 RGF queries to only one or two queries in these iterations.
	
\begin{algorithm}[t]
	\caption{\small SWITCH Attack under $\ell_p$ norm constraint}
	\label{alg:attack} 
	\begin{algorithmic} 
		\Require
		Input image $\mathbf{x}$, true label $t$ of $\mathbf{x}$ (untargeted attack) or target class $t$ (targeted attack), feed-forward function of the target model $f$, feed-forward function of the surrogate model $m$, attack objective loss function $\mathcal{L}(\cdot,\cdot)$.
		\renewcommand{\algorithmicrequire}{\textbf{Parameters:}}  
		\Require
		Image learning rate $\eta$, the radius $\epsilon$ of $\ell_p$ norm ball, $p\in \{2, \infty\}$, RGF parameter $\delta$, activate SWITCH$_\text{RGF}$ flag USE$_\text{RGF}$.
		\Ensure $\mathbf{x}_\mathrm{adv}$ that satisfies $\Vert \mathbf{x}_\mathrm{adv}-\mathbf{x}\Vert_p \le \epsilon$.
		
	\end{algorithmic}
	\begin{algorithmic}[1]		
		\State Initialize the adversarial example $\mathbf{x}_\mathrm{adv} \gets \mathbf{x}$
		\State $L_{\mathrm{last}} \gets \mathcal{L}(f(\mathbf{x}_\mathrm{adv}), t)$
		\While{\text{\textbf{not} successful}}
		\State $\hat{\bm{\mathrm{g}}} \gets \nabla_{\mathbf{x}_\mathrm{adv}} \mathcal{L} (m(\mathbf{x}_\mathrm{adv}), t)$
		\State $\overline{\hat{\bm{\mathrm{g}}}} \gets \frac{\hat{\bm{\mathrm{g}}}}{\Vert \hat{\bm{\mathrm{g}}} \Vert_2}$ if $p=2$, or $\overline{\hat{\bm{\mathrm{g}}}} \gets \text{sign} (\hat{\bm{\mathrm{g}}})$ if $p=\infty$
		\State $\mathbf{x}_{\mathrm{temp}} \gets \text{Clip}_\epsilon (\mathbf{x}_\mathrm{adv} + \eta \cdot \overline{\hat{\bm{\mathrm{g}}}}) $ \label{line:x_temp}
		\State $L_1 \gets \mathcal{L}(f(\mathbf{x}_{\mathrm{temp}}), t)$ \label{alg:loss1_x_temp}
		\If{$L_1 \geq L_{\mathrm{last}}$} 
		\State $\mathbf{x}_\mathrm{adv} \gets \mathbf{x}_{\mathrm{temp}},\quad  L_{\mathrm{last}} \gets L_1$ \Comment{accept $\mathbf{x}_{\mathrm{temp}}$ directly}
		\Else \Comment{consider switching gradient direction}
		\State $\mathbf{x}_{\mathrm{temp}}^\prime \gets \text{Clip}_\epsilon (\mathbf{x}_\mathrm{adv} - \eta \cdot \overline{\hat{\bm{\mathrm{g}}}}) $ \label{line:x_temp_prime}
		\State $L_2 \gets \mathcal{L}(f(\mathbf{x}_{\mathrm{temp}}^\prime), t)$ \label{alg:loss2_x_temp}
		\If{$L_2 \geq L_\text{last}$}
		\State $\mathbf{x}_\mathrm{adv} \gets \mathbf{x}_{\mathrm{temp}}^\prime,\quad L_{\mathrm{last}} \gets L_2$ \Comment{switch}
		\Else
		\If{USE$_\text{RGF}$}
		\State $\hat{\mathbf{g}}_\text{RGF} \gets \frac{1}{q} \sum_{i=1}^q \left(\frac{\mathcal{L}(f(x+\delta u_i),t) - \mathcal{L}(f(x),t)}{\delta}\cdot u_i \right)$ 
		\State $\mathbf{x}_{\mathrm{adv}} \gets \text{Clip}_\epsilon (\mathbf{x}_\mathrm{adv} + \eta \cdot \hat{\mathbf{g}}_\text{RGF})$
		\State $L_{\mathrm{last}} \gets \mathcal{L}(f(\mathbf{x}_{\mathrm{adv}}),t)$
		\Else
		\If{$L_2 \geq L_1$}
		\State $\mathbf{x}_\mathrm{adv} \gets \mathbf{x}_{\mathrm{temp}}^\prime,\quad L_{\mathrm{last}} \gets L_2$ \Comment{switch}
		\Else
		\State $\mathbf{x}_\mathrm{adv} \gets \mathbf{x}_{\mathrm{temp}},\quad L_{\mathrm{last}} \gets L_1$ \Comment{not switch}
		\EndIf
		\EndIf
		\EndIf
		\EndIf
		\EndWhile
		\State\Return $\mathbf{x}_{\mathrm{adv}}$
	\end{algorithmic}
\end{algorithm}

\section{Experiment}
	
	\subsection{Experiment Setting}
	\begin{table}[h]
		\small
		\tabcolsep=0.1cm
		\caption{Different attack methods adopt different strategies.}
		\label{tab:compared_methods}
		\resizebox{1\linewidth}{!}{
			\begin{tabular}{c|ccc}
				\toprule
				\textbf{Method} & \textbf{Surrogate Model(s)} & \textbf{Gradient Estimation}  & \textbf{Based on Random Search}  \\
				\midrule
				Bandits~\cite{ilyas2018prior} & $\times$ &  $\checkmark$ & $\times$ \\
				RGF~\cite{nesterov2017random} & $\times$ &  $\checkmark$ & $\times$ \\
				P-RGF~\cite{cheng2019improving} & $\checkmark$ &  $\checkmark$ & $\times$ \\
				PPBA \cite{li2020projection} & $\times$ &  $\times$ & $\checkmark$ \\
				Parsimonious \cite{moonICML19} & $\times$ &  $\times$ & $\times$ \\
				SignHunter \cite{al2020sign} & $\times$ &  $\times$ & $\times$ \\
				Square Attack \cite{ACFH2020square} & $\times$ &  $\times$ & $\checkmark$ \\
				SWITCH & $\checkmark$ &  $\times$ & $\times$ \\
				\bottomrule
		\end{tabular}}
		
	\end{table}
	\label{sec:expr_setting}
	\noindent\textbf{Datasets and Target Models.} Three datasets, namely, CIFAR-10 \cite{krizhevsky2009learning}, CIFAR-100 \cite{krizhevsky2009learning}, and TinyImageNet \cite{russakovsky2015imagenet}, are used for the experiments. We randomly select \nn{1000} tested images from their validation sets for evaluation, all of which are correctly classified by Inception-v3 network~\cite{szegedy2016rethinking}. In the CIFAR-10 and CIFAR-100 datasets, we follow Yan \textit{et al}. \cite{guo2019subspace} to select the following target models: (1) a 272-layers PyramidNet+Shakedrop network (PyramidNet-272) \cite{han2017deep,yamada2019shakedrop} trained using AutoAugment \cite{cubuk2019autoaugment}; (2) a model obtained through a neural architecture search called GDAS \cite{dong2019searching}; (3) a wide residual network with 28 layers and 10 times width expansion (WRN-28) \cite{Zagoruyko2016WRN}; and (4) a wide residual network with 40 layers (WRN-40) \cite{Zagoruyko2016WRN}. In the TinyImageNet dataset, we select (1) ResNeXt-101~(32x4d)~\cite{Xie2017ResNext}, (2) ResNeXt-101~(64x4d)~\cite{Xie2017ResNext}, and (3) DenseNet-121 with a growth rate of 32 \cite{Huang_2017_CVPR} as the target models.
		
	\begin{table*}[t]
		\small
		\tabcolsep=0.1cm
		\caption{The ratio of performing switch operation and the corresponding increased loss in all iterations of attacking \nn{1000} images of CIFAR-10 and CIFAR-100 datasets. $L_{\text{switched}} > L_{\text{last}}$ ratio: The ratio of samples with $L_{\text{switched}}$ larger than the loss of previous iteration ($L_{\text{last}}$). $L_{\text{switched}} > L_{\mathbf{x}_{\mathrm{temp}}}$ ratio: the ratio of samples with $L_{\text{switched}}$ larger than the loss of $\mathbf{x}_{\mathrm{temp}}$ that produced by using the surrogate gradient before switching.}
		\label{tab:ablation_study_CIFAR_untargeted}
		\resizebox{1\linewidth}{!}{
			\begin{tabular}{c|c|c|cccc|cccc|cccc}
				\toprule
				\B{Attack Type} & \B{Dataset} & \B{Norm} & \multicolumn{4}{c|}{\B{switch ratio}} &  \multicolumn{4}{c|}{\B{$L_{\text{switched}} > L_{\text{last}}$ ratio}} &  \multicolumn{4}{c}{\B{$L_{\text{switched}} > L_{\mathbf{x}_{\mathrm{temp}}}$ ratio}} \\
				&	& &  PyramidNet-272 & GDAS & WRN-28 & WRN-40 & PyramidNet-272 & GDAS & WRN-28 & WRN-40 &  PyramidNet-272 & GDAS & WRN-28 & WRN-40 \\ 
				\midrule
				\multirow{4}{*}{untargeted} & \multirow{2}{*}{CIFAR-10} &  $\ell_2$  & 62.0\% & 56.1\% & 60.5\% & 57.3\% & 59.8\% & 66.7\% & 60.4\% & 68.9\% & 94.7\% & 96.6\% & 93.3\% & 95.6\% \\
				& & $\ell_\infty$ & 54.8\% & 51.1\% & 52.7\% & 52.2\% & 72.6\% & 77.7\% & 80.0\% & 83.0\% & 98.2\% & 98.8\% & 98.6\% & 99.0\% \\
				\cmidrule(rl){2-15}
				& \multirow{2}{*}{CIFAR-100} & $\ell_2$ & 52.9\% & 51.8\% & 52.7\% & 54.9\% & 74.1\% & 74.4\% & 73.7\% & 69.4\% & 95.5\% & 97.6\% & 97.5\% & 97.0\% \\ 
				& & $\ell_\infty$ &  51.0\% & 49.8\% & 50.6\% & 50.8\% & 86.7\% & 92.3\% & 90.9\% & 86.4\% & 98.9\% & 99.9\% & 99.4\% & 98.8\% \\
				\midrule
				\multirow{4}{*}{targeted} & \multirow{2}{*}{CIFAR-10} &  $\ell_2$ & 75.9\% & 74.2\% & 73.2\% & 73.4\% & 48.8\% & 51.0\% & 47.0\% & 53.3\% & 88.0\% & 88.5\% & 88.3\% & 90.2\% \\
				& & $\ell_\infty$ & 79.6\% & 77.9\% & 75.9\% & 77.6\% & 60.4\% & 58.7\% & 58.5\% & 60.6\% & 96.2\% & 96.3\% & 95.1\% & 96.3\% \\
				\cmidrule(rl){2-15}
				& \multirow{2}{*}{CIFAR-100} & $\ell_2$ & 77.7\% & 76.2\% & 78.2\% & 78.1\% & 53.0\% & 56.0\% & 54.9\% & 53.5\% & 87.5\% & 88.7\% & 88.1\% & 87.2\% \\
				& & $\ell_\infty$ & 83.2\% & 80.1\% & 83.7\% & 82.4\% & 66.8\% & 66.3\% & 66.9\% & 61.0\% & 95.8\% & 96.4\% & 96.3\% & 94.1\% \\
				\bottomrule
		\end{tabular}}
	\end{table*}
	\begin{table*}[ht]
		\small
		\tabcolsep=0.1cm
		\caption{The attack success rates, the average queries, and the average cosine similarities between the surrogate and true gradients with different surrogate models. We perform SWITCH under $\ell_2$ norm to break a WRN-28 network on the CIFAR-10 dataset. ``Random'' represents using a random vector from the normal distribution as the surrogate gradient in each iteration.}
		\label{tab:ablation_study_cosine_grad}
		\resizebox{1\linewidth}{!}{
			\begin{tabular}{c|c|ccccccccc}
				\toprule
				\B{Attack Type} & \B{Metric} & Random & VGG-19(BN) & ResNet-50 & ResNet-110 & PreResNet-110 & ResNeXT-101(8$\times$64d) & ResNeXT-101(16$\times$64d) & DenseNet-100 & DenseNet-190 \\ 
				\midrule
				\multirow{3}{*}{Untargeted} &  Success Rate  & 81.8\% & 96.5\% & 97.4\% &  97.5\% & 95.8\% & 97.7\% & 96.1\% & 96.1\% & 98.6\%  \\
				&  Avg. Query  & 751 & 98 & 103 &  147 & 113 & 81 & 63 & 116 & 48  \\
				&  cosine grad  & 0.026 & 0.11 & 0.117 & 0.108 & 0.138 & 0.226 & 0.235 & 0.221 & 0.184 \\
				\midrule
				\multirow{3}{*}{Targeted} &  Success Rate  & 50.9\% & 82.8\% & 89.3\% &  87.5\% & 84.5\% & 87.4\% & 80.1\% & 86.9\% & 91.8\%  \\
				&  Avg. Query  & 1537 & 891 & 698 &  809 & 807 & 768 & 765 & 713 & 429  \\
				&  cosine grad  & 0.022 & 0.086 & 0.093 & 0.085 & 0.104 & 0.182 & 0.189 & 0.135 & 0.131 \\
				\bottomrule
		\end{tabular}}

	\end{table*}
	
	\noindent\textbf{Method Setting.} In $\ell_2$ norm attacks, the maximum distortion $\epsilon$ is set to 1.0 on CIFAR-10 and CIFAR-100, and 2.0 on TinyImageNet. In $\ell_\infty$ norm attacks, $\epsilon$ is set to $8/255$. The learning rate $\eta$ is set based on experimental performance. In $\ell_2$ norm attacks, $\eta$ is set to $\epsilon/10$. In $\ell_\infty$ norm attacks, $\eta$ is set to 0.003 ($\approx \epsilon/10$), except for untargeted attacks on CIFAR-10 and CIFAR-100 in which $\eta$ is set to 0.01. Surrogate model $m$ is selected as ResNet-110 on CIFAR-10 and CIFAR-100, and ResNet-101 on TinyImageNet. In targeted attacks, the target class is set to $y_{adv} = (y+1)\mod C$, where $y_{adv}$ is the target class, $y$ is the true class, and $C$ is the number of classes. 
	
	\noindent\textbf{Compared Methods.} The compared query-based black-box attacks are listed in Tab.~\ref{tab:compared_methods}. It includes Bandits~\cite{ilyas2018prior}, RGF~\cite{nesterov2017random}, prior-guided RGF (P-RGF)~\cite{cheng2019improving}, Projection \& Probability-driven Black-box Attack (PPBA) \cite{li2020projection}, Parsimonious \cite{moonICML19}, SignHunter \cite{al2020sign}, and Square Attack \cite{ACFH2020square}. These methods consist of different strategies. Bandits, RGF, and P-RGF estimate the approximate gradient to optimize the adversarial example; P-RGF improves query efficiency of RGF by utilizing surrogate models, and uses the same surrogate model $m$ as our method does; PPBA and Square Attack rely on random search optimization; PPBA utilizes a low-frequency constrained sensing matrix; Parsimonious optimizes a discrete surrogate problem by finding the solution among the vertices of the $\ell_\infty$ ball. We use the PyTorch~\cite{paszke2019pytorch} framework for all experiments, and the official PyTorch implementations of PPBA, Bandits, SignHunter, and Square Attack are used in the experiments. For experimental consistency, we translate the codes of Parsimonious, RGF, and P-RGF from the official TensorFlow~\cite{abadi2016tensorflow} version into the PyTorch version. Parsimonious only supports $\ell_\infty$ norm attacks. We set the maximum number of queries to \nn{10000} in all experiments. We set the same $\epsilon$ value as our method to limit the perturbation of these attacks. The default parameters of these methods are presented in the supplementary material.
	
	\noindent\textbf{Evaluation Metric.} Following previous studies~\cite{cheng2019improving,guo2019subspace}, we report a successful attack if the attack successfully finds an adversarial example within \nn{10000} queries. We report the attack success rate and the average/median number of queries over successful attacks. To present more comprehensive metrics, we also report the average and median query over all the samples by setting the queries of failed samples to \nn{10000} in the supplementary material.

	\subsection{Ablation Study}
	
	\begin{table*}[t]
		\small	
		\tabcolsep=0.1cm
		\caption{Experimental results of untargeted attack on CIFAR-10 and CIFAR-100 datasets.}
		\label{tab:CIFAR_untargetd_result}
		\resizebox{1\linewidth}{!}{
			\begin{tabular}{c|c|c|cccc|cccc|cccc}
				\toprule
				\B Dataset & \B{Norm} & \B{Attack} & \multicolumn{4}{c|}{\B{Attack Success Rate}} &  \multicolumn{4}{c|}{\B{Avg. Query}} &  \multicolumn{4}{c}{\B{Median Query}} \\
				& & & PyramidNet-272 & GDAS & WRN-28 & WRN-40 & PyramidNet-272 & GDAS & WRN-28 & WRN-40 & PyramidNet-272 & GDAS & WRN-28 & WRN-40 \\ 
				\midrule
				\multirow{21}{*}{CIFAR-10} & \multirow{10}{*}{$\ell_2$} & RGF \cite{nesterov2017random} & 100\% & 98.9\% & 99.4\% & 100\% & 1173 & 889 & 1622 & 1352 & 1020 & 667 & 1173 & 1071 \\
				& & P-RGF \cite{cheng2019improving} & 100\% & 99.8\% & 99.1\% & 99.6\% & 853 & 473 & 1101 & 831 & 666 & 239 & 468 & 428 \\
				& & Bandits  \cite{ilyas2018prior} & 100\% & 100\% & 99.4\% & 99.7\% & 692 & 332 & 807 & 665 & 474 & 210 & 334 & 310 \\
				& & PPBA \cite{li2020projection} & 100\% & 99.9\% & 97.1\% & 99.1\% & 1056 & 634 & 1339 & 1118 & 833 & 485 & 754 & 734 \\
				& & SignHunter \cite{al2020sign} & 97\% & 85.8\% & 97.7\% & 99\% & 794 & 512 & 1209 & 1038 & 487 & 267 & 595 & 564 \\
				& & Square Attack \cite{ACFH2020square} & 99.8\% & 100\% & 96.6\% & 98.4\% & 767 & 363 & 866 & 744 & 435 & 160 & 315 & 308 \\
				& & NO SWITCH & 39.8\% & 77.1\% & 80.5\% & 82.3\% & 152 & 58 & 39 & 53 & 21 & 12 & 10 & 9 \\
				& & SWITCH & 90\% & 95.5\% & 97.3\% & 97.1\% & 490 & 121 & 139 & 83 & 67 & 17 & 14 & 13 \\
				& & SWITCH$_\text{RGF}$ & 100\% & 100\% & 99.4\% & 100\% & 555 & 326 & 638 & 516 & 369 & 107 & 60 & 27 \\
				\cmidrule(rl){2-15} & \multirow{11}{*}{$\ell_\infty$} & RGF \cite{nesterov2017random} & 98.8\% & 93.8\% & 98.7\% & 99.1\% & 942 & 645 & 1194 & 960 & 667 & 460 & 663 & 612 \\
				& & P-RGF \cite{cheng2019improving} & 97.3\% & 97.9\% & 97.7\% & 98\% & 742 & 337 & 703 & 564 & 408 & 128 & 236 & 217 \\
				& & Bandits \cite{ilyas2018prior} & 99.6\% & 100\% & 99.4\% & 99.9\% & 1015 & 391 & 611 & 542 & 560 & 166 & 224 & 228 \\
				& & PPBA \cite{li2020projection} & 96.6\% & 99.5\% & 96.8\% & 97.6\% & 860 & 334 & 698 & 574 & 322 & 122 & 188 & 196 \\
				& & Parsimonious \cite{moonICML19} & 100\% & 100\% & 100\% & 100\% & 701 & 345 & 891 & 738 & 523 & 230 & 424 & 375 \\
				& & SignHunter \cite{al2020sign} & 99.4\% & 91.8\% & 100\% & 100\% & 379 & 253 & 506 & 415 & 189 & 106 & 205 & 188 \\
				& & Square Attack \cite{ACFH2020square} & 100\% & 100\% & 99.9\% & 100\% & 332 & 126 & 403 & 342 & 182 & 54 & 144 & 139 \\
				& & NO SWITCH & 48.9\% & 84.3\% & 89.4\% & 91.8\% & 422 & 147 & 72 & 120 & 12 & 4 & 3 & 3 \\
				& & SWITCH & 77.2\% & 96\% & 97.2\% & 98\% & 522 & 133 & 119 & 68 & 29 & 4 & 3 & 3 \\
				& & SWITCH$_{\text{RGF}}$ & 92.8\% & 98.2\% & 97.4\% & 98.2\% & 525 & 143 & 254 & 218 & 161 & 5 & 3 & 3 \\
				\midrule
				\multirow{21}{*}{CIFAR-100} & \multirow{10}{*}{$\ell_2$} & RGF \cite{nesterov2017random} & 100\% & 99.8\% & 99.6\% & 99.5\% & 565 & 554 & 876 & 985 & 459 & 408 & 663 & 612 \\
				& & P-RGF \cite{cheng2019improving} & 100\% & 99.8\% & 99.6\% & 99.3\% & 407 & 281 & 565 & 765 & 280 & 180 & 296 & 356 \\
				& & Bandits  \cite{ilyas2018prior} & 100\% & 100\% & 99.9\% & 99.6\% & 188 & 170 & 312 & 325 & 96 & 90 & 148 & 134 \\
				& & PPBA \cite{li2020projection} & 100\% & 100\% & 99.8\% & 99.4\% & 439 & 362 & 623 & 715 & 322 & 280 & 411 & 419 \\
				& & SignHunter \cite{al2020sign} & 97.5\% & 94.1\% & 99.1\% & 99.1\% & 293 & 219 & 565 & 586 & 128 & 82 & 183 & 192 \\
				& & Square Attack \cite{ACFH2020square} & 100\% & 100\% & 99.5\% & 99.2\% & 217 & 139 & 326 & 372 & 98 & 55 & 120 & 119 \\
				& & NO SWITCH & 68.5\% & 84.2\% & 73.6\% & 74.7\% & 113 & 96 & 58 & 63 & 11 & 8 & 11 & 10 \\
				& & SWITCH & 95.7\% & 98.5\% & 96.2\% & 96.1\% & 128 & 85 & 141 & 124 & 19 & 11 & 15 & 17 \\
				& & SWITCH$_\text{RGF}$ & 100\% & 100\% & 99.9\% & 99.7\% & 241 & 160 & 306 & 395 & 107 & 15 & 66 & 84 \\
				\cmidrule(rl){2-15} & \multirow{11}{*}{$\ell_\infty$} & RGF \cite{nesterov2017random} & 99.8\% & 98.8\% & 99\% & 99\% & 389 & 423 & 552 & 627 & 256 & 255 & 357 & 357 \\
				& & P-RGF \cite{cheng2019improving} & 99.3\% & 98.7\% & 97.6\% & 97.8\% & 308 & 238 & 351 & 478 & 147 & 116 & 138 & 181 \\
				& & Bandits \cite{ilyas2018prior} & 100\% & 100\% & 99.8\% & 99.8\% & 266 & 209 & 262 & 260 & 68 & 56 & 106 & 92 \\
				& & PPBA \cite{li2020projection} & 99.6\% & 99.9\% & 98.7\% & 98.6\% & 251 & 168 & 349 & 279 & 105 & 73 & 128 & 77 \\
				& & Parsimonious \cite{moonICML19} & 100\% & 100\% & 100\% & 99.9\% & 287 & 185 & 383 & 422 & 204 & 103 & 215 & 213 \\
				& & SignHunter \cite{al2020sign} & 99\% & 97.3\% & 99.8\% & 100\% & 125 & 119 & 211 & 255 & 52 & 37 & 59 & 60 \\
				& & Square Attack \cite{ACFH2020square} & 100\% & 100\% & 100\% & 99.8\% & 76 & 57 & 150 & 162 & 17 & 19 & 46 & 36 \\
				& & NO SWITCH & 81.8\% & 92.1\% & 86.4\% & 87.2\% & 243 & 84 & 126 & 191 & 3 & 3 & 3 & 3 \\
				& & SWITCH & 93.4\% & 97.5\% & 93.5\% & 94.3\% & 145 & 64 & 109 & 91 & 4 & 3 & 4 & 4 \\
				& & SWITCH$_\text{RGF}$ & 99.3\% & 98.5\% & 98.4\% & 97.7\% & 136 & 67 & 166 & 144 & 6 & 3 & 4 & 4 \\
				\bottomrule
		\end{tabular}}

	\end{table*}
	
	\begin{table*}[t]
		\small
		\tabcolsep=0.1cm
		\caption{Experimental results of targeted attack under $\ell_2$ norm constraint on CIFAR-10 and CIFAR-100 datasets.}
		\label{tab:CIFAR_targetd_l2_result}
		\resizebox{1\linewidth}{!}{
			\begin{tabular}{c|c|cccc|cccc|cccc}
				\toprule
				\B Dataset & \B Attack & \multicolumn{4}{c|}{\B{Attack Success Rate}} &  \multicolumn{4}{c|}{\B{Avg. Query}} &  \multicolumn{4}{c}{\B{Median Query}} \\
				& &  PyramidNet-272 & GDAS & WRN-28 & WRN-40 & PyramidNet-272 & GDAS & WRN-28 & WRN-40 & PyramidNet-272 & GDAS & WRN-28 & WRN-40 \\ 
				\midrule
				\multirow{9}{*}{CIFAR-10} & RGF \cite{nesterov2017random} & 100\% & 100\% & 97.5\% & 99.8\% & 1795 & 1249 & 2454 & 2176 & 1632 & 1071 & 2040 & 1938 \\
				&  P-RGF \cite{cheng2019improving} & 99.9\% & 100\% & 97.1\% & 99.7\% & 1698 & 1149 & 2140 & 1738 & 1548 & 966 & 1613 & 1444 \\
				&  Bandits  \cite{ilyas2018prior} & 96.3\% & 100\% & 81.5\% & 84.2\% & 1966 & 905 & 2092 & 2088 & 1426 & 650 & 1268 & 1240 \\
				&  PPBA \cite{li2020projection} & 96\% & 99.8\% & 82.1\% & 88.1\% & 2062 & 1180 & 2623 & 2610 & 1479 & 885 & 1887 & 1845 \\
				&  SignHunter \cite{al2020sign} & 98.9\% & 99.9\% & 92.7\% & 97.3\% & 1681 & 1047 & 1953 & 1863 & 1409 & 791 & 1510 & 1480 \\
				&  Square Attack \cite{ACFH2020square} & 95.1\% & 99.5\% & 88.8\% & 93.5\% & 1700 & 885 & 1983 & 1697 & 1070 & 442 & 1070 & 1024 \\
				&  NO SWITCH & 14.8\% & 43.1\% & 53.2\% & 56.1\% & 180 & 114 & 73 & 120 & 46 & 28 & 23 & 22 \\
				&  SWITCH & 71.4\% & 89.2\% & 89.6\% & 90.7\% & 1245 & 526 & 587 & 550 & 296 & 92 & 85 & 78 \\
				&  SWITCH$_\text{RGF}$ & 100\% & 100\% & 97.6\% & 99.8\% & 1146 & 746 & 1566 & 1325 & 894 & 482 & 851 & 841 \\
				\midrule
				\multirow{9}{*}{CIFAR-100} & RGF \cite{nesterov2017random} & 99.9\% & 99.6\% & 98.2\% & 96.2\% & 1542 & 1538 & 2374 & 2775 & 1326 & 1275 & 1989 & 2346 \\
				& P-RGF \cite{cheng2019improving} & 99.9\% & 99.7\% & 97.7\% & 96.1\% & 1528 & 1445 & 2295 & 2662 & 1288 & 1184 & 1870 & 2196 \\
				& Bandits  \cite{ilyas2018prior} & 97.5\% & 98.4\% & 82.3\% & 62.3\% & 2063 & 1620 & 3003 & 3200 & 1418 & 1080 & 2204 & 2344 \\
				&  PPBA \cite{li2020projection} & 95.4\% & 96.4\% & 82.3\% & 63.8\% & 2087 & 1606 & 3008 & 3477 & 1568 & 1140 & 2274 & 2874 \\
				& SignHunter \cite{al2020sign} & 98.4\% & 96.4\% & 91.6\% & 89.4\% & 1494 & 1423 & 1938 & 2078 & 1176 & 1072 & 1505 & 1610 \\
				&  Square Attack \cite{ACFH2020square} & 96.6\% & 94.2\% & 87.4\% & 85.1\% & 1469 & 1281 & 2091 & 2242 & 873 & 707 & 1250 & 1468 \\
				& NO SWITCH & 9.6\% & 14.6\% & 11.1\% & 11.6\% & 788 & 622 & 776 & 627 & 65 & 42 & 62 & 64 \\
				& SWITCH & 73.6\% & 78\% & 74.2\% & 73.5\% & 997 & 748 & 949 & 1077 & 309 & 164 & 276 & 313 \\
				& SWITCH$_\text{RGF}$ & 99.9\% & 99.8\% & 97.8\% & 96.4\% & 1101 & 992 & 1743 & 2078 & 839 & 698 & 1259 & 1538 \\
				\bottomrule
		\end{tabular}}
		
	\end{table*}
	\noindent\textbf{The Ratio of Increased Loss with SWITCH}. To inspect the effectiveness of SWITCH, we conduct one-step empirical analysis on CIFAR-10 and CIFAR-100 datasets, \textit{i.e.,} we check whether the loss could be increased or decreased by following the switched gradients. Specifically, we first count the proportion of switch operations performed in all attack iterations (switch ratio). Then, only the samples that are updated by using the switched gradient are sent to the target model to record a loss $L_{\text{switched}}$. The ratio of samples with $L_{\text{switched}}$ larger than the loss of previous iteration ($L_{\text{last}}$) is counted as ``$L_{\text{switched}} > L_{\text{last}}$ ratio''. The ratio of $L_{\text{switched}} > L_{\mathbf{x}_{\mathrm{temp}}}$ is counted to check whether the switched gradient is better than the gradient before switching, where $\mathbf{x}_{\mathrm{temp}}$ is the image made by using the surrogate gradient before switching. Results of Tab.~\ref{tab:ablation_study_CIFAR_untargeted} show that (1) switching is better than not switching (see $L_{\text{switched}} > L_{\mathbf{x}_{\mathrm{temp}}}$ ratio); (2) switching gradients are rather effective in optimization, since the losses of a large fraction of samples are increased from the last iteration after following the switched gradient ($L_{\text{switched}} > L_{\text{last}}$) even though they have  $L_{\mathbf{x}_{\mathrm{temp}}} < L_{\text{last}}$; (3) targeted attacks perform more switch operations than untargeted ones because their surrogate gradients deviates too much from true gradients.
	
	\noindent\textbf{Transferability of Gradient Directions}. We study the effects of SWITCH with different surrogate models. Tab. \ref{tab:ablation_study_cosine_grad} shows the relationship between the query number and the cosine similarity w.r.t. the true gradients. It shows that with a larger cosine similarity, the query efficiency is higher. This is because the a higher cosine similarity implies a more useful surrogate gradient so that the improvement of the loss in each iteration could be larger, thus saving the overall queries. Specifically, the cosine similarity (w.r.t. the true gradient) of the gradient of a surrogate model is significantly larger than that of a random direction, demonstrating the necessity and effectiveness of utilizing the transferable gradient.
	
	\begin{table}[t]
		\small
		\tabcolsep=0.1cm
		\caption{Experimental results of untargeted attacks under $\ell_2$ norm on TinyImageNet dataset. D$_{121}$: DenseNet-121, R$_{32}$: ResNeXt-101~(32$\times$4d), R$_{64}$: ResNeXt-101~(64$\times$4d).}
		\label{tab:TinyImageNet_untargeted_l2_result}
		\resizebox{1\linewidth}{!}{
			\begin{tabular}{c|ccc|ccc|ccc}
				\toprule
				\B{Attack} & \multicolumn{3}{c|}{\B{Attack Success Rate}} &  \multicolumn{3}{c|}{\B{Avg. Query}} &  \multicolumn{3}{c}{\B{Median Query}} \\
				&  D$_{121}$ & R$_{32}$ & R$_{64}$  & D$_{121}$ & R$_{32}$ & R$_{64}$ & D$_{121}$ & R$_{32}$ & R$_{64}$ \\ 
				\midrule
				RGF \cite{nesterov2017random} & 99.4\% & 94.1\% & 96.1\% & 1485 & 2651 & 2579 & 1173 & 2052 & 2000 \\
				P-RGF \cite{cheng2019improving} & 99.5\% & 94.2\% & 94.8\% & 1218 & 2092 & 1985 & 792 & 1288 & 1272 \\
				Bandits \cite{ilyas2018prior} & 98.7\% & 93\% & 94.6\% & 1032 & 2005 & 1896 & 576 & 1198 & 1184 \\
				PPBA \cite{li2020projection} & 99.7\% & 96.9\% & 97.5\% & 1151 & 2004 & 1838 & 851 & 1445 & 1396 \\
				SignHunter \cite{al2020sign} & 97.6\% & 93\% & 94.4\% & 560 & 1290 & 1238 & 188 & 585 & 601 \\
				Square Attack \cite{ACFH2020square} & 98.5\% & 92.2\% & 93.6\% & 566 & 1287 & 1164 & 187 & 501 & 416 \\
				NO SWITCH & 25.6\% & 17.8\% & 16.1\% & 399 & 641 & 499 & 35 & 44 & 39 \\
				SWITCH & 98.6\% & 97.1\% & 96.9\% & 174 & 307 & 299 & 54 & 75 & 77 \\
				SWITCH$_\text{RGF}$ & 99.4\% & 97.7\% & 97.4\% & 1010 & 1700 & 1552 & 640 & 1003 & 1000 \\
				\bottomrule
		\end{tabular}}
	\end{table}
	
	\begin{table}[t]
		\small
		\tabcolsep=0.1cm
		\caption{Experimental results of targeted attack under $\ell_2$ norm on TinyImageNet dataset. D$_{121}$: DenseNet-121, R$_{32}$: ResNeXt-101~(32$\times$4d), R$_{64}$: ResNeXt-101~(64$\times$4d).}
		\label{tab:TinyImageNet_targeted_l2_result}
		\resizebox{1\linewidth}{!}{
			\begin{tabular}{c|ccc|ccc|ccc}
				\toprule
				\B{Attack} & \multicolumn{3}{c|}{\B{Attack Success Rate}} &  \multicolumn{3}{c|}{\B{Avg. Query}} &  \multicolumn{3}{c}{\B{Median Query}} \\
				& D$_{121}$ & R$_{32}$ & R$_{64}$  & D$_{121}$ & R$_{32}$ & R$_{64}$ & D$_{121}$ & R$_{32}$ & R$_{64}$ \\ 
				\midrule
				RGF \cite{nesterov2017random} & 87.6\% & 71.6\% & 75.9\% & 3735 & 4845 & 4821 & 3264 & 4335 & 4386 \\
				P-RGF \cite{cheng2019improving} & 84.1\% & 67.9\% & 72.1\% & 3714 & 4584 & 4620 & 3252 & 4148 & 4212 \\
				Bandits \cite{ilyas2018prior} & 56.4\% & 36.8\% & 36.6\% & 3907 & 4229 & 4503 & 3586 & 3818 & 4236 \\
				PPBA \cite{li2020projection} & 70.1\% & 50.3\% & 51.9\% & 3734 & 4256 & 4488 & 3120 & 3887 & 4111 \\
				SignHunter \cite{al2020sign} & 85.3\% & 67.4\% & 72.2\% & 2650 & 3342 & 3412 & 1914 & 2663 & 2901 \\
				Square Attack \cite{ACFH2020square} & 70.1\% & 49.3\% & 53.9\% & 2595 & 3287 & 3391 & 1777 & 2541 & 2603 \\
				NO SWITCH & 4.3\% & 3.5\% & 3.6\% & 889 & 201 & 610 & 69 & 62 & 98 \\
				SWITCH & 74.7\% & 68.4\% & 68.5\% & 1241 & 1453 & 1532 & 536 & 538 & 610 \\
				SWITCH$_\text{RGF}$ & 83\% & 71.1\% & 75.8\% & 3486 & 4111 & 4150 & 2880 & 3603 & 3710 \\    
				\bottomrule
		\end{tabular}}

	\end{table}
	
	\noindent\textbf{NO SWITCH versus SWITCH versus SWITCH$_\text{RGF}$.} To validate the effectiveness of the gradient switch operation in terms of the attack success rate, we consider a baseline NO SWITCH which is the algorithm without switching the surrogate gradient. NO~SWITCH consecutively follows the gradient $\hat{\mathbf{g}}$ of the surrogate model $m$ during the optimization. The results are presented in all the  tables of experimental results (Tabs.~\ref{tab:CIFAR_untargetd_result}, \ref{tab:CIFAR_targetd_l2_result}, \ref{tab:TinyImageNet_untargeted_l2_result}, \ref{tab:TinyImageNet_targeted_l2_result}, \ref{tab:TinyImageNet_defensive_models_linf_result}) and Fig. \ref{fig:query_threshold_attack_success_rate}. We derive the following conclusions based on these results: 
	
	(1) NO~SWITCH requires only one query in each attack iteration which is fewer than SWITCH and SWITCH$_\text{RGF}$, thereby resulting high query efficiency. However, NO~SWITCH obtains unsatisfactory attack success rate and its success rate can hardly exceed 90\%. 
	
	\definecolor{mypink}{RGB}{219, 48, 122}
	\definecolor{greenyellow}{rgb}{0.68, 0.7, 0.18} 
	
	(2) SWITCH and SWITCH$_\text{RGF}$ outperforms NO~SWITCH in terms of attack success rate by a large margin. Although SWITCH$_\text{RGF}$ uses the most queries among three SWITCH versions, it significantly surpasses SWITCH in terms of the success rate. Readers can refer to Fig.~\ref{fig:query_threshold_attack_success_rate} for more details, which demonstrates \textcolor{red}{SWITCH$_\text{RGF}$} and \textcolor{greenyellow}{SWITCH} outperform \textcolor{mypink}{NO SWITCH} in different query budgets.
	\subsection{Comparisons with State-of-the-Arts}
	
	\begin{table}[t]
		\small
		\tabcolsep=0.1cm
		\caption{Experimental results of attacking defensive models under $\ell_\infty$ norm constraint on the TinyImageNet dataset, where CD is ComDefend \cite{jia2019comdefend} and FD is Feature Distillation \cite{liu2019feature}. All defensive models adopt ResNet-50 as the backbone.}
		\label{tab:TinyImageNet_defensive_models_linf_result}
		\resizebox{1\linewidth}{!}{
			\begin{tabular}{c|ccc|ccc}
				\toprule
				\B{Attack} & \multicolumn{3}{c|}{\B{Attack Success Rate}} &  \multicolumn{3}{c}{\B{Avg. Query}}\\
				& CD \cite{jia2019comdefend} & FD \cite{liu2019feature} & JPEG \cite{guo2018countering} & CD \cite{jia2019comdefend} & FD \cite{liu2019feature} & JPEG \cite{guo2018countering}  \\ 
				\midrule
				RGF \cite{nesterov2017random}  & 31.3\% & 10.2\% & 3.2\% & 2446 & 2433 & 2552   \\
				P-RGF \cite{cheng2019improving} & 37.3\% & 25.2\%  & 10.2\%  & 1946 & 1958 & 2605  \\
				Bandits \cite{ilyas2018prior} & 39.6\% & 12.5\% & 8.1\% & 893 & 1272 &  1698  \\
				PPBA \cite{li2020projection} & 18.3\% & 72.9\% &  48\%  & 528 & 1942 &  2387  \\
				Parsimonious \cite{moonICML19} & 83.9\% & 99.5\% & 96.1\% &866 & 359 & 410  \\
				SignHunter \cite{al2020sign} & 87.6\% & 40.3\% & 99.5\% & 208 & 558 & 260  \\
				Square Attack \cite{ACFH2020square} & 36.1\% & 99.5\%& 97.9\% & 57 & 167 & 301 \\
				NO SWITCH & 55.9\% & 40\% &  45.8\% &757 & 843 &  634  \\
				SWITCH & 92.5\% & 93.2\% & 80\% & 365 & 353 & 698 \\
				SWITCH$_\text{RGF}$ & 74.7\% & 69.7\% & 43.6\% & 1448 & 1723 & 2442 \\
				\bottomrule
		\end{tabular}}
		\vspace{-1cm}
	\end{table}

	\begin{figure*}[t]
		\captionsetup[sub]{font={tiny}}
		\centering 
		\small
		\begin{minipage}[b]{.24\textwidth}
			\includegraphics[width=\linewidth]{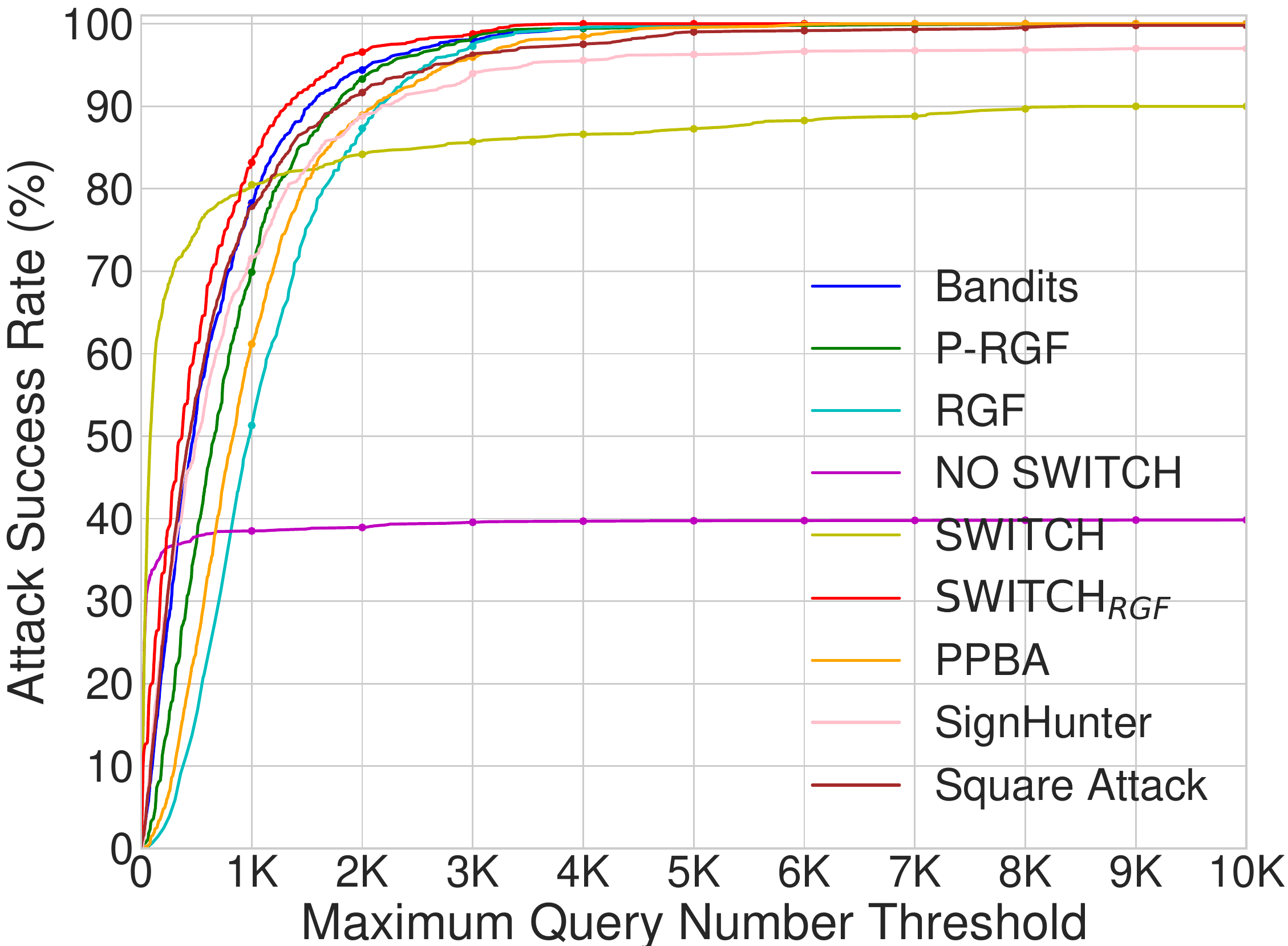}
			\subcaption{Untargeted $\ell_2$ attack PyramidNet-272 on CIFAR-10}
			\label{fig:untargeted_l2_pyramidnet272_CIFAR-10}
		\end{minipage}
		\begin{minipage}[b]{.24\textwidth}
			\includegraphics[width=\linewidth]{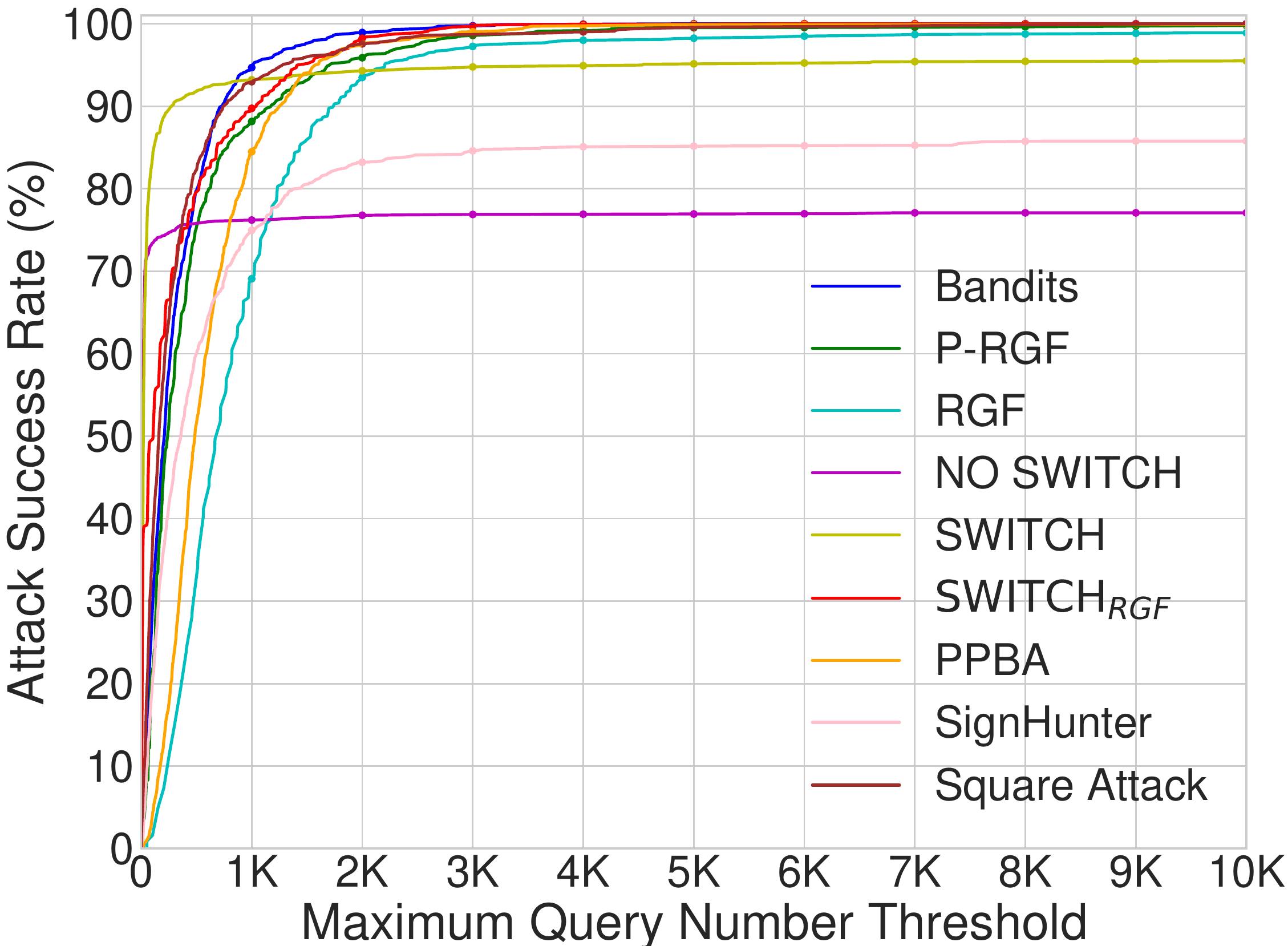}
			\subcaption{Untargeted $\ell_2$ attack GDAS on CIFAR-10}
			\label{fig:untargeted_l2_gdas_CIFAR-10}
		\end{minipage}
		\begin{minipage}[b]{.24\textwidth}
			\includegraphics[width=\linewidth]{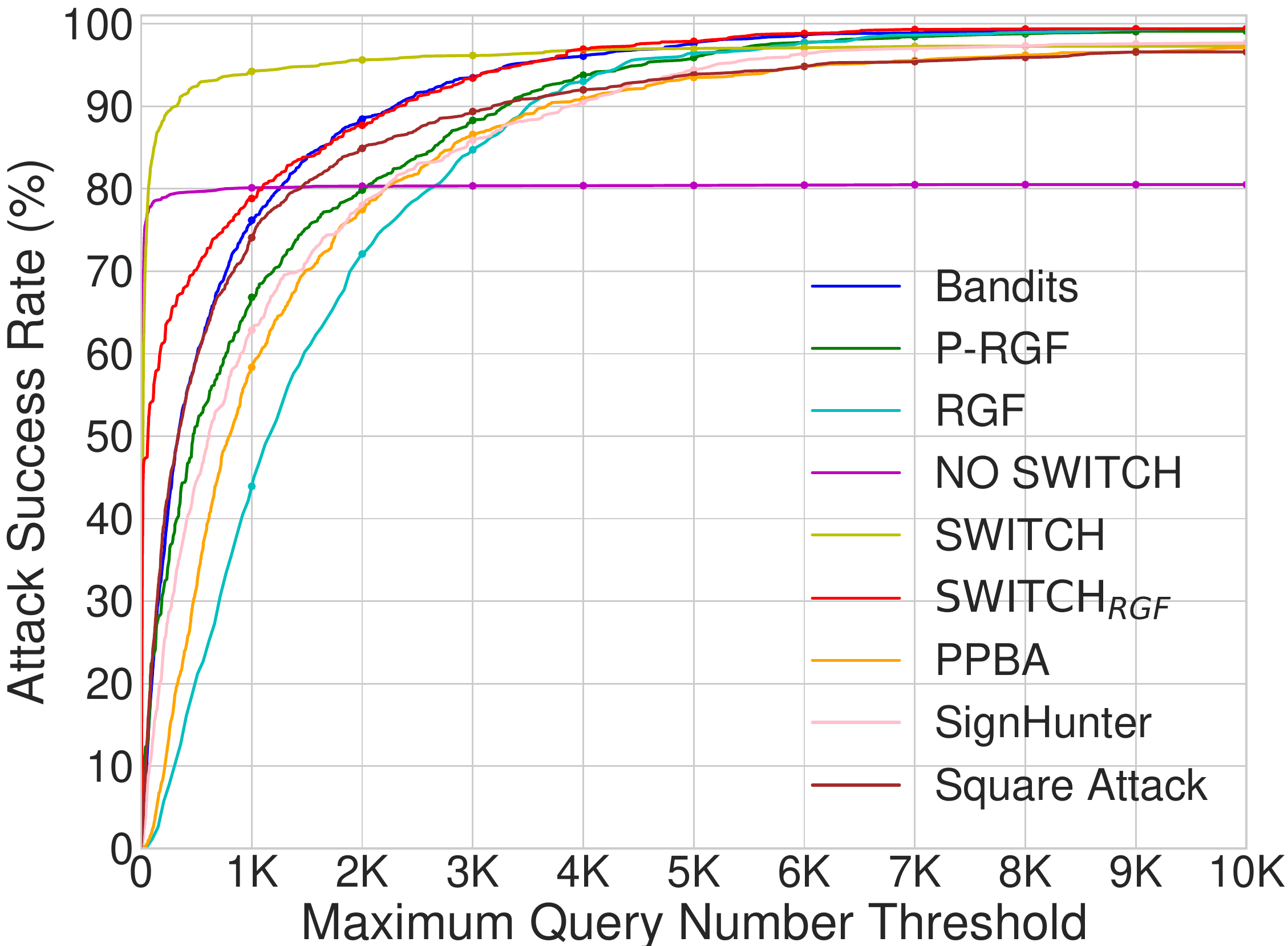}
			\subcaption{Untargeted $\ell_2$ attack WRN-28 on CIFAR-10}
			\label{fig:untargeted_l2_WRN-28_CIFAR-10}
		\end{minipage}
		\begin{minipage}[b]{.24\textwidth}
			\includegraphics[width=\linewidth]{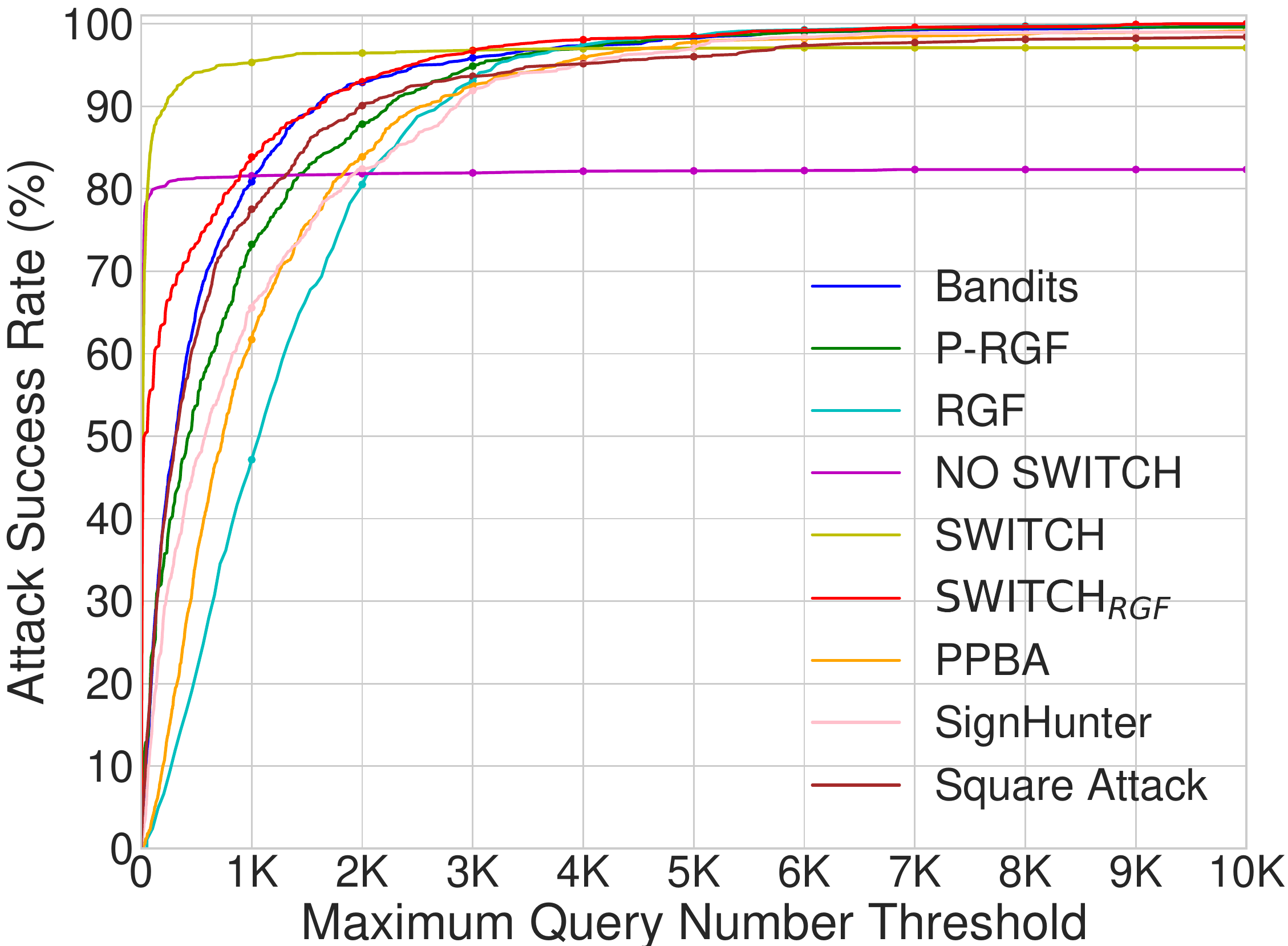}
			\subcaption{Untargeted $\ell_2$ attack WRN-40 on CIFAR-10}
			\label{fig:untargeted_l2_WRN-40_CIFAR-10}
		\end{minipage}
		\begin{minipage}[b]{.24\textwidth}
			\includegraphics[width=\linewidth]{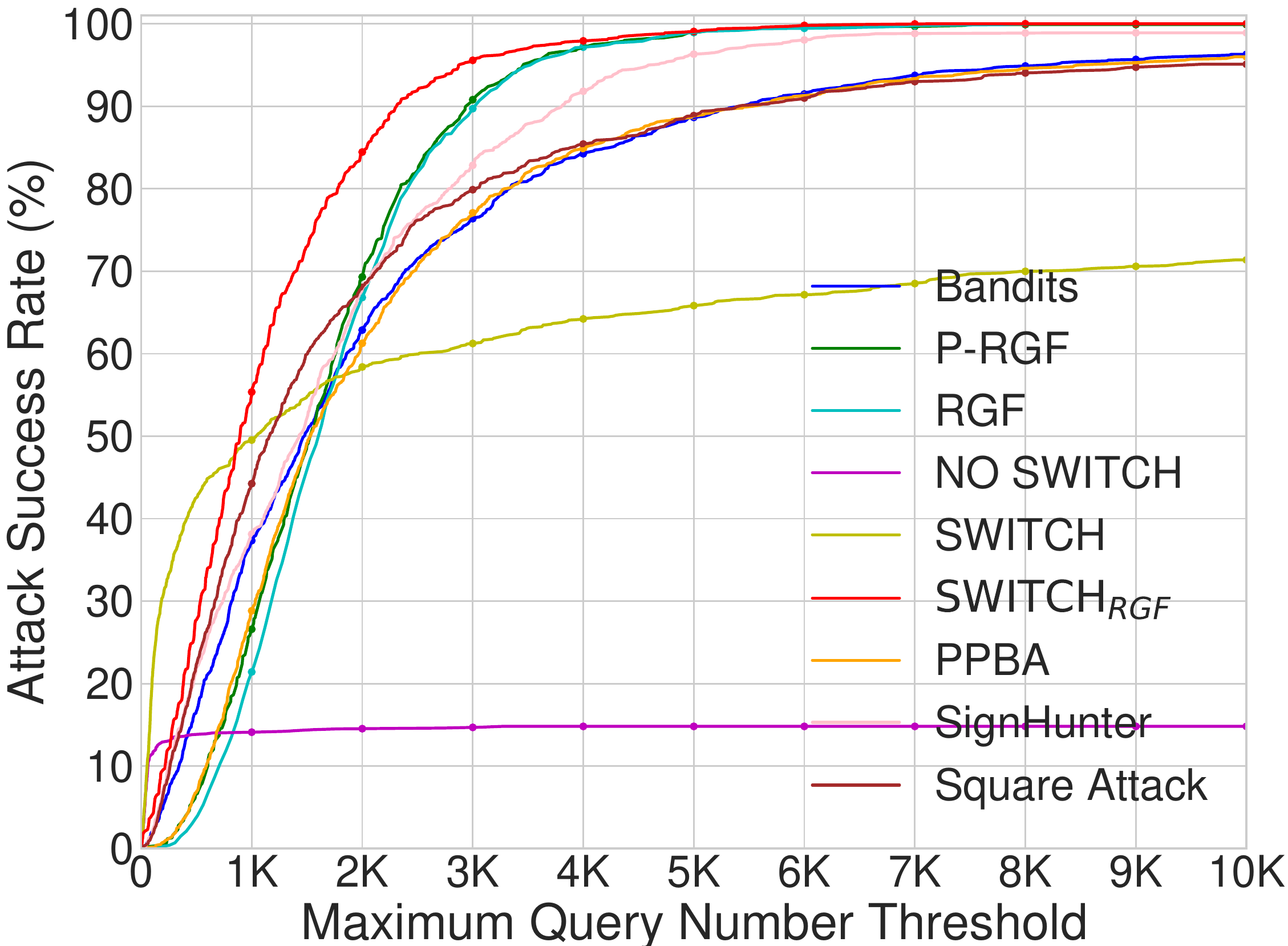}
			\subcaption{Targeted $\ell_2$ attack PyramidNet-272 on CIFAR-10}
			\label{fig:targeted_l2_pyramidnet272_CIFAR-10}
		\end{minipage}
		\begin{minipage}[b]{.24\textwidth}
			\includegraphics[width=\linewidth]{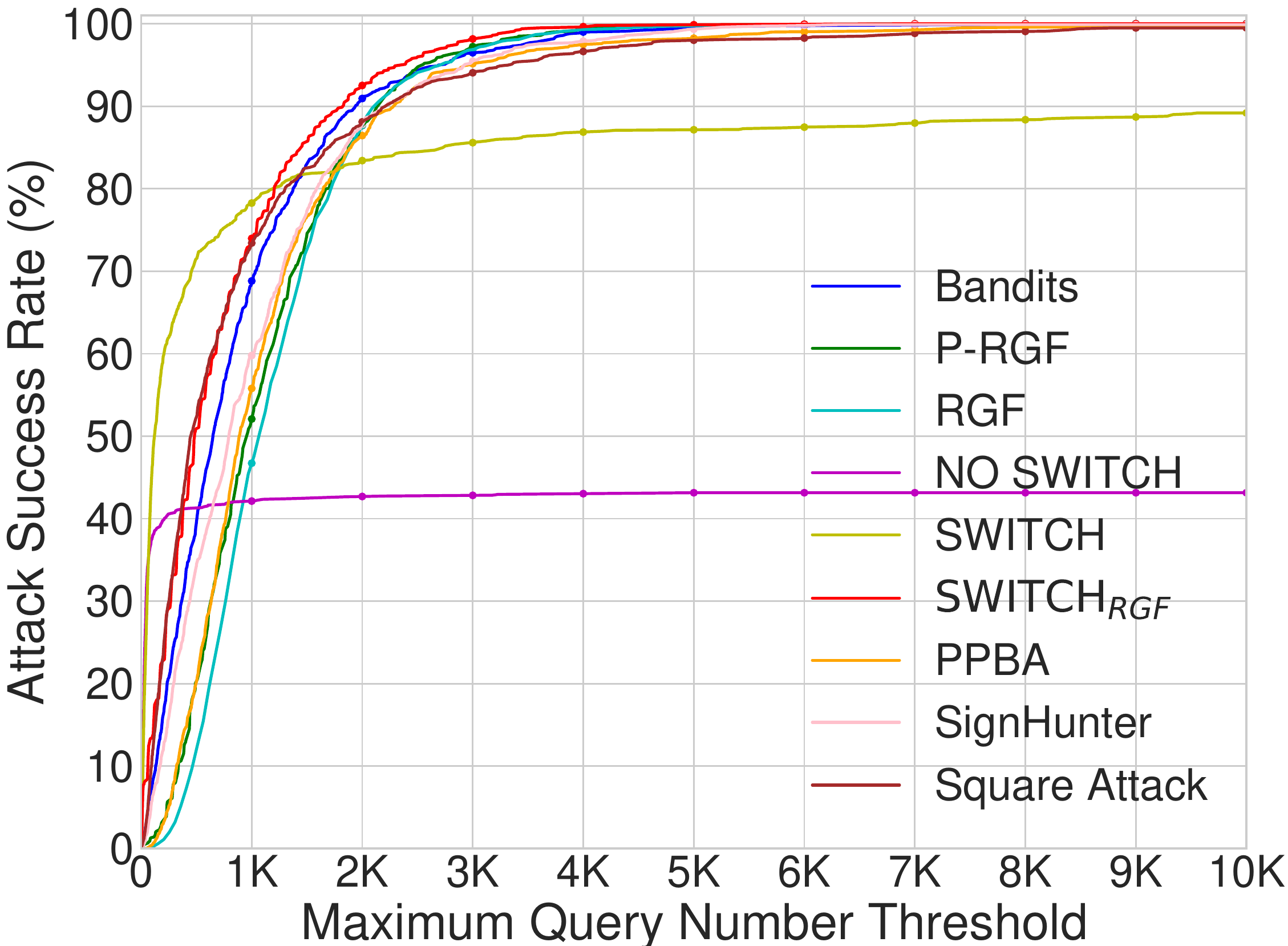}
			\subcaption{Targeted $\ell_2$ attack GDAS on CIFAR-10}
			\label{fig:targeted_l2_gdas_CIFAR-10}
		\end{minipage}
		\begin{minipage}[b]{.24\textwidth}
			\includegraphics[width=\linewidth]{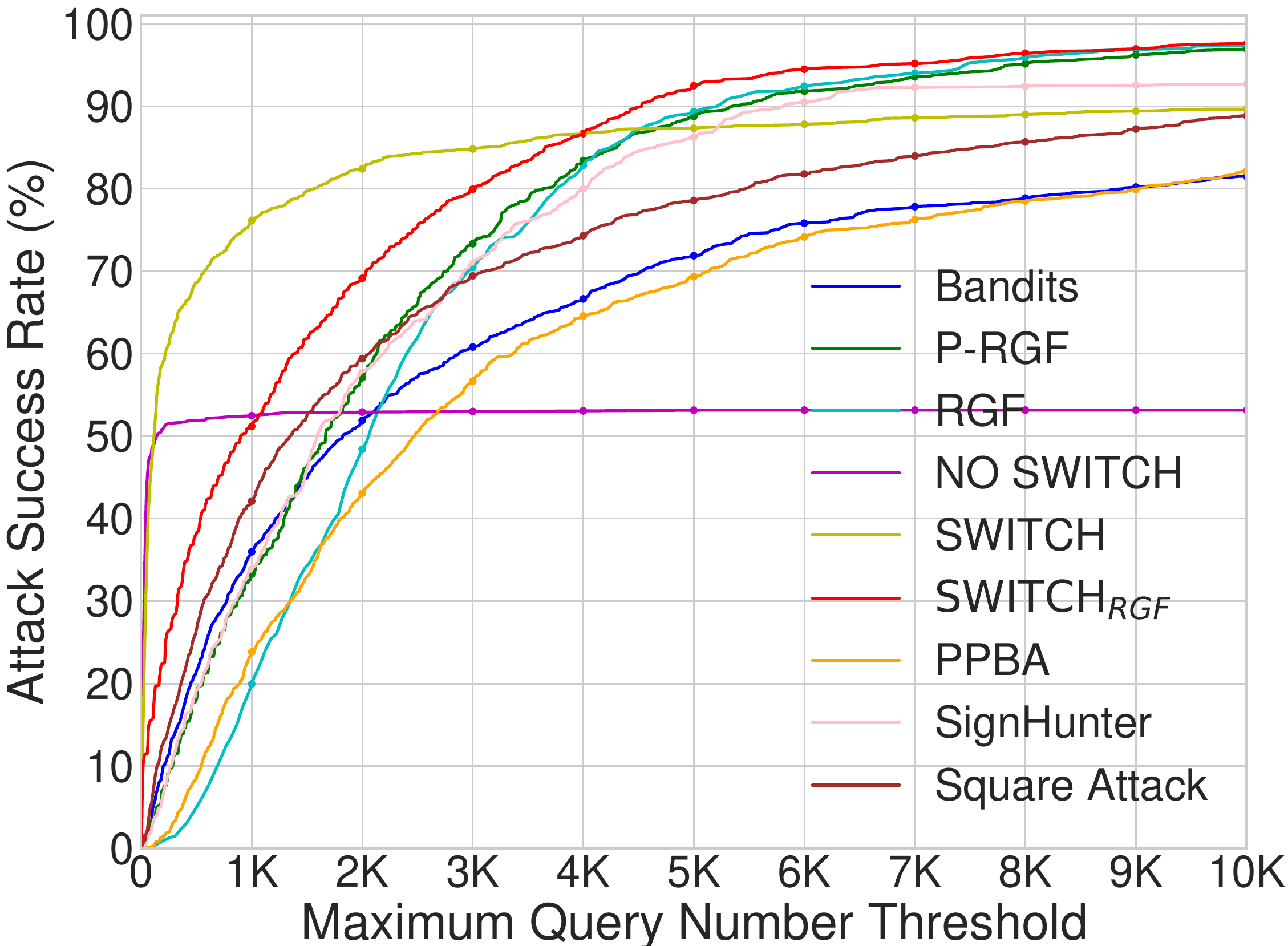}
			\subcaption{Targeted $\ell_2$ attack WRN-28 on CIFAR-10}
			\label{fig:targeted_l2_WRN-28_CIFAR-10}
		\end{minipage}
		\begin{minipage}[b]{.24\textwidth}
			\includegraphics[width=\linewidth]{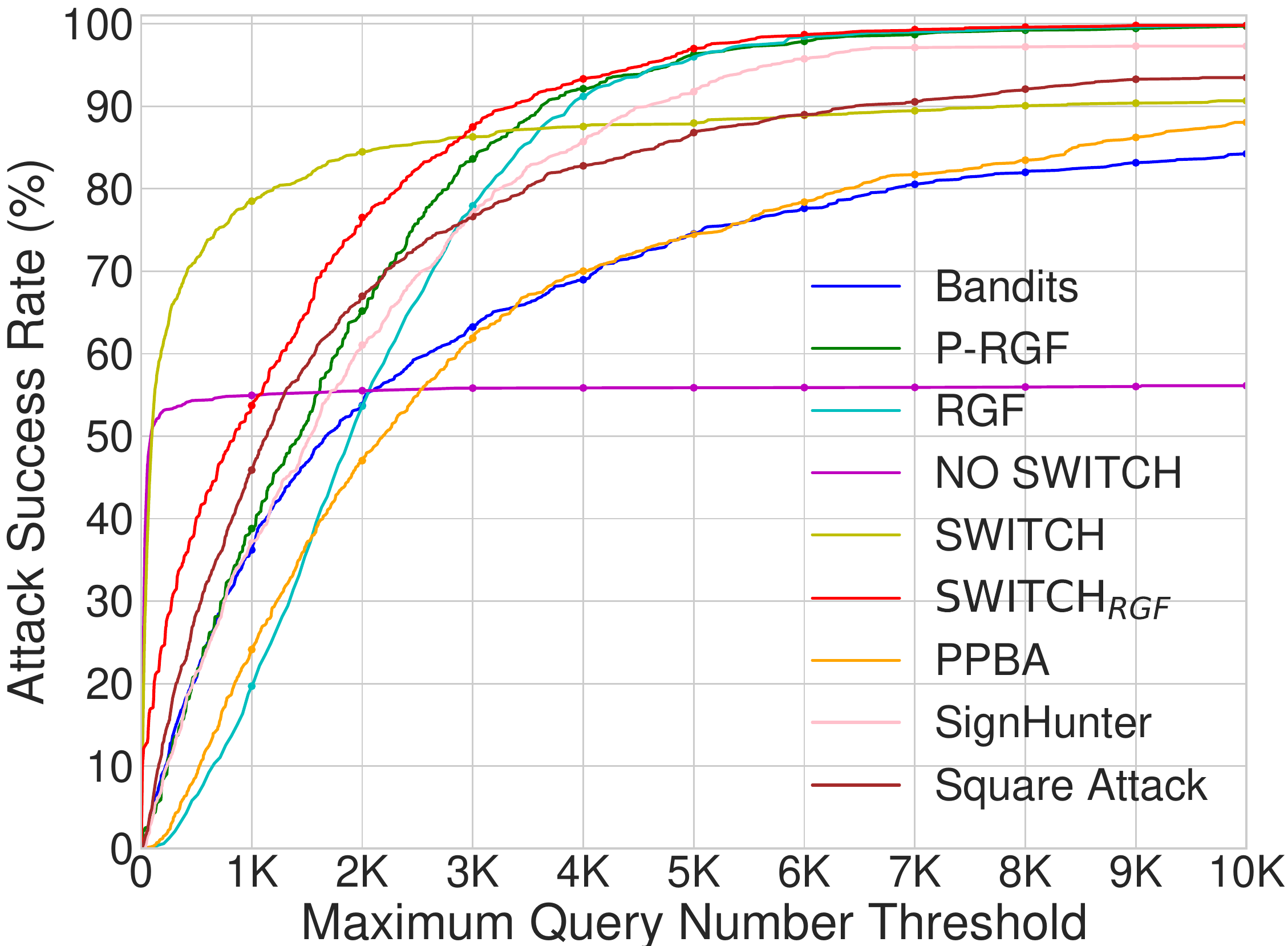}
			\subcaption{Targeted $\ell_2$ attack WRN-40 on CIFAR-10}
			\label{fig:targeted_l2_WRN-40_CIFAR-10}
		\end{minipage}
		\begin{minipage}[b]{.25\textwidth}
			\includegraphics[width=\linewidth]{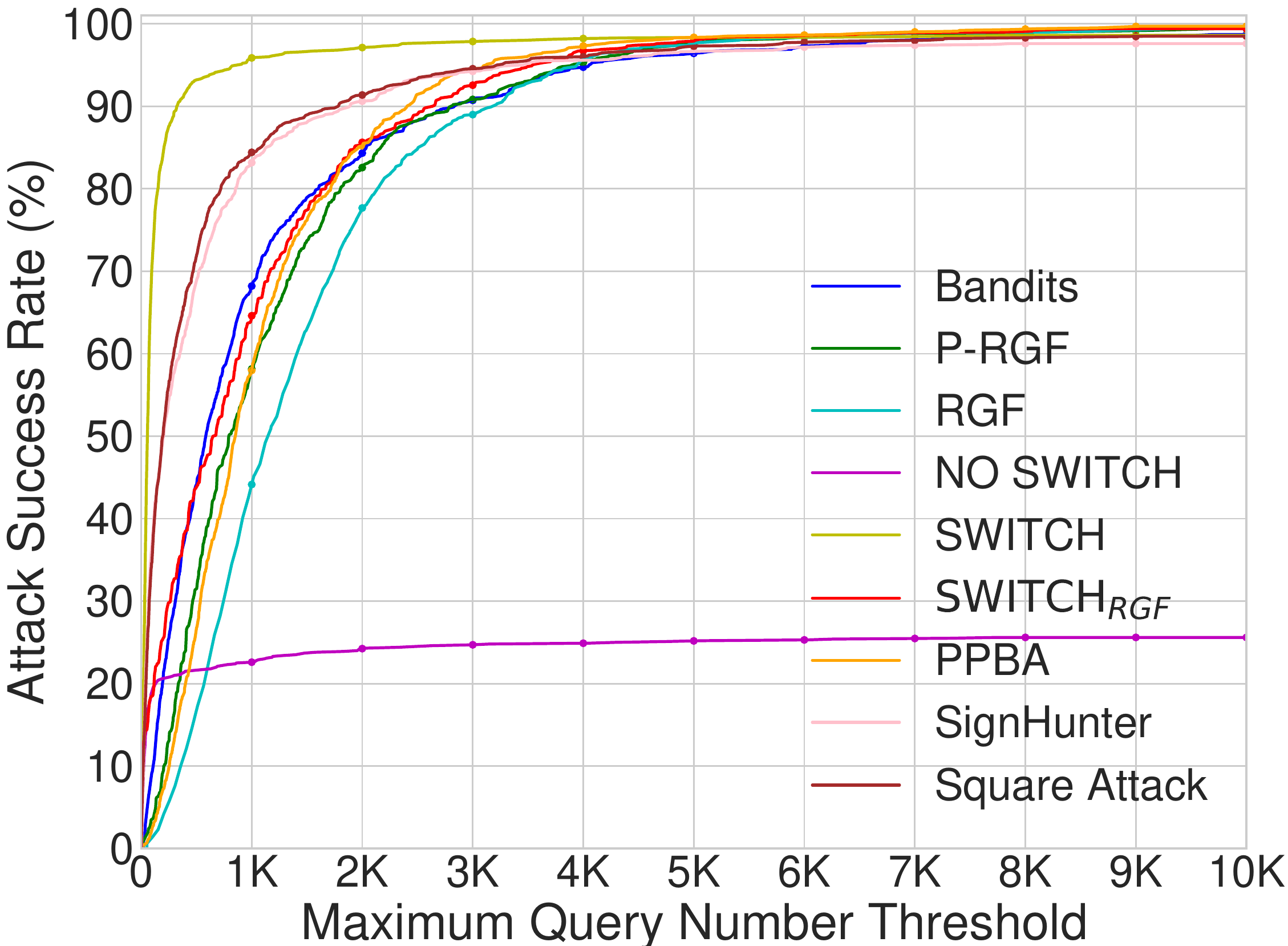}
			\subcaption{Untargeted $\ell_2$ attack D$_{121}$ on TinyImageNet}
			\label{fig:untargeted_l2_densenet121_TinyImageNet}
		\end{minipage}
		\begin{minipage}[b]{.25\textwidth}
			\includegraphics[width=\linewidth]{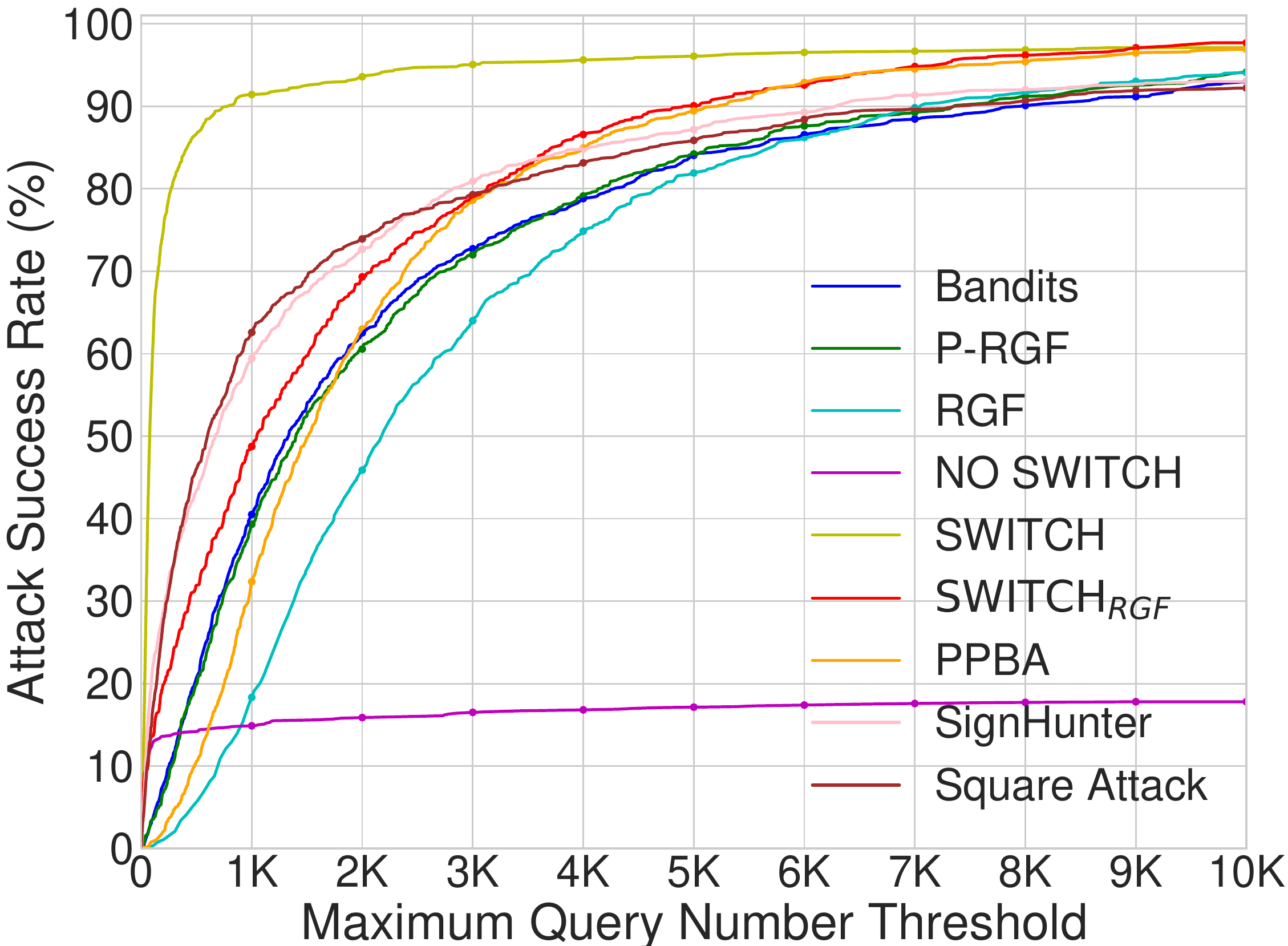}
			\subcaption{Untargeted $\ell_2$ attack R$_{32}$ on TinyImageNet}
			\label{fig:untargeted_l2_resnext101_32x4d_TinyImageNet}
		\end{minipage}
		\begin{minipage}[b]{.25\textwidth}
			\includegraphics[width=\linewidth]{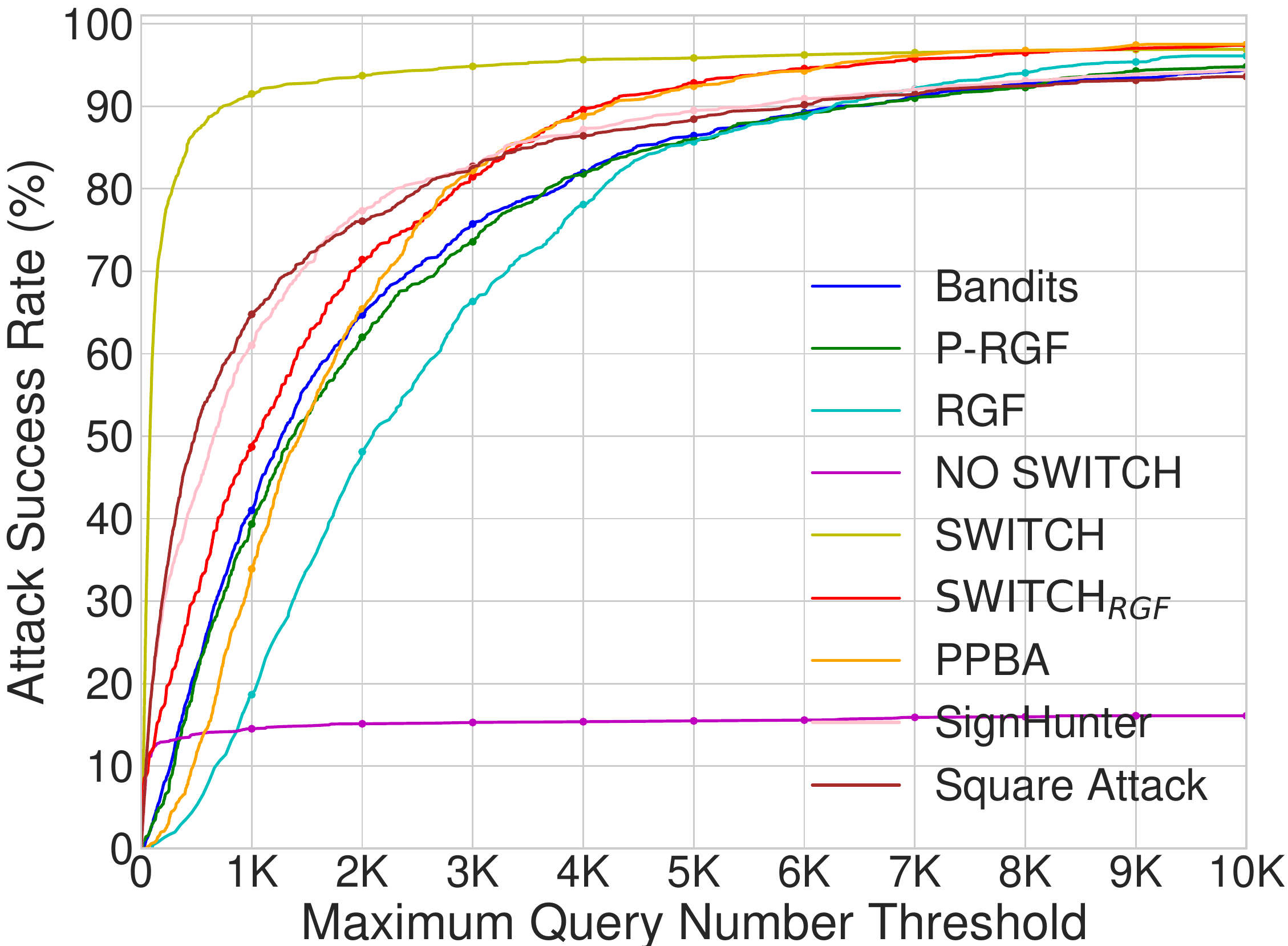}
			\subcaption{Untargeted $\ell_2$ attack R$_{64}$ on TinyImageNet}
			\label{fig:untargeted_l2_resnext101_64x4d_TinyImageNet}
		\end{minipage}
		\begin{minipage}[b]{.25\textwidth}
			\includegraphics[width=\linewidth]{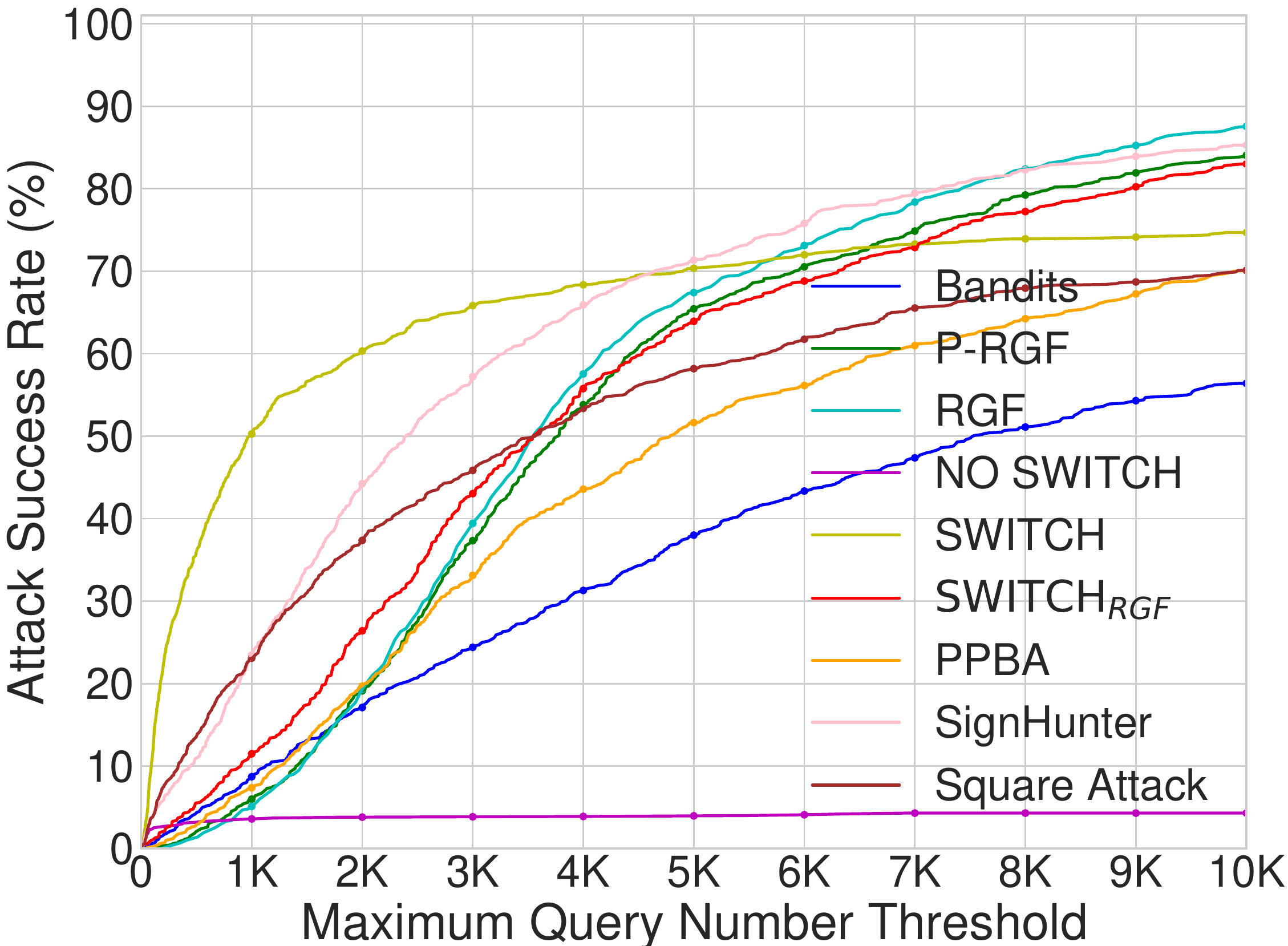}
			\subcaption{Targeted $\ell_2$ attack D$_{121}$ on TinyImageNet}
			\label{fig:targeted_l2_densenet121_TinyImageNet}
		\end{minipage}
		\begin{minipage}[b]{.25\textwidth}
			\includegraphics[width=\linewidth]{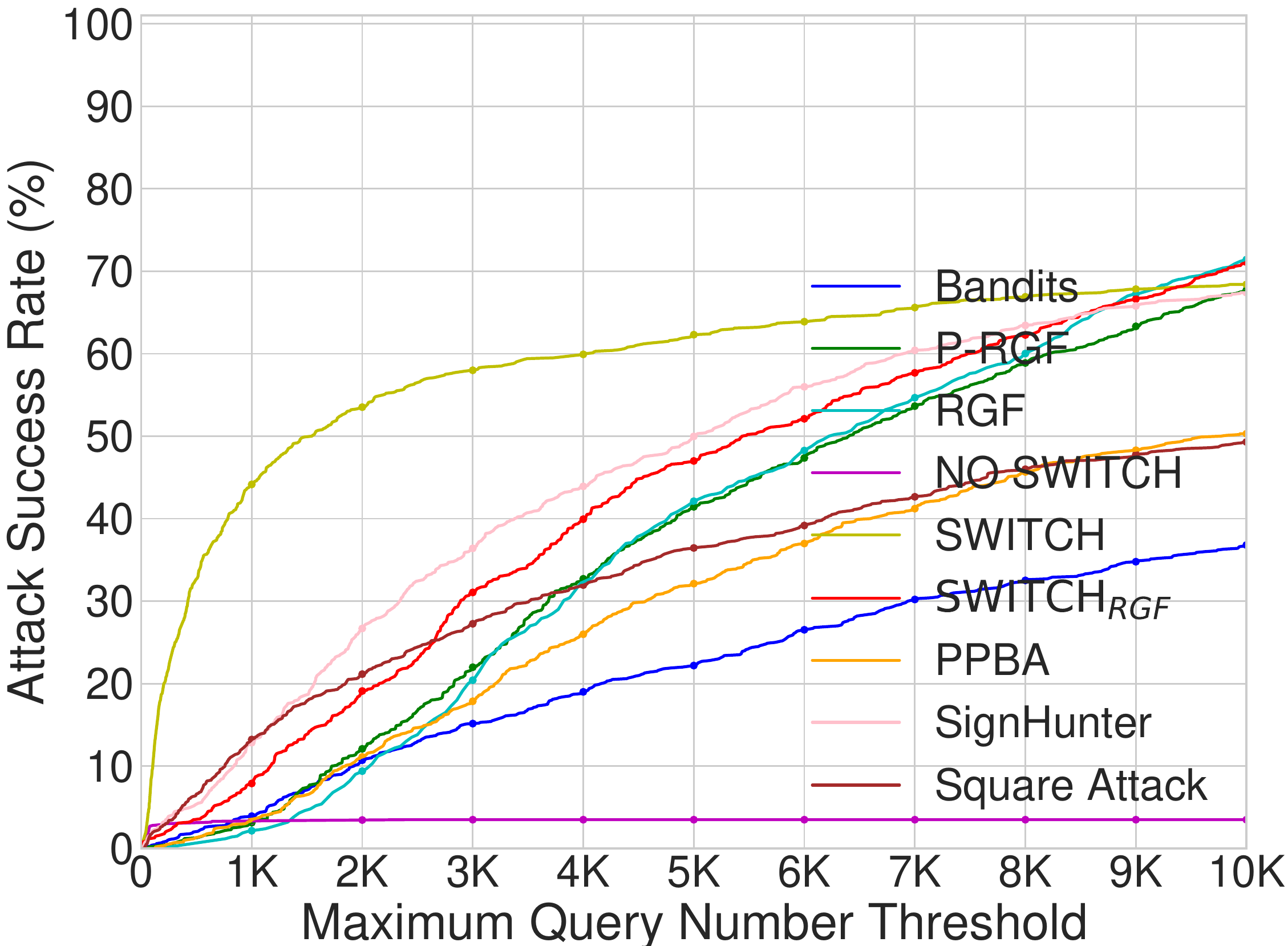}
			\subcaption{Targeted $\ell_2$ attack R$_{32}$ on TinyImageNet}
			\label{fig:targeted_l2_resnext101_32x4d_TinyImageNet}
		\end{minipage}
		\begin{minipage}[b]{.25\textwidth}
			\includegraphics[width=\linewidth]{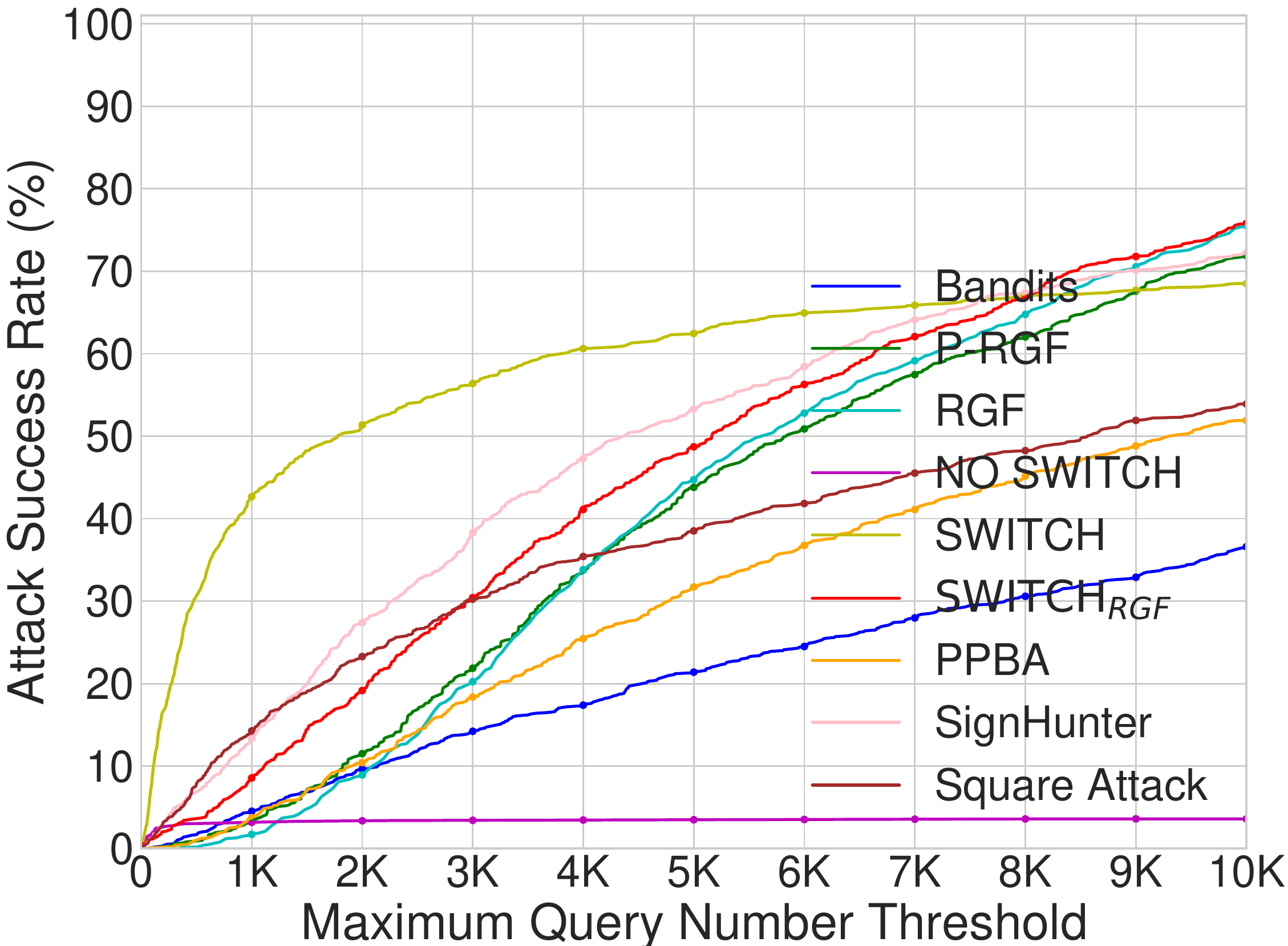}
			\subcaption{Targeted $\ell_2$ attack R$_{64}$ on TinyImageNet}
			\label{fig:targeted_l2_resnext101_64x4d_TinyImageNet}
		\end{minipage}
		
		\caption{Detailed comparisons of attack success rates at different maximum limited queries, where D$_{121}$ is DenseNet-121, R$_{32}$ is ResNeXt-101~(32x4d) and R$_{64}$ is ResNeXt-101~(64x4d).}
		\label{fig:query_threshold_attack_success_rate}
	\end{figure*}
	\noindent\textbf{Results of Attacks on Normal Models.} Tab.~\ref{tab:CIFAR_untargetd_result} and Tab.~\ref{tab:CIFAR_targetd_l2_result} show the experimental results of CIFAR-10 and CIFAR-100 datasets. Tab.~\ref{tab:TinyImageNet_untargeted_l2_result} and Tab.~\ref{tab:TinyImageNet_targeted_l2_result} show the results of TinyImageNet dataset. We can conclude the following:
	
	(1) In untargeted $\ell_2$ attacks, SWITCH usually achieves similar success rates to other state-of-the-art attacks with much fewer queries. In $\ell_\infty$ attacks and targeted attacks, the success rates of SWITCH are worse than those of untargeted ones. However, SWITCH features fast initial convergence, which manifests its advantage in those attacks under the setting of extremely limited queries. Meanwhile, SWITCH$_\text{RGF}$ makes up the shortcoming of SWITCH and raises the success rate to a high level. 
	
	(3) SWITCH$_\text{RGF}$ consumes significant fewer queries than RGF, and also than P-RGF even though they use the same surrogate model, because the amount of queries used to estimate gradients are saved in those iterations when the surrogate gradient is useful.
	
	
	\noindent\textbf{Results of Attacks on Defensive Models.} To inspect the performance of attacking defensive models, we conduct the experiments of attacking three defensive models, \textit{i.e.,} ComDefend (CD) \cite{jia2019comdefend}, Feature Distillation (FD) \cite{liu2019feature}, and JPEG \cite{guo2018countering}. CD is a defensive model that consists of a compression CNN and a reconstruction CNN to transform the adversarial image to defend against attacks. FD is a DNN-oriented compression defensive framework. JPEG is a image transformation-based defensive method. All the defensive models adopt the backbone of ResNet-50. Results of Tab.~\ref{tab:TinyImageNet_defensive_models_linf_result} conclude:
	
	(1) SWITCH achieves the best success rate in attacking CD among all attacks, and it also achieves the comparable success rate with state-of-the-art attacks in attacking FD.
	
	(2) SWITCH outperforms SWITCH$_\text{RGF}$ because gradient estimates obtained from RGF is almost useless in attacking these defensive models, possibly due to the phenomenon of obfuscated gradients \cite{obfuscated_gradients}. Readers can refer to success rates of RGF in Tab. \ref{tab:TinyImageNet_defensive_models_linf_result}.

	\noindent\textbf{Attack Success Rates at Different Maximum Queries.} To investigate the results more comprehensively, we present the attack success rates by limiting different maximum query numbers in Fig.~\ref{fig:query_threshold_attack_success_rate}. It shows that (1) \textcolor{greenyellow}{SWITCH} is rather useful for the attack scenario with an extremely low query limitation. In the difficult targeted attacks (row 2 and row 4), SWITCH exhibits the highest success rate under \nn{1000} limited queries. This is because of the skillful use of the exploring direction (surrogate gradient $\pm\overline{\hat{\bm{\mathrm{g}}}}$), which is more effective than those attacks that follow random directions (\textit{e.g.,} Square Attack), resulting in SWITCH's fast initial convergence. In real scenarios, the query budgets are often extremely limited, making SWITCH more useful in these situations. (2) \textcolor{greenyellow}{SWITCH} may converge to a lower success rate given a large query number threshold, which is a drawback of SWITCH. However, \textcolor{red}{SWITCH$_\text{RGF}$} avoids this problem while often remains the superior query efficiency. 
	\vspace{-1cm}
	\section{Conclusion}
	In this study, we propose a highly efficient black-box attack named SWITCH. It exploits the gradient of a surrogate model and then switches it if following such gradient cannot improve the value of attack objective function. We also develop an extension called SWITCH$_\text{RGF}$ to further improve the success rate. 
	Despite the simplicity of our proposed methods, the strategy is effective and explainable, and extensive experiments demonstrate that our proposed methods achieve state-of-the-art performance.

	\bibliographystyle{ACM-Reference-Format}
	\bibliography{main}
	
	\appendix
	\twocolumn[{%
		\centering
		\LARGE \bfseries Supplementary Material \\[1.5em]
		
	}]
	\section{Experimental Settings}
	\subsection{Hyperparameters of Compared Methods}
	\noindent \textbf{RGF and P-RGF attack.} Hyperparameters of RGF~\cite{nesterov2017random} and P-RGF~\cite{cheng2019improving} attacks are listed in Tab. \ref{tab:RGF_conf}. The experiments of RGF and P-RGF are conducted by using the PyTorch version translated from the official TensorFlow implementation. 
	\begin{table}[htbp]
		\small
		\tabcolsep=0.1cm
		
		\caption{The hyperparameters of RGF and P-RGF attacks.}
		\label{tab:RGF_conf}
		\begin{center}
			\scalebox{1}{
				\begin{tabular}{c|p{5cm}|c}
					\toprule
					Norm & Hyperparameter & Value \\
					\midrule
					\multirow{2}{*}{$\ell_2$} & $h$, image learning rate  & 0.1 \\
					& $\sigma$, sampling variance & 1e-4 \\
					\midrule
					\multirow{2}{*}{$\ell_\infty$} & $h$, image learning rate  & 0.005 \\
					& $\sigma$, sampling variance & 1e-4 \\
					\midrule
					$\ell_2$, $\ell_\infty$ & surrogate model used in CIFAR-10/100 & ResNet-110 \\
					$\ell_2$, $\ell_\infty$ & surrogate model used in TinyImageNet & ResNet-101 \\
					\bottomrule
			\end{tabular}}
		\end{center}
		
	\end{table}
	
	\noindent \textbf{Bandits.} Hyperparameters of Bandits \cite{ilyas2018prior} are listed in Tab. \ref{tab:bandits_conf}. The OCO learning rate is used to update the prior, where the prior is an alias of gradient $\bm{\mathrm{g}}$ for updating the input image.
	
	\noindent \textbf{Parsimonious.} Hyperparameters of Parsimonious \cite{moonICML19} are listed in Tab. \ref{tab:parsimonious_conf}. Parsimonious only supports $\ell_\infty$ norm attack. We follow the official implementation to set the initial block size to 4 in all experiments.
	
	\noindent \textbf{SignHunter.} Hyperparameters of SignHunter \cite{al2020sign} are listed in Tab. \ref{tab:SignHunter_conf}. 
	
	\noindent \textbf{Square Attack.} Hyperparameters of Square Attack \cite{ACFH2020square} are shown in Tab. \ref{tab:SquareAttack_conf}. Square Attack randomly samples a square window filled with perturbations in each iteration. The area of this square window in the first iteration is initialized as $ H \times W \times p$, which is then reduced in subsequent iterations. 
	
	\begin{table}[htbp]
		\small
		\tabcolsep=0.1cm
		\caption{The hyperparameters of Bandits.}
		\label{tab:bandits_conf}
		\begin{center}
			\scalebox{1}{
				\begin{tabular}{c|p{5cm}|c}
					\toprule
					Norm & Hyperparameter & Value \\
					\midrule
					\multirow{5}{*}{$\ell_2$} & $\delta$, finite difference probe & 0.01 \\
					& $\eta$, image learning rate  & 0.1 \\
					& $\eta_g$, OCO learning rate & 0.1 \\
					& $\tau$,  Bandits exploration & 0.3 \\
					& maximum query times & \nn{10000} \\
					\midrule
					\multirow{5}{*}{$\ell_\infty$} & $\delta$, finite difference probe & 0.1 \\
					& $\eta$, image learning rate  & 1/255 \\
					& $\eta_g$, OCO learning rate & 1.0 \\
					& $\tau$,  Bandits exploration & 0.3 \\
					& maximum query times & \nn{10000} \\
					\bottomrule
			\end{tabular}}
		\end{center}
		
	\end{table}

	\begin{table}[htbp]
		\small
		\tabcolsep=0.1cm
		\caption{The hyperparameters of Parsimonious.}
		\label{tab:parsimonious_conf}
		\begin{center}
			\scalebox{1}{
				\begin{tabular}{p{6cm}|c}
					\toprule
					Hyperparameter & Value \\
					\midrule
					the number of iterations in local search  & 1 \\
					$k$, initial block size & 4 \\
					batch size & 64 \\
					no hierarchical evaluation & False\\
					
					\bottomrule
			\end{tabular}}
		\end{center}
		
	\end{table}
	\begin{table}[htbp]
		\small
		\tabcolsep=0.1cm
		\caption{The hyperparameters of SignHunter.}
		\label{tab:SignHunter_conf}
		\begin{center}
			\scalebox{1}{
				\begin{tabular}{p{6cm}|c}
					\toprule
					Hyperparameter & Value \\
					\midrule
					$\epsilon$, radius of $\ell_2$ norm ball in CIFAR-10/100 & 1.0 \\
					$\epsilon$, radius of $\ell_2$ norm ball in TinyImageNet & 2.0 \\
					$\epsilon$, maximum radius of $\ell_\infty$ norm ball & 8/255 \\
					\bottomrule
			\end{tabular}}
		\end{center}
		
	\end{table}
	
	\begin{table}[htbp]
		\small
		\tabcolsep=0.1cm
		\caption{The hyperparameters of Square Attack.}
		\label{tab:SquareAttack_conf}
		\begin{center}
			\scalebox{1}{
				\begin{tabular}{p{6cm}|c}
					\toprule
					Hyperparameter & Value \\
					\midrule
					$p$, initial probability of changing a coordinate & 0.05 \\
					\bottomrule
			\end{tabular}}
		\end{center}
		
	\end{table}
	
	\begin{table}[htbp]
		\small
		\tabcolsep=0.1cm
		\caption{The hyperparameters of PPBA.}
		\label{tab:PPBA_conf}
		\begin{center}
			\scalebox{1}{
				\begin{tabular}{p{1.9cm}|p{5cm}|c}
					\toprule
					Dataset & Hyperparameter & Value \\
					\midrule
					\multirow{9}{2cm}{CIFAR-10/100} & order & strided \\
					& $\rho$, the change of measurement vector $z$ for $\ell_\infty$ norm attack & 0.001 \\
					& $\rho$, the change of measurement vector $z$ for $\ell_2$ norm attack & 0.01 \\
					& $\mu$, the momentum of calculating effective probability & 1 \\
					& frequency dim, used in initialize blocks & 11 \\
					& low-frequency dimension & \nn{1500} \\
					& stride, used in initialize blocks & 7 \\
					& number of samples per iteration & 1 \\
					\midrule
					\multirow{9}{2cm}{TinyImageNet} & order & strided \\
					& $\rho$, the change of measurement vector $z$ for $\ell_\infty$ norm attack & 0.001 \\
					& $\rho$, the change of measurement vector $z$ for $\ell_2$ norm attack & 0.01 \\
					& $\mu$, the momentum of calculating effective probability & 1 \\
					& frequency dim, used in initialize block & 15 \\
					& low-frequency dimension & \nn{1500} \\
					& stride, used in initialize block & 7 \\
					& number of samples per iteration & 1 \\
					\bottomrule
			\end{tabular}}
		\end{center}
		
	\end{table}
	\begin{table}[t]
		\small
		\tabcolsep=0.1cm
		\caption{The hyperparameters of SWITCH attack.}
		\label{tab:SWITCH_conf}
		\begin{center}
			\scalebox{1}{
				\begin{tabular}{c|p{4.8cm}|c}
					\toprule
					Norm & Hyperparameter & Value \\
					\midrule
					\multirow{3}{*}{$\ell_2$} & $\eta$, image learning rate for CIFAR-10 \& CIFAR-100 & 0.1 \\
					& $\eta$, image learning rate for TinyImageNet & 0.2 \\
					\midrule
					\multirow{6}{*}{$\ell_\infty$} & $\eta$, image learning rate for untargeted attack in CIFAR-10 \& CIFAR-100 & 0.01 \\
					& $\eta$, image learning rate for targeted attack in CIFAR-10 \& CIFAR-100 & 0.003 \\
					& $\eta$, image learning rate in TinyImageNet dataset & 0.003 \\
					\midrule
					\multirow{2}{*}{$\ell_2$, $\ell_\infty$} & $m$, surrogate model used in CIFAR-10 \& CIFAR-100 dataset & ResNet-110 \\
					\multirow{2}{*}{$\ell_2$, $\ell_\infty$} & $m$, surrogate model used in TinyImageNet dataset & ResNet-101 \\
					\bottomrule
			\end{tabular}}
		\end{center}
		
	\end{table}
	
	\noindent \textbf{PPBA.} Hyperparameters of Projection \& Probability-driven Black-box Attack (PPBA) \cite{li2020projection} are shown in Tab.~\ref{tab:PPBA_conf}. We follow the official implementation to initialize the random matrix (strided) with the dimension of \nn{1500}.

	\noindent \textbf{SWITCH.} Hyperparameters of the proposed SWITCH are listed in Tab. \ref{tab:SWITCH_conf}. The configurations of the surrogate model used in SWITCH is the pre-trained ResNet-110 in CIFAR-10 and CIFAR-100 datasets, and ResNet-101 in TinyImageNet dataset.

	\section{Experimental Results}

	\noindent\textbf{Experimental Results Including Failed Examples.} In our original paper, we report the experimental results (attack success rate and the average/median number of
	queries) over successful attacks, the samples of failed attacks are excluded in the statistics. To present the results more comprehensively, we also report the average and median query over all the samples by setting the queries of failed samples to \nn{10000} in this supplementary material. The experimental results are shown in Tab. \ref{tab:TinyImageNet_untargeted_l2_result_include_failed}, Tab. \ref{tab:TinyImageNet_untargeted_linf_result_include_failed}, Tab. \ref{tab:TinyImageNet_targeted_l2_result_include_failed}, Tab. \ref{tab:CIFAR_untargetd_result_include_failed}, and Tab. \ref{tab:CIFAR_targetd_l2_result_include_failed}. We can draw the following conclusions based on the results of these tables.
	
	(1) The number of the average and median query depends on the attack success rate. If the success rate is low such that the queries of many samples are set to \nn{10000}, then the average query is low. If the success rate is below 50\%, then the median query is \nn{10000}.
	
	(2) Compared to the results of the average and median query in the original paper, the average and the median query of SWITCH is lower than NO SWITCH because its attack success rate is higher than NO SWITCH.
	
\begin{figure*}[t]
		\setlength{\abovecaptionskip}{0pt}
		\setlength{\belowcaptionskip}{0pt}
		\captionsetup[sub]{font={scriptsize}}
		\centering
		\begin{minipage}[b]{.23\textwidth}
			\includegraphics[width=\linewidth]{figures/query_threshold_success_rate/CIFAR-10_pyramidnet272_l2_untargeted_attack_query_threshold_success_rate_dict.pdf}
			\subcaption{untargeted $\ell_2$ attack PyramidNet-272}
		\end{minipage}
		\begin{minipage}[b]{.23\textwidth}
			\includegraphics[width=\linewidth]{figures/query_threshold_success_rate/CIFAR-10_gdas_l2_untargeted_attack_query_threshold_success_rate_dict.pdf}
			\subcaption{untargeted $\ell_2$ attack GDAS}
		\end{minipage}
		\begin{minipage}[b]{.23\textwidth}
			\includegraphics[width=\linewidth]{figures/query_threshold_success_rate/CIFAR-10_WRN-28-10-drop_l2_untargeted_attack_query_threshold_success_rate_dict.pdf}
			\subcaption{untargeted $\ell_2$ attack WRN-28}
		\end{minipage}
		\begin{minipage}[b]{.23\textwidth}
			\includegraphics[width=\linewidth]{figures/query_threshold_success_rate/CIFAR-10_WRN-40-10-drop_l2_untargeted_attack_query_threshold_success_rate_dict.pdf}
			\subcaption{untargeted $\ell_2$ attack WRN-40}
		\end{minipage}
		\begin{minipage}[b]{.23\textwidth}
			\includegraphics[width=\linewidth]{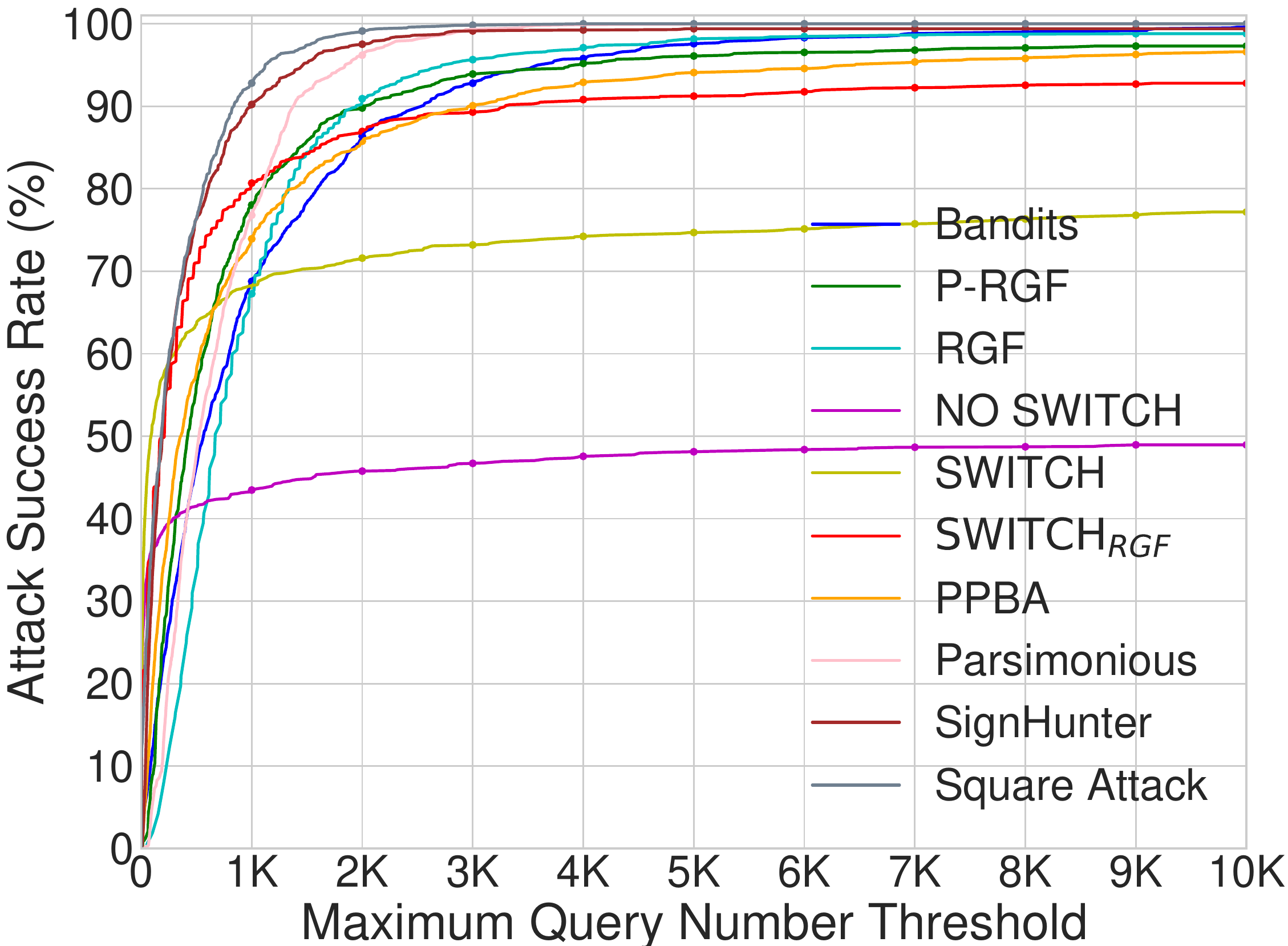}
			\subcaption{untargeted $\ell_\infty$ attack PyramidNet-272}
		\end{minipage}
		\begin{minipage}[b]{.23\textwidth}
			\includegraphics[width=\linewidth]{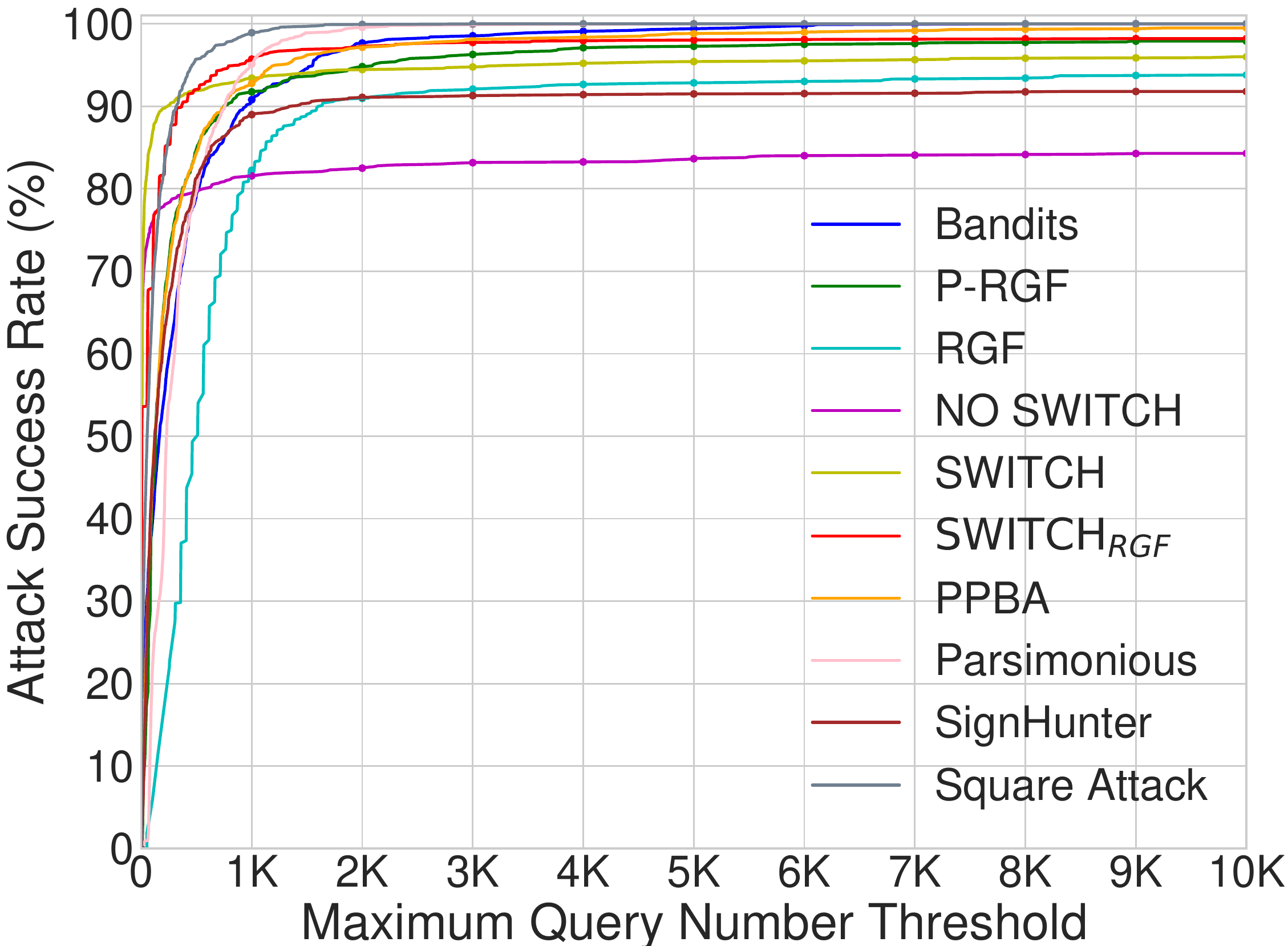}
			\subcaption{untargeted $\ell_\infty$ attack GDAS}
		\end{minipage}
		\begin{minipage}[b]{.23\textwidth}
			\includegraphics[width=\linewidth]{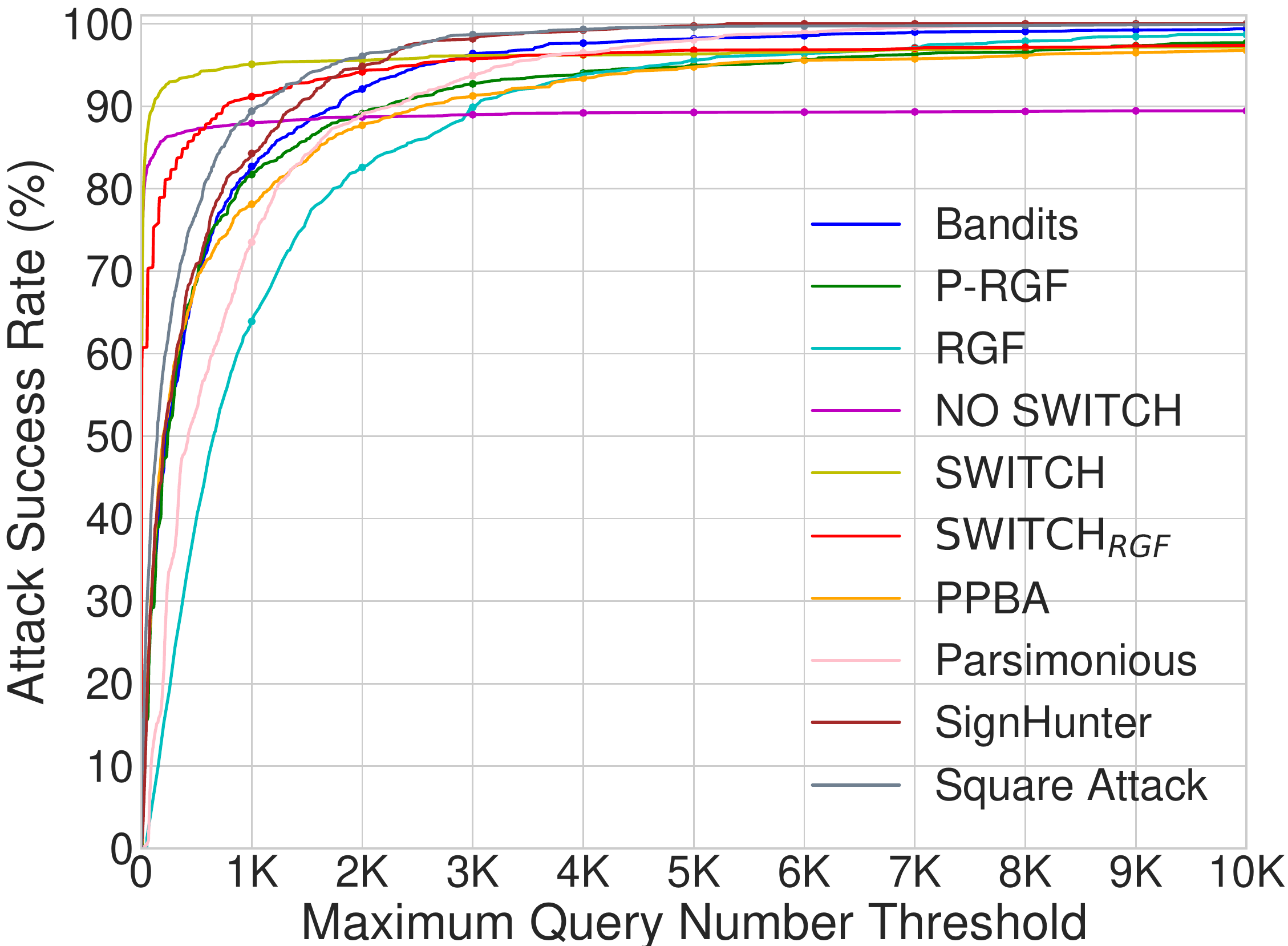}
			\subcaption{untargeted $\ell_\infty$ attack WRN-28}
		\end{minipage}
		\begin{minipage}[b]{.23\textwidth}
			\includegraphics[width=\linewidth]{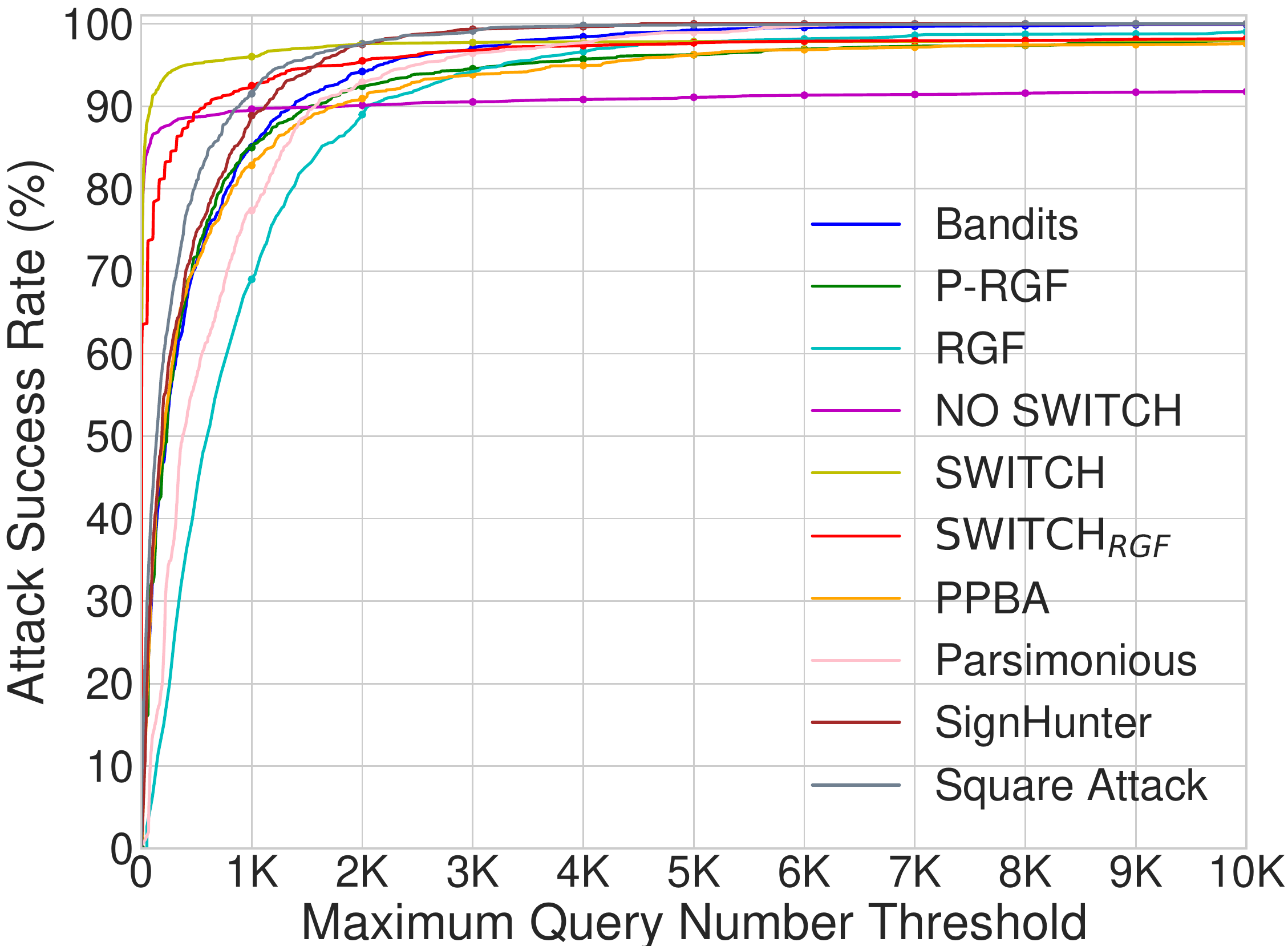}
			\subcaption{untargeted $\ell_\infty$ attack WRN-40}
		\end{minipage}
		\begin{minipage}[b]{.23\textwidth}
			\includegraphics[width=\linewidth]{figures/query_threshold_success_rate/CIFAR-10_pyramidnet272_l2_targeted_attack_query_threshold_success_rate_dict.pdf}
			\subcaption{targeted $\ell_2$ attack PyramidNet-272}
		\end{minipage}
		\begin{minipage}[b]{.23\textwidth}
			\includegraphics[width=\linewidth]{figures/query_threshold_success_rate/CIFAR-10_gdas_l2_targeted_attack_query_threshold_success_rate_dict.pdf}
			\subcaption{targeted $\ell_2$ attack GDAS}
		\end{minipage}
		\begin{minipage}[b]{.23\textwidth}
			\includegraphics[width=\linewidth]{figures/query_threshold_success_rate/CIFAR-10_WRN-28-10-drop_l2_targeted_attack_query_threshold_success_rate_dict.pdf}
			\subcaption{targeted $\ell_2$ attack WRN-28}
		\end{minipage}
		\begin{minipage}[b]{.23\textwidth}
			\includegraphics[width=\linewidth]{figures/query_threshold_success_rate/CIFAR-10_WRN-40-10-drop_l2_untargeted_attack_query_threshold_success_rate_dict.pdf}
			\subcaption{targeted $\ell_2$ attack WRN-40}
		\end{minipage}
		\caption{Comparison of attack success rate at different limited maximum queries on CIFAR-10 dataset.}
		\label{fig:query_to_attack_success_rate_CIFAR-10}
	\end{figure*}

\begin{figure*}[htbp]
		\setlength{\abovecaptionskip}{0pt}
		\setlength{\belowcaptionskip}{0pt}
		\captionsetup[sub]{font={scriptsize}}
		\centering
		\begin{minipage}[b]{.23\textwidth}
			\includegraphics[width=\linewidth]{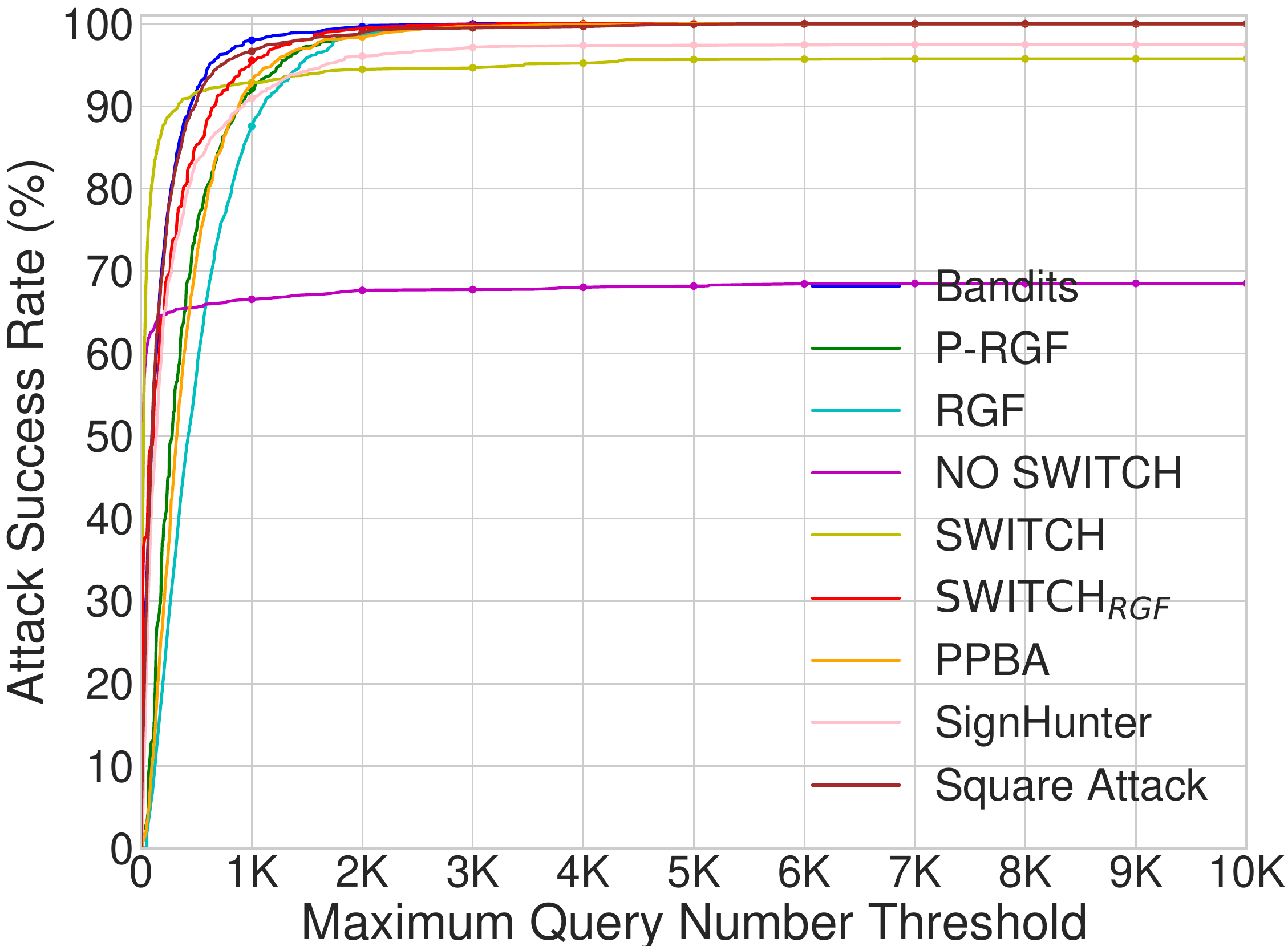}
			\subcaption{untargeted $\ell_2$ attack PyramidNet-272}
		\end{minipage}
		\begin{minipage}[b]{.23\textwidth}
			\includegraphics[width=\linewidth]{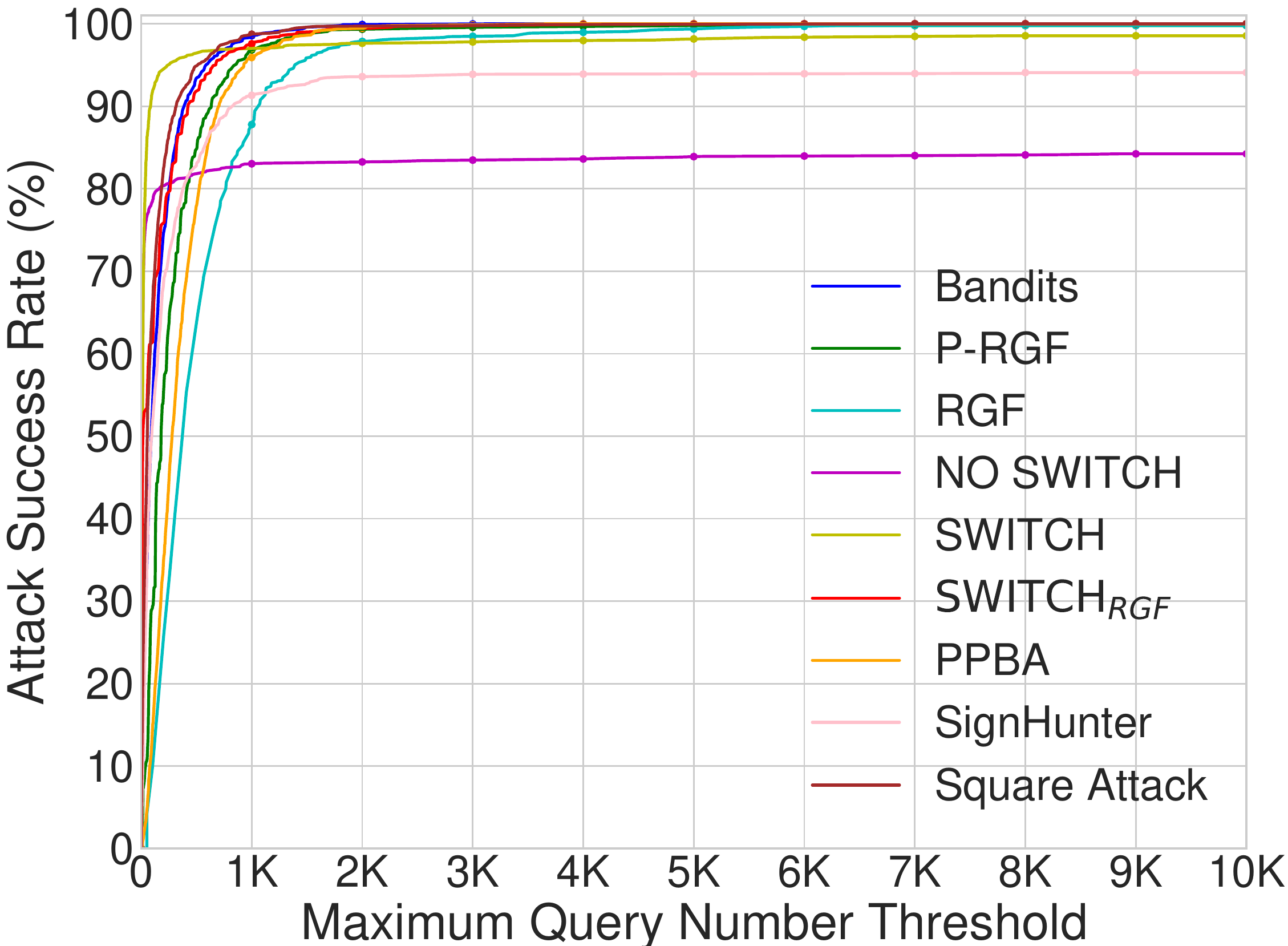}
			\subcaption{untargeted $\ell_2$ attack GDAS}
		\end{minipage}
		\begin{minipage}[b]{.23\textwidth}
			\includegraphics[width=\linewidth]{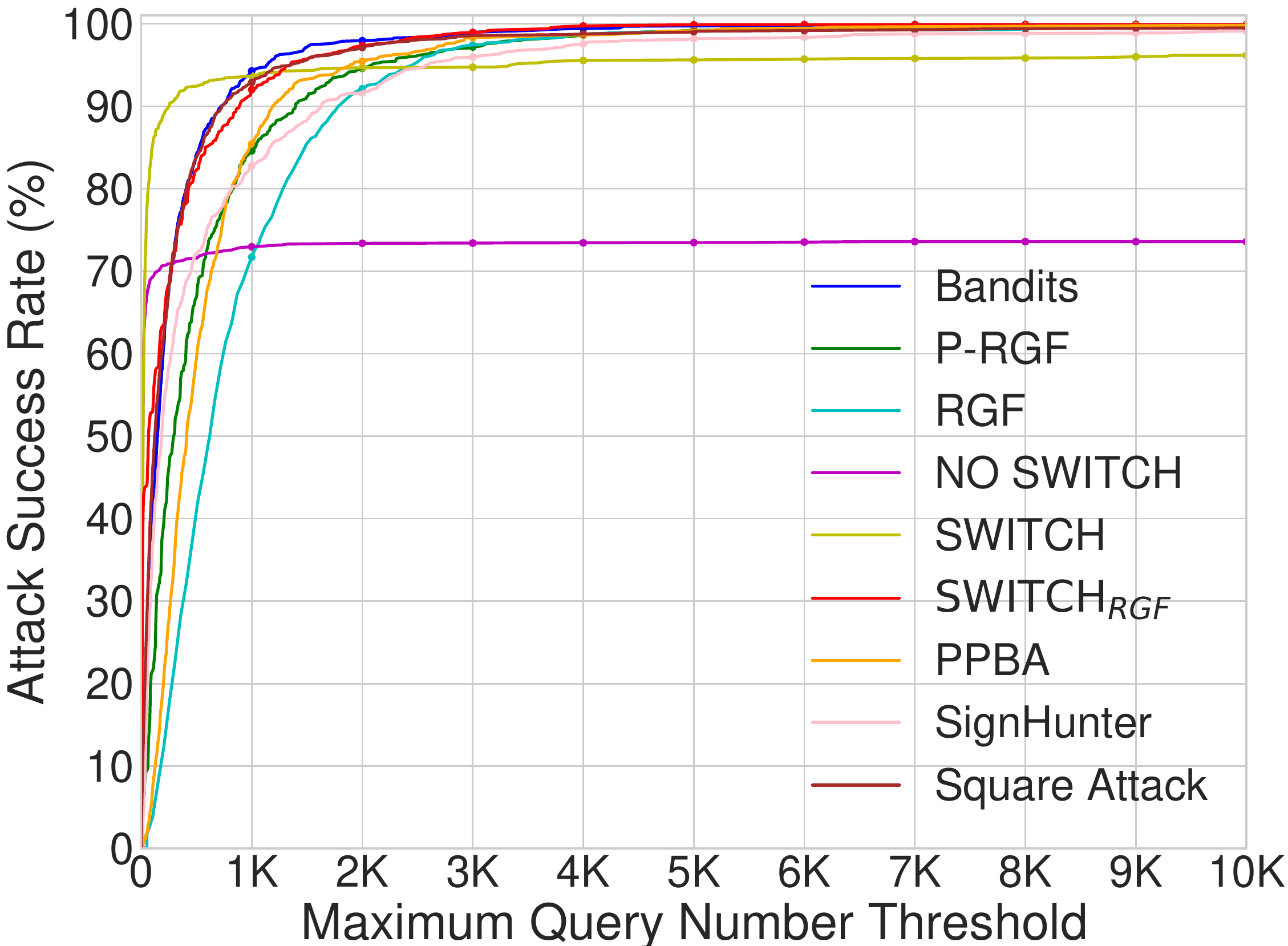}
			\subcaption{untargeted $\ell_2$ attack WRN-28}
		\end{minipage}
		\begin{minipage}[b]{.23\textwidth}
			\includegraphics[width=\linewidth]{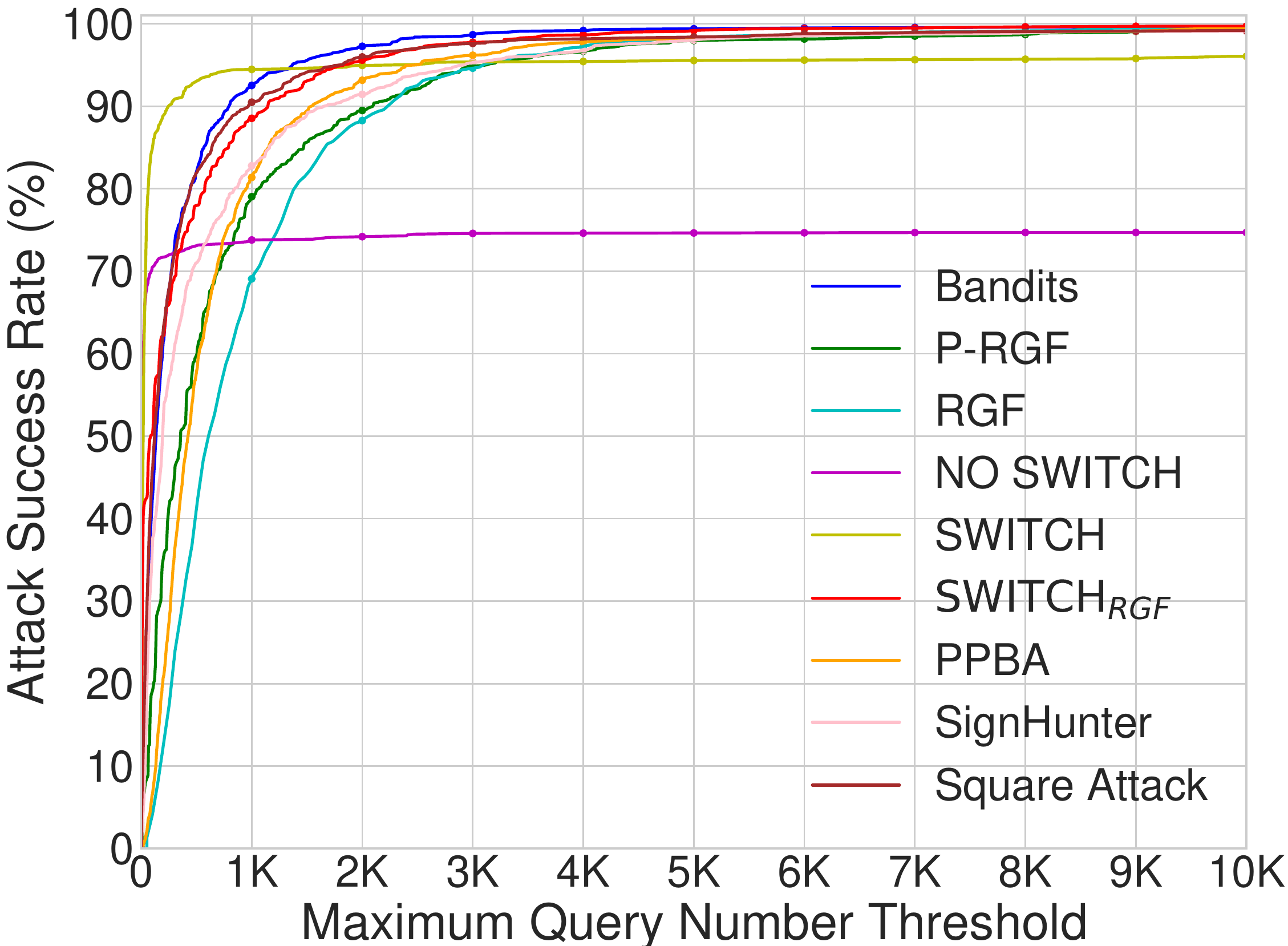}
			\subcaption{untargeted $\ell_2$ attack WRN-40}
		\end{minipage}
		\begin{minipage}[b]{.23\textwidth}
			\includegraphics[width=\linewidth]{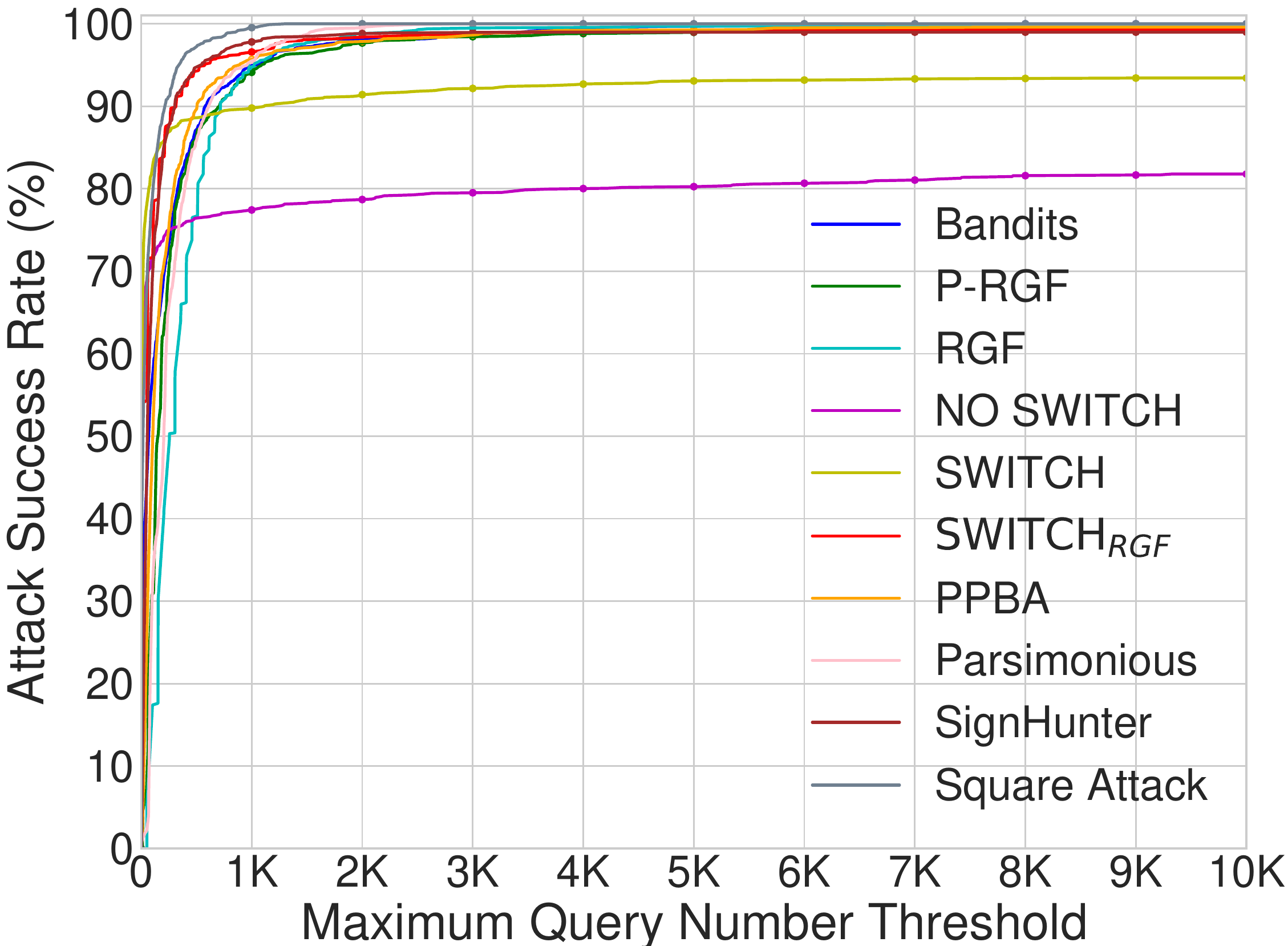}
			\subcaption{untargeted $\ell_\infty$ attack PyramidNet-272}
		\end{minipage}
		\begin{minipage}[b]{.23\textwidth}
			\includegraphics[width=\linewidth]{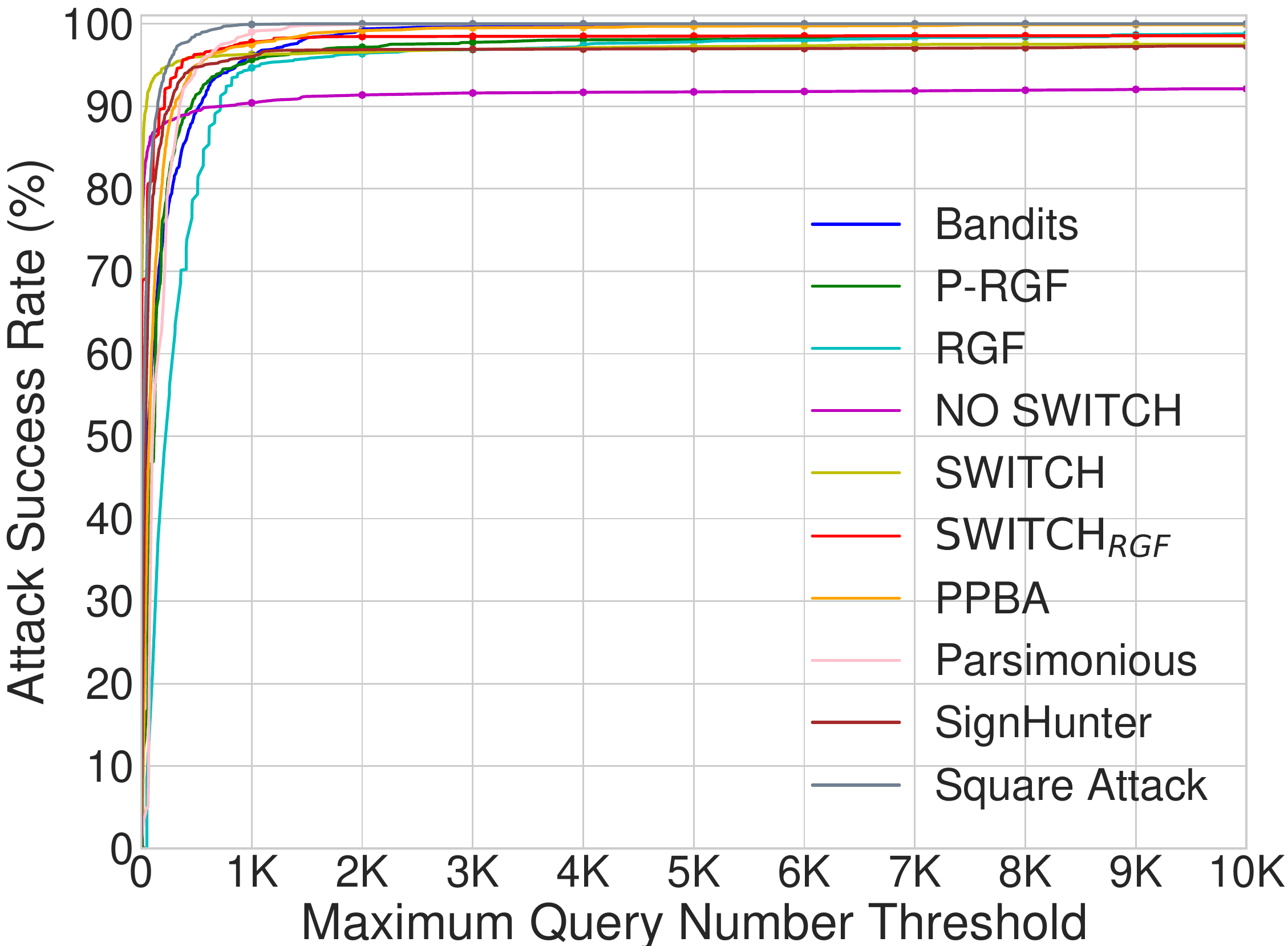}
			\subcaption{untargeted $\ell_\infty$ attack GDAS}
		\end{minipage}
		\begin{minipage}[b]{.23\textwidth}
			\includegraphics[width=\linewidth]{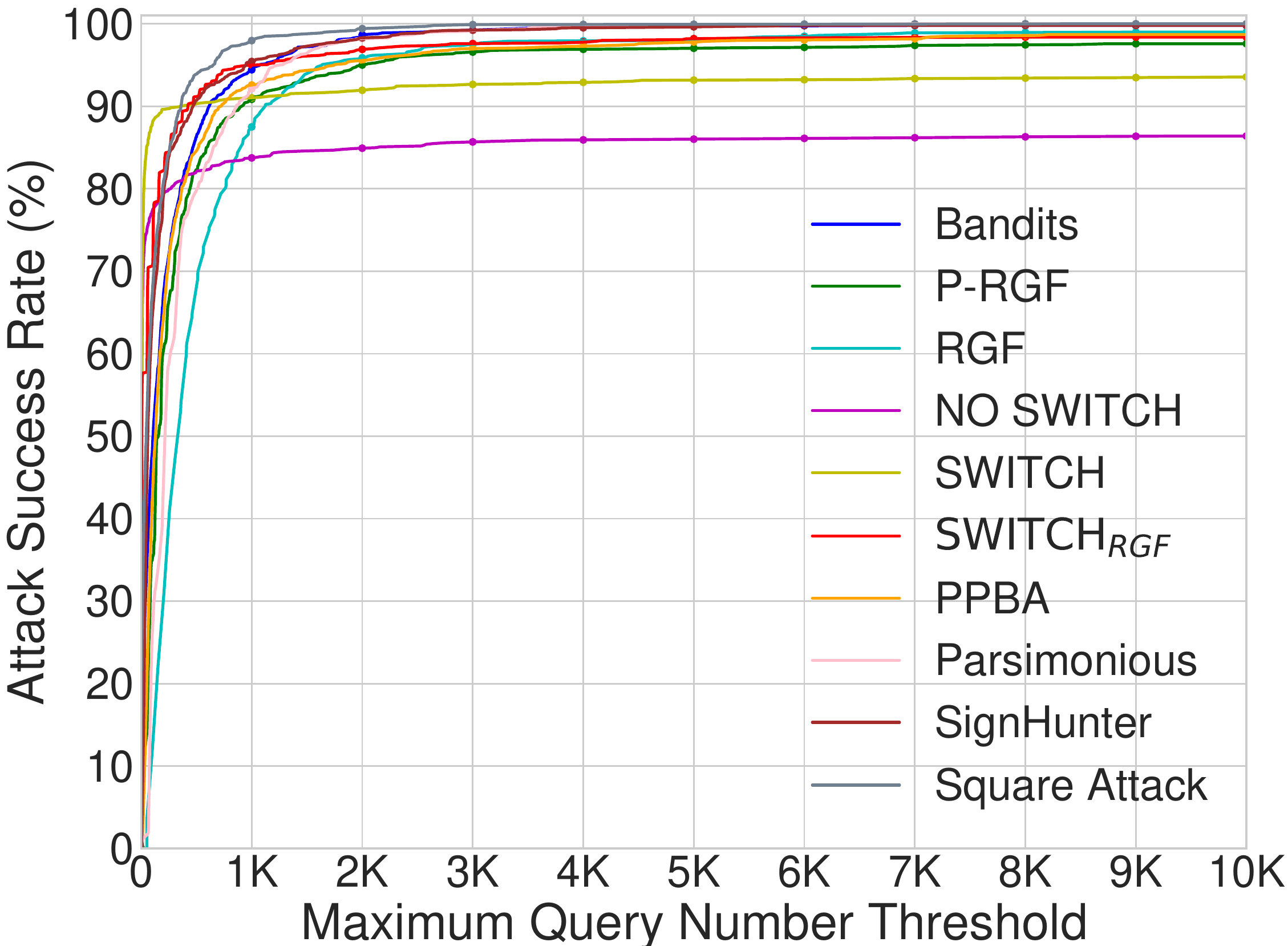}
			\subcaption{untargeted $\ell_\infty$ attack WRN-28}
		\end{minipage}
		\begin{minipage}[b]{.23\textwidth}
			\includegraphics[width=\linewidth]{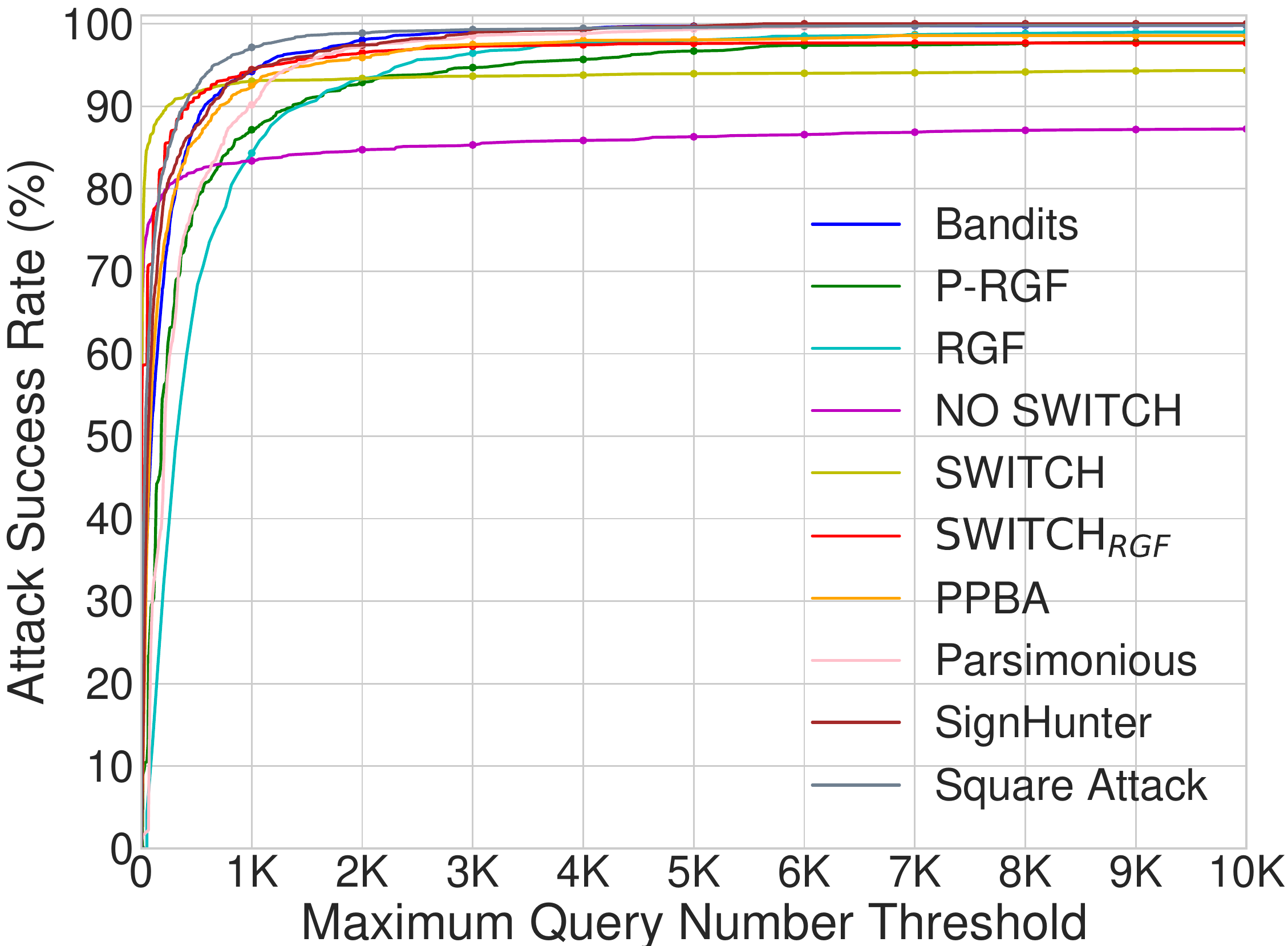}
			\subcaption{untargeted $\ell_\infty$ attack WRN-40}
		\end{minipage}
		\begin{minipage}[b]{.23\textwidth}
			\includegraphics[width=\linewidth]{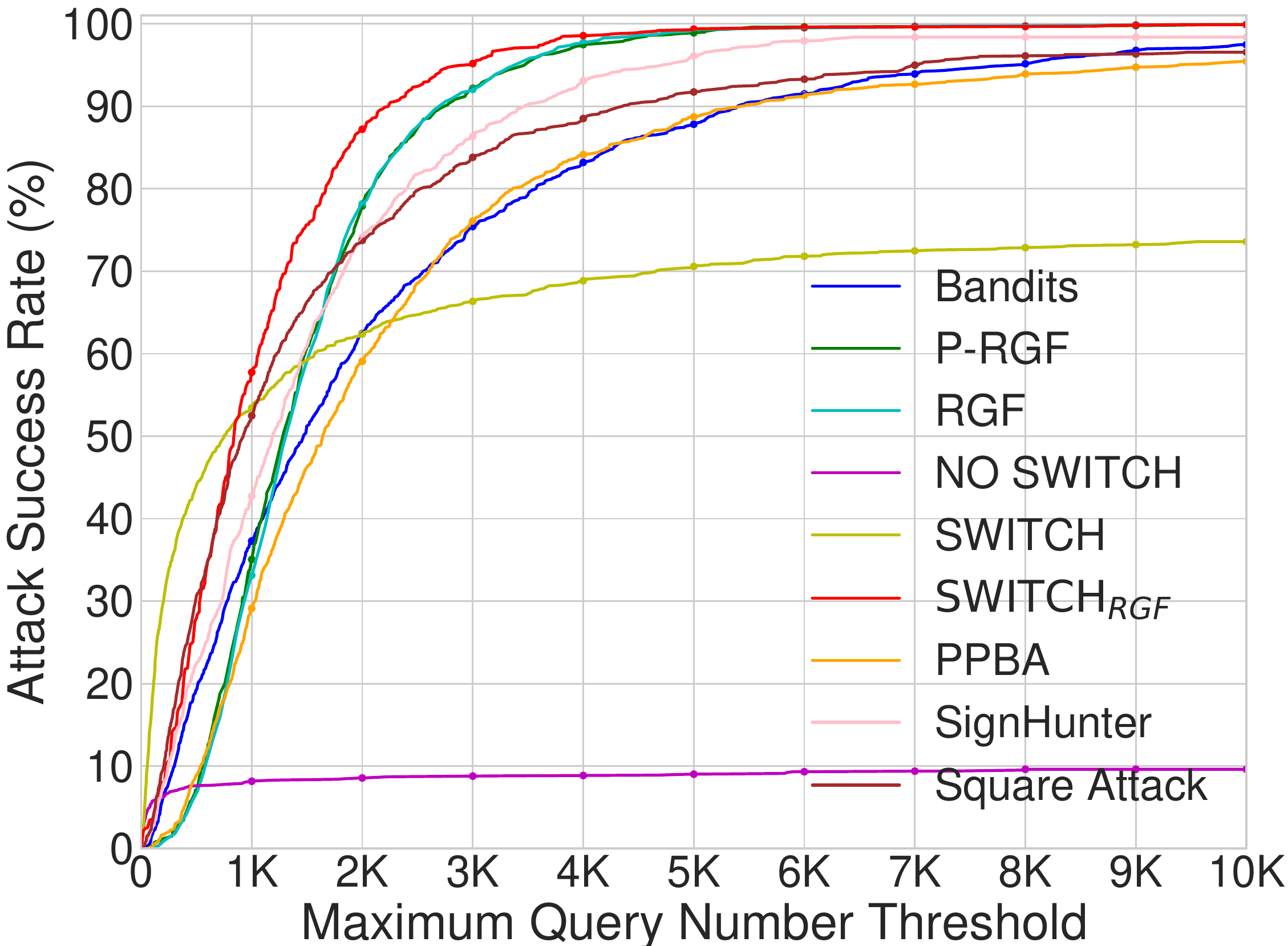}
			\subcaption{targeted $\ell_2$ attack PyramidNet-272}
		\end{minipage}
		\begin{minipage}[b]{.23\textwidth}
			\includegraphics[width=\linewidth]{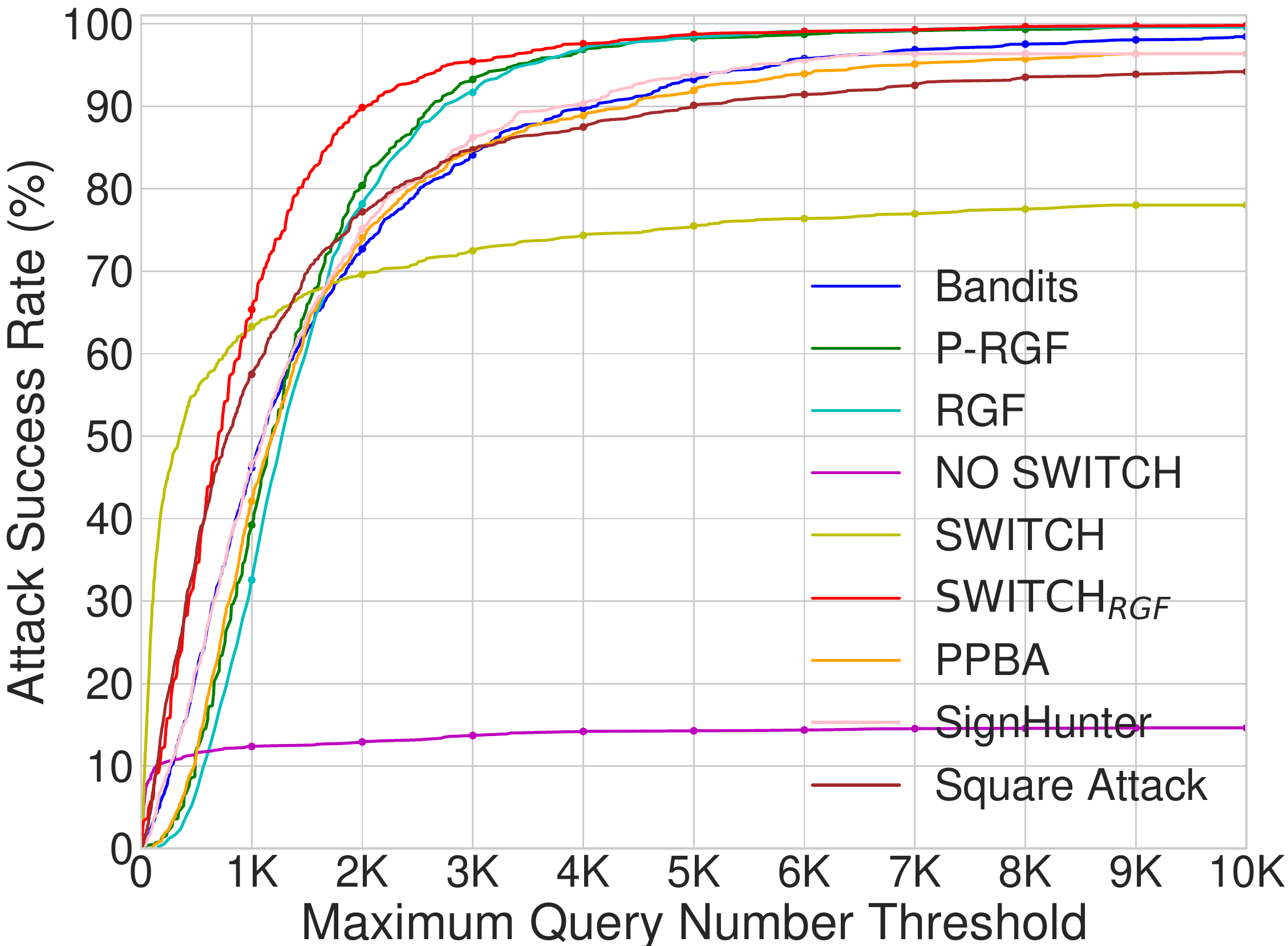}
			\subcaption{targeted $\ell_2$ attack GDAS}
		\end{minipage}
		\begin{minipage}[b]{.23\textwidth}
			\includegraphics[width=\linewidth]{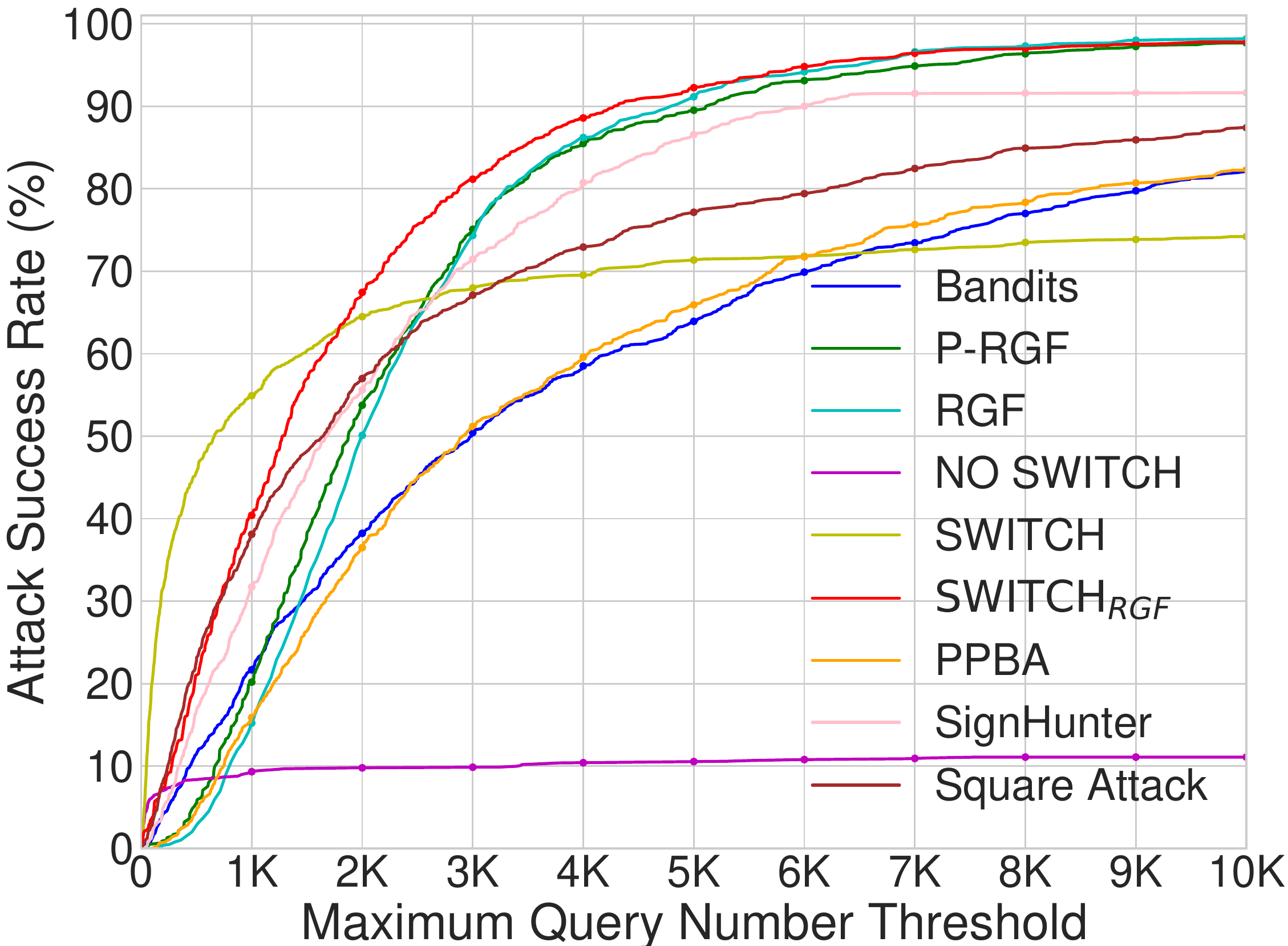}
			\subcaption{targeted $\ell_2$ attack WRN-28}
		\end{minipage}
		\begin{minipage}[b]{.23\textwidth}
			\includegraphics[width=\linewidth]{figures/query_threshold_success_rate/CIFAR-100_WRN-40-10-drop_l2_untargeted_attack_query_threshold_success_rate_dict.pdf}
			\subcaption{targeted $\ell_2$ attack WRN-40}
		\end{minipage}
		\caption{Comparison of attack success rate at different limited maximum queries on CIFAR-100 dataset.}
		\label{fig:query_to_attack_success_rate_CIFAR-100}
	\end{figure*}

\begin{figure*}[bp]
		\setlength{\abovecaptionskip}{0pt}
		\setlength{\belowcaptionskip}{0pt}
		\captionsetup[sub]{font={scriptsize}}
		\centering 
		\begin{minipage}[b]{.3\textwidth}
			\includegraphics[width=\linewidth]{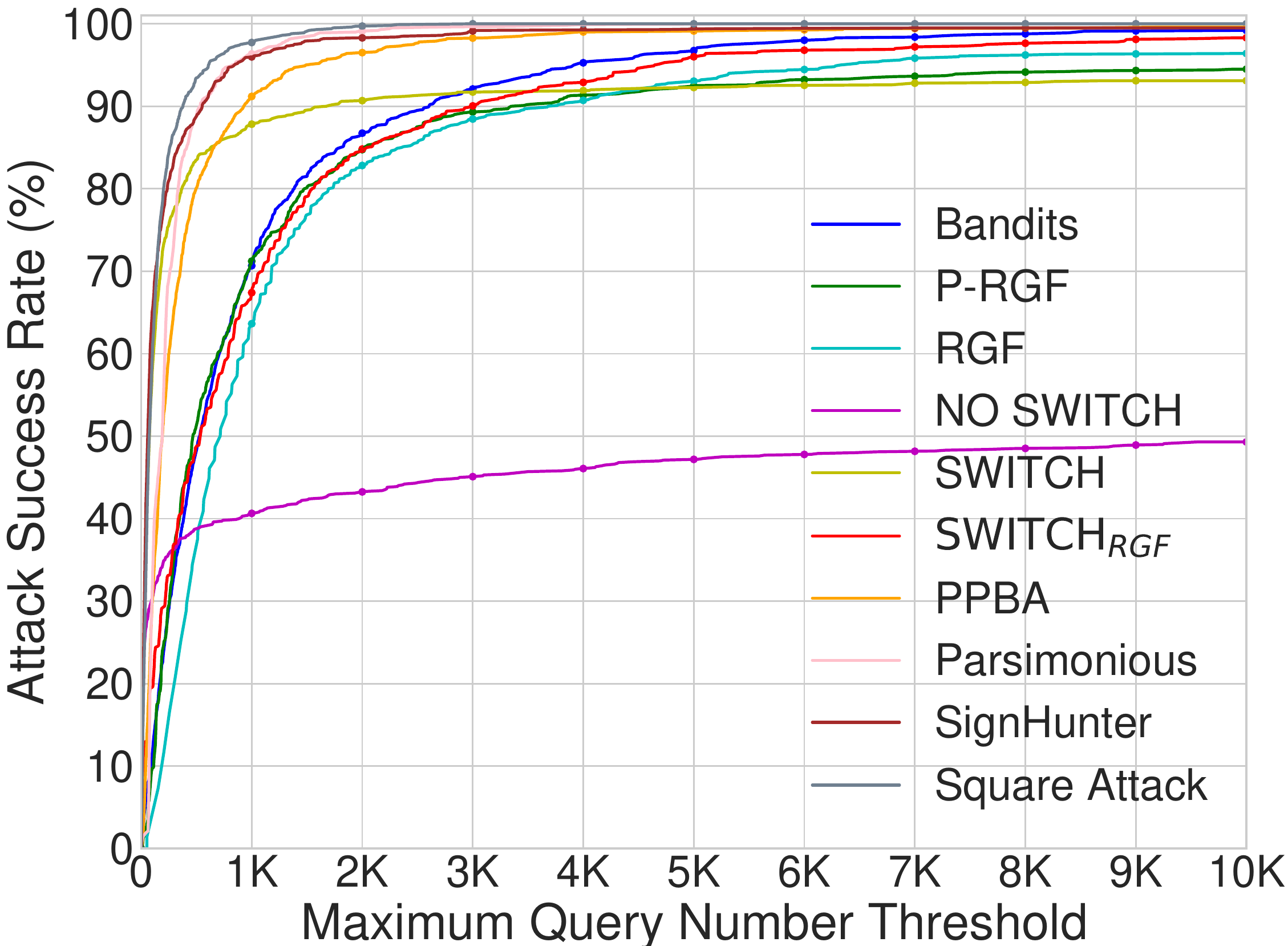}
			\subcaption{untargeted $\ell_\infty$ attack DenseNet-121}
		\end{minipage}
		\begin{minipage}[b]{.3\textwidth}
			\includegraphics[width=\linewidth]{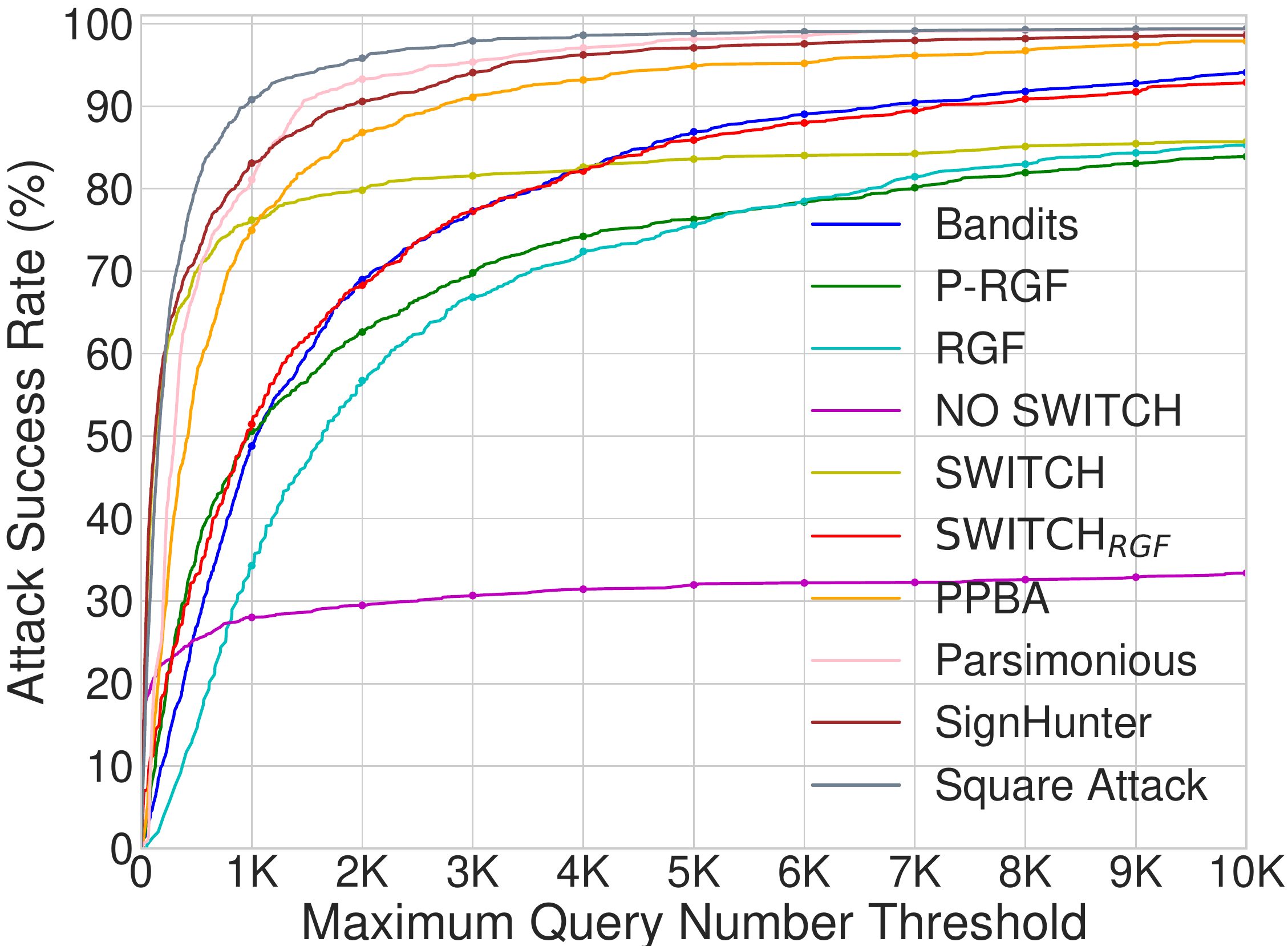}
			\subcaption{untargeted $\ell_\infty$ attack ResNext-101~(32$\times$4d)}
		\end{minipage}
		\begin{minipage}[b]{.3\textwidth}
			\includegraphics[width=\linewidth]{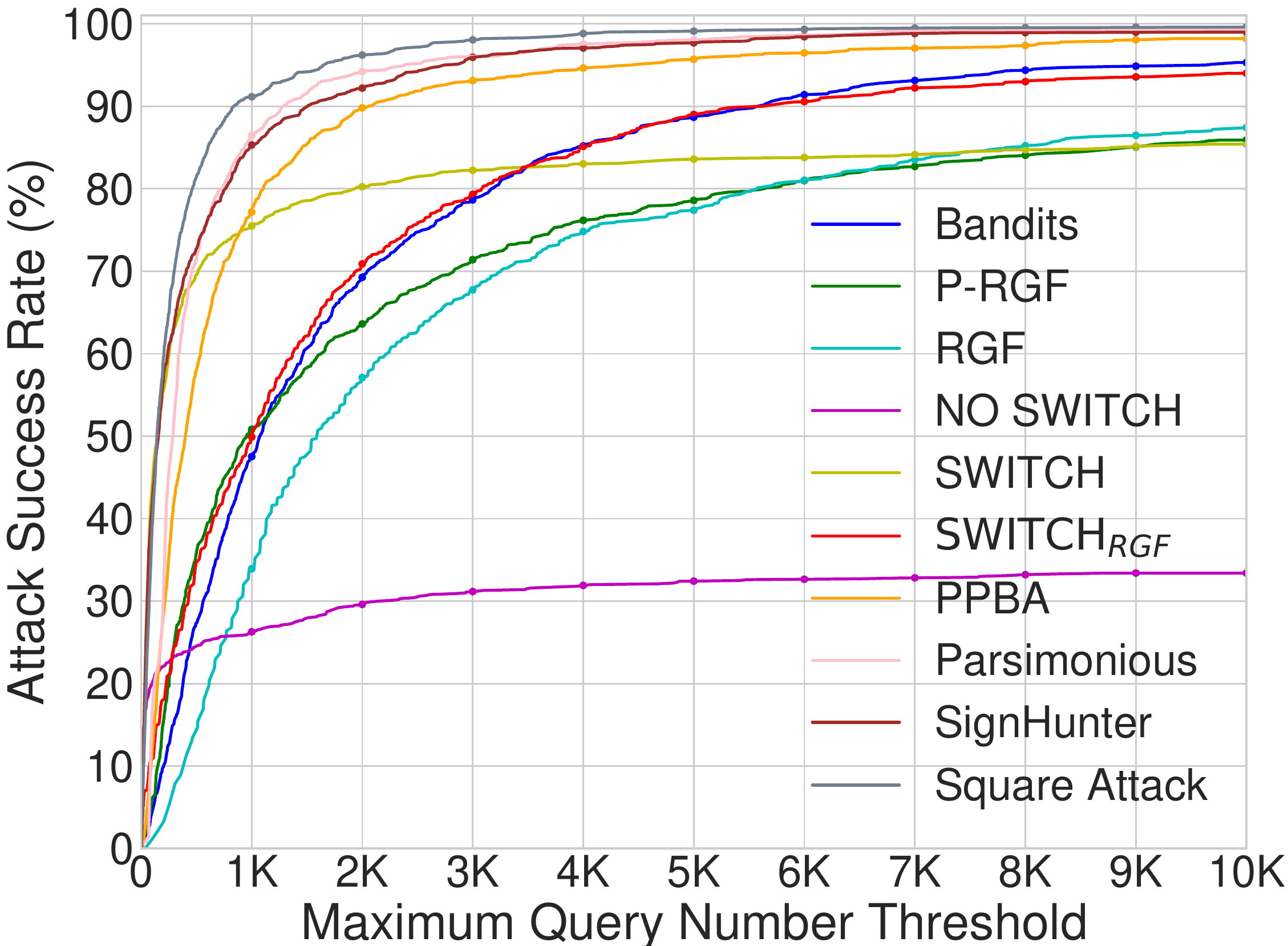}
			\subcaption{untargeted $\ell_\infty$ attack ResNext-101~(64$\times$4d)}
		\end{minipage}
		
		\begin{minipage}[b]{.3\textwidth}
			\includegraphics[width=\linewidth]{figures/query_threshold_success_rate/TinyImageNet_densenet121_l2_untargeted_attack_query_threshold_success_rate_dict.pdf}
			\subcaption{untargeted $\ell_2$ attack DenseNet-121}
		\end{minipage}
		\begin{minipage}[b]{.3\textwidth}
			\includegraphics[width=\linewidth]{figures/query_threshold_success_rate/TinyImageNet_resnext32_4_l2_untargeted_attack_query_threshold_success_rate_dict.pdf}
			\subcaption{untargeted $\ell_2$ attack ResNext-101~(32$\times$4d)}
		\end{minipage}
		\begin{minipage}[b]{.3\textwidth}
			\includegraphics[width=\linewidth]{figures/query_threshold_success_rate/TinyImageNet_resnext64_4_l2_untargeted_attack_query_threshold_success_rate_dict.pdf}
			\subcaption{untargeted $\ell_2$ attack ResNext-101~(64$\times$4d)}
		\end{minipage}
		
		\begin{minipage}[b]{.3\textwidth}
			\includegraphics[width=\linewidth]{figures/query_threshold_success_rate/TinyImageNet_densenet121_l2_targeted_attack_query_threshold_success_rate_dict.pdf}
			\subcaption{targeted $\ell_2$ attack DenseNet-121}
		\end{minipage}
		\begin{minipage}[b]{.3\textwidth}
			\includegraphics[width=\linewidth]{figures/query_threshold_success_rate/TinyImageNet_resnext32_4_l2_targeted_attack_query_threshold_success_rate_dict.pdf}
			\subcaption{targeted $\ell_2$ attack ResNext-101~(32$\times$4d)}
		\end{minipage}
		\begin{minipage}[b]{.3\textwidth}
			\includegraphics[width=\linewidth]{figures/query_threshold_success_rate/TinyImageNet_resnext64_4_l2_targeted_attack_query_threshold_success_rate_dict.pdf}
			\subcaption{targeted $\ell_2$ attack ResNext-101~(64$\times$4d)}
		\end{minipage}
		\caption{Comparison of the attack success rate at different limited maximum queries in TinyImageNet dataset.}
		\label{fig:query_to_attack_success_rate_TinyImageNet}
	\end{figure*}

	\noindent\textbf{Attack Success Rates at Different Maximum Queries.} Fig. \ref{fig:query_to_attack_success_rate_CIFAR-10}, Fig. \ref{fig:query_to_attack_success_rate_CIFAR-100}, and Fig. \ref{fig:query_to_attack_success_rate_TinyImageNet} show the attack success rates by limiting different maximum queries in each example. 
	
\begin{figure*}[htbp]
		\setlength{\abovecaptionskip}{0pt}
		\setlength{\belowcaptionskip}{0pt}
		\captionsetup[sub]{font={scriptsize}}
		\centering 
		\begin{minipage}[b]{.23\textwidth}
			\includegraphics[width=\linewidth]{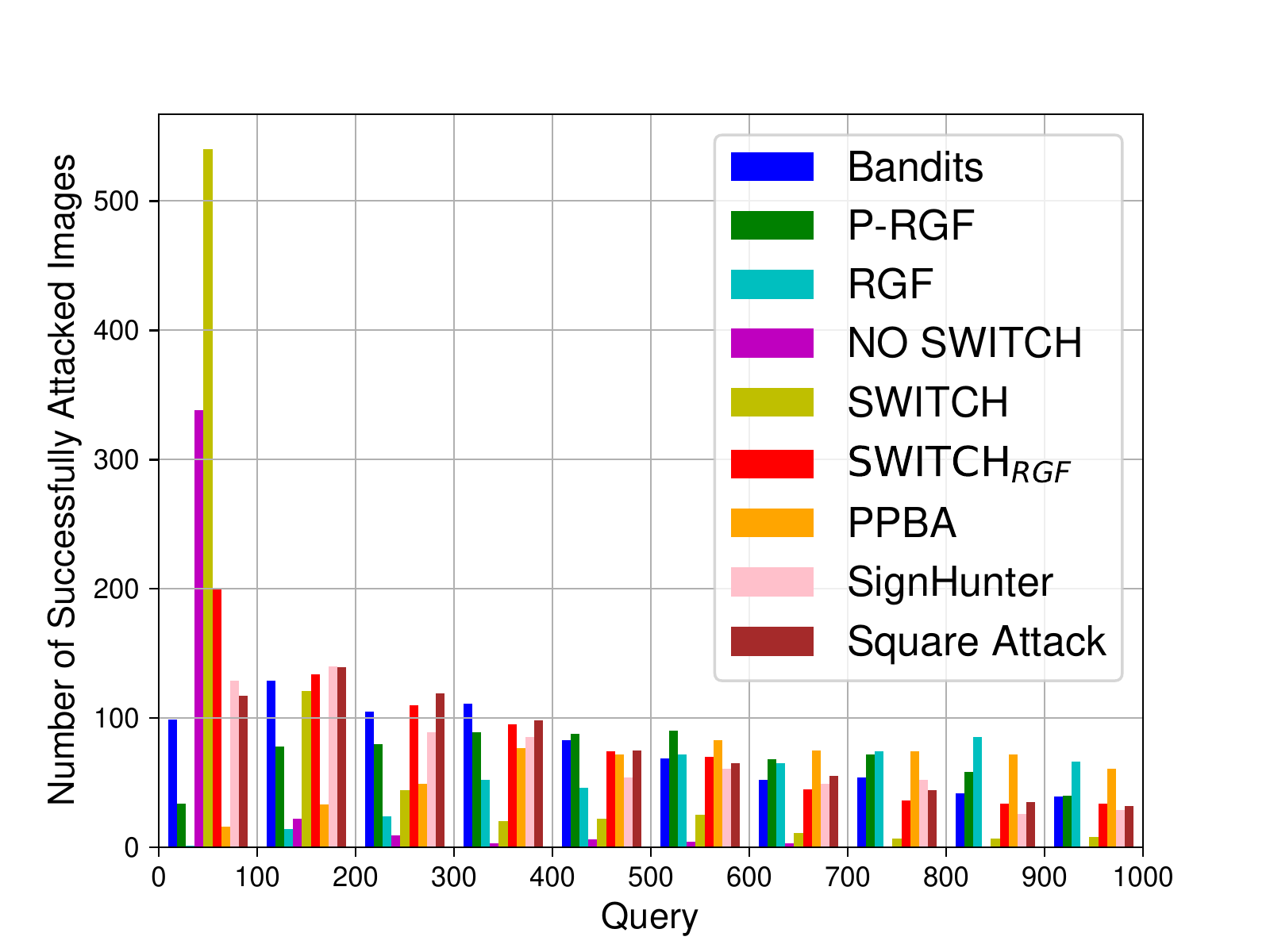}
			\subcaption{untargeted $\ell_2$ attack PyramidNet-272}
		\end{minipage}
		\begin{minipage}[b]{.23\textwidth}
			\includegraphics[width=\linewidth]{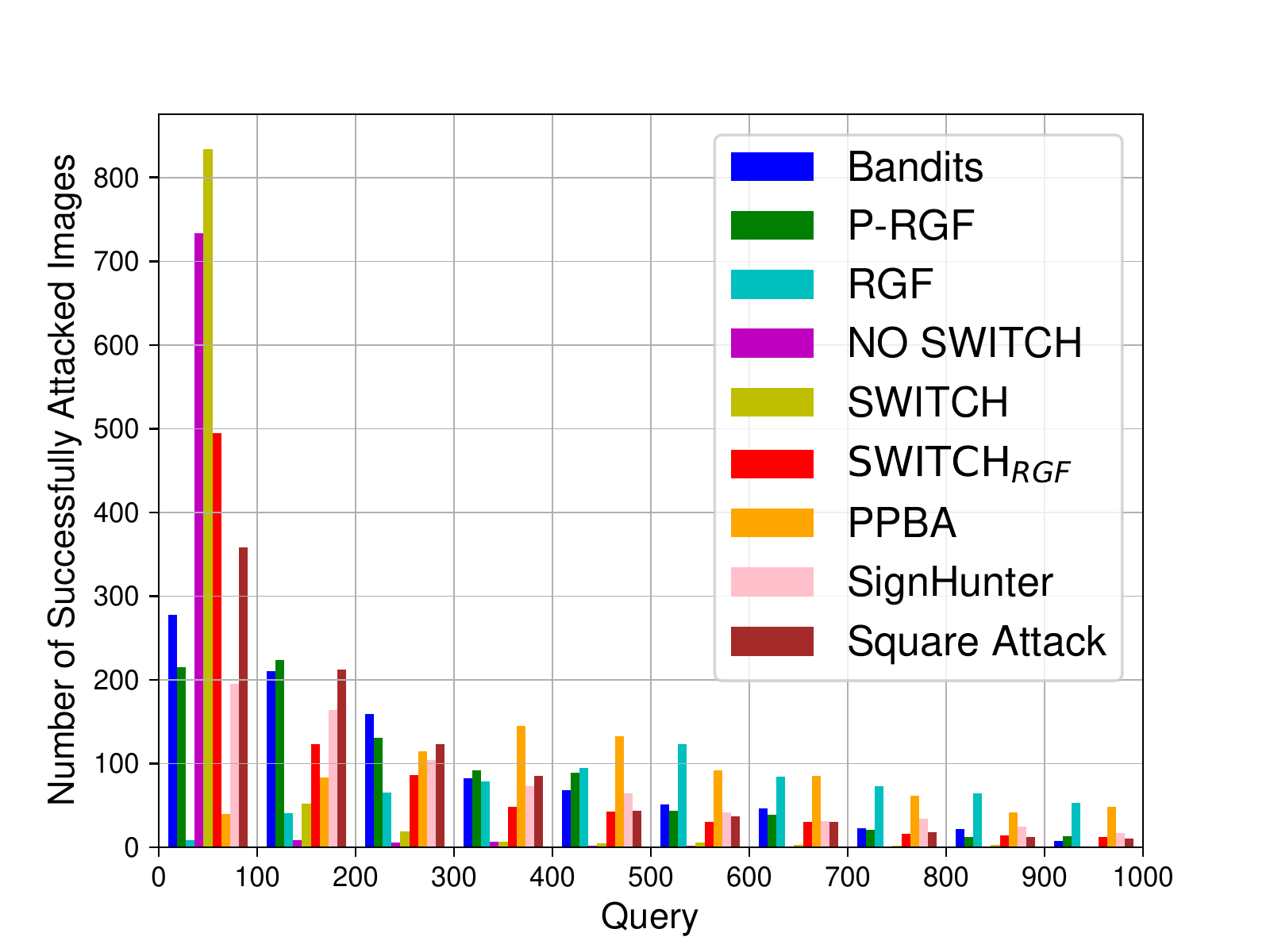}
			\subcaption{untargeted $\ell_2$ attack GDAS}
		\end{minipage}
		\begin{minipage}[b]{.23\textwidth}
			\includegraphics[width=\linewidth]{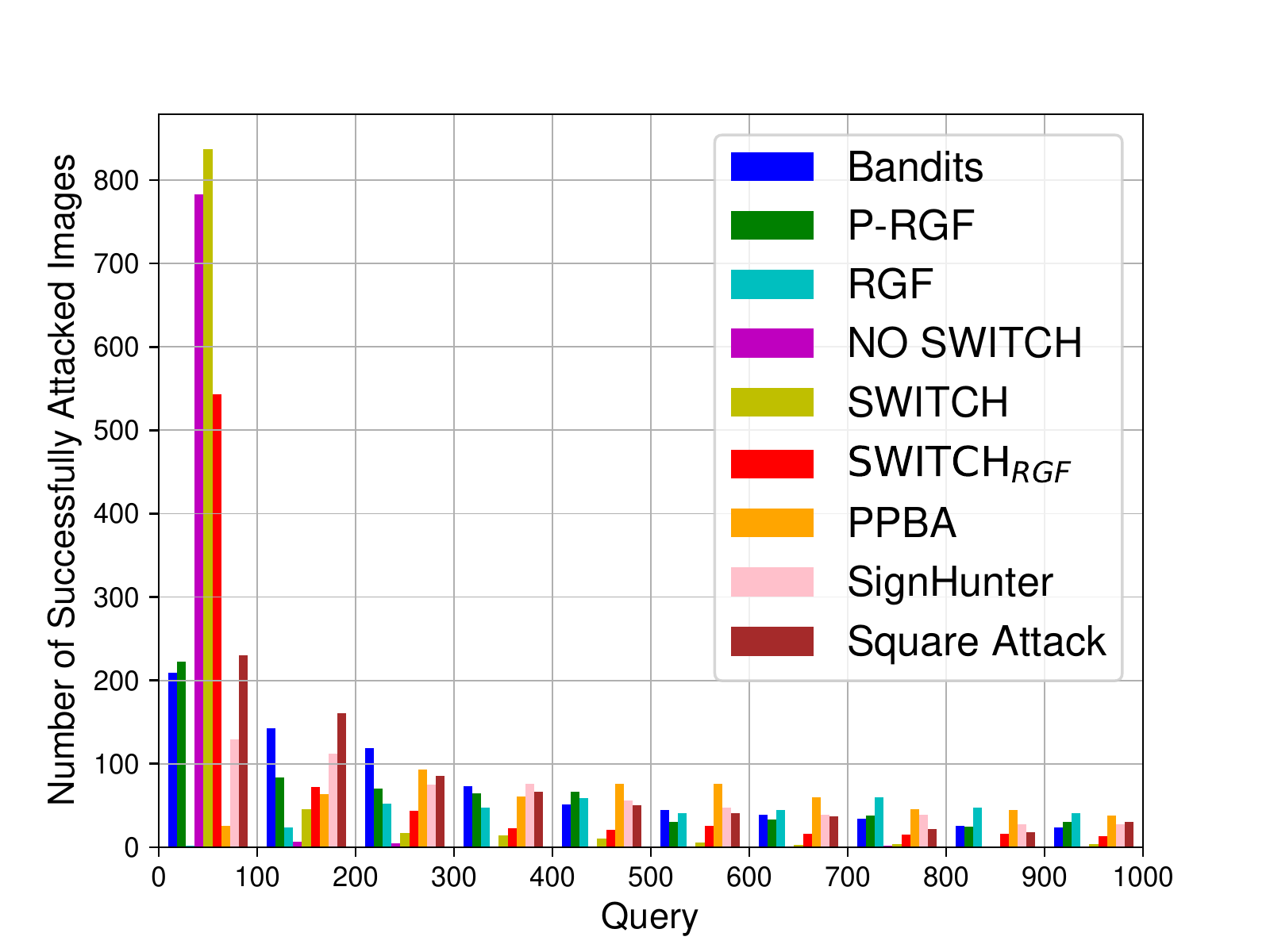}
			\subcaption{untargeted $\ell_2$ attack WRN-28}
		\end{minipage}
		\begin{minipage}[b]{.23\textwidth}
			\includegraphics[width=\linewidth]{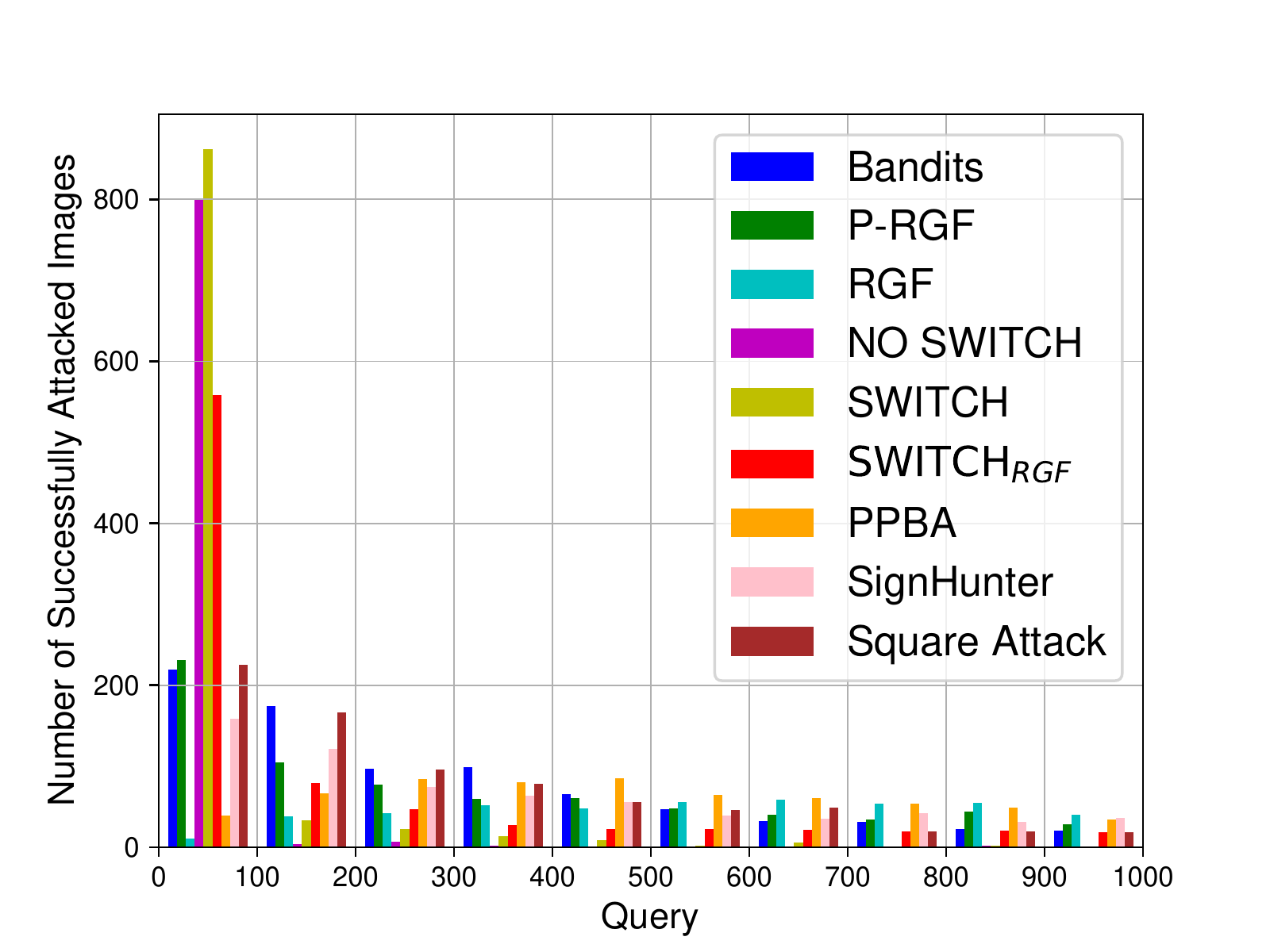}
			\subcaption{untargeted $\ell_2$ attack WRN-40}
		\end{minipage}
		\begin{minipage}[b]{.23\textwidth}
			\includegraphics[width=\linewidth]{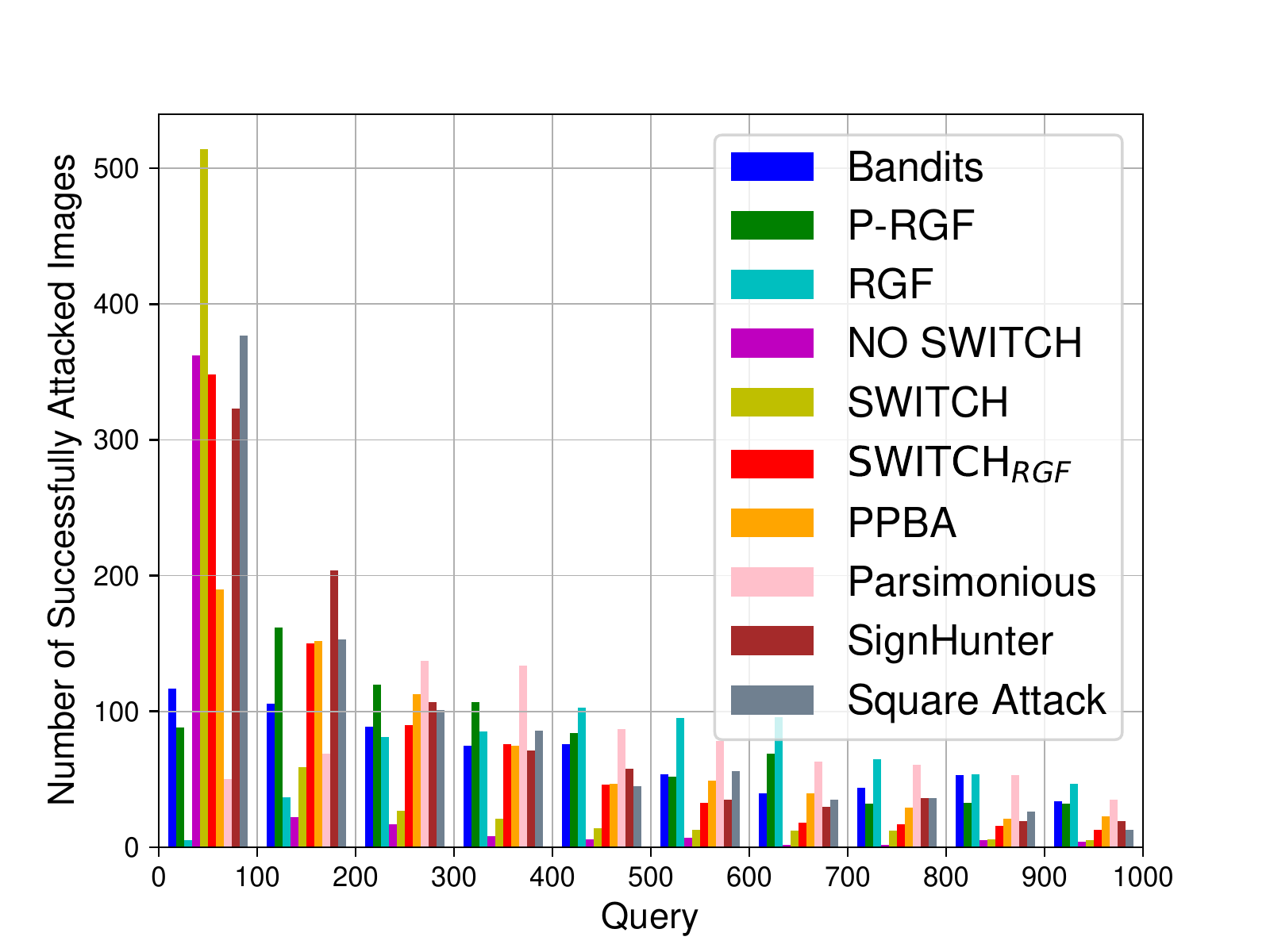}
			\subcaption{untargeted $\ell_\infty$ attack PyramidNet-272}
		\end{minipage}
		\begin{minipage}[b]{.23\textwidth}
			\includegraphics[width=\linewidth]{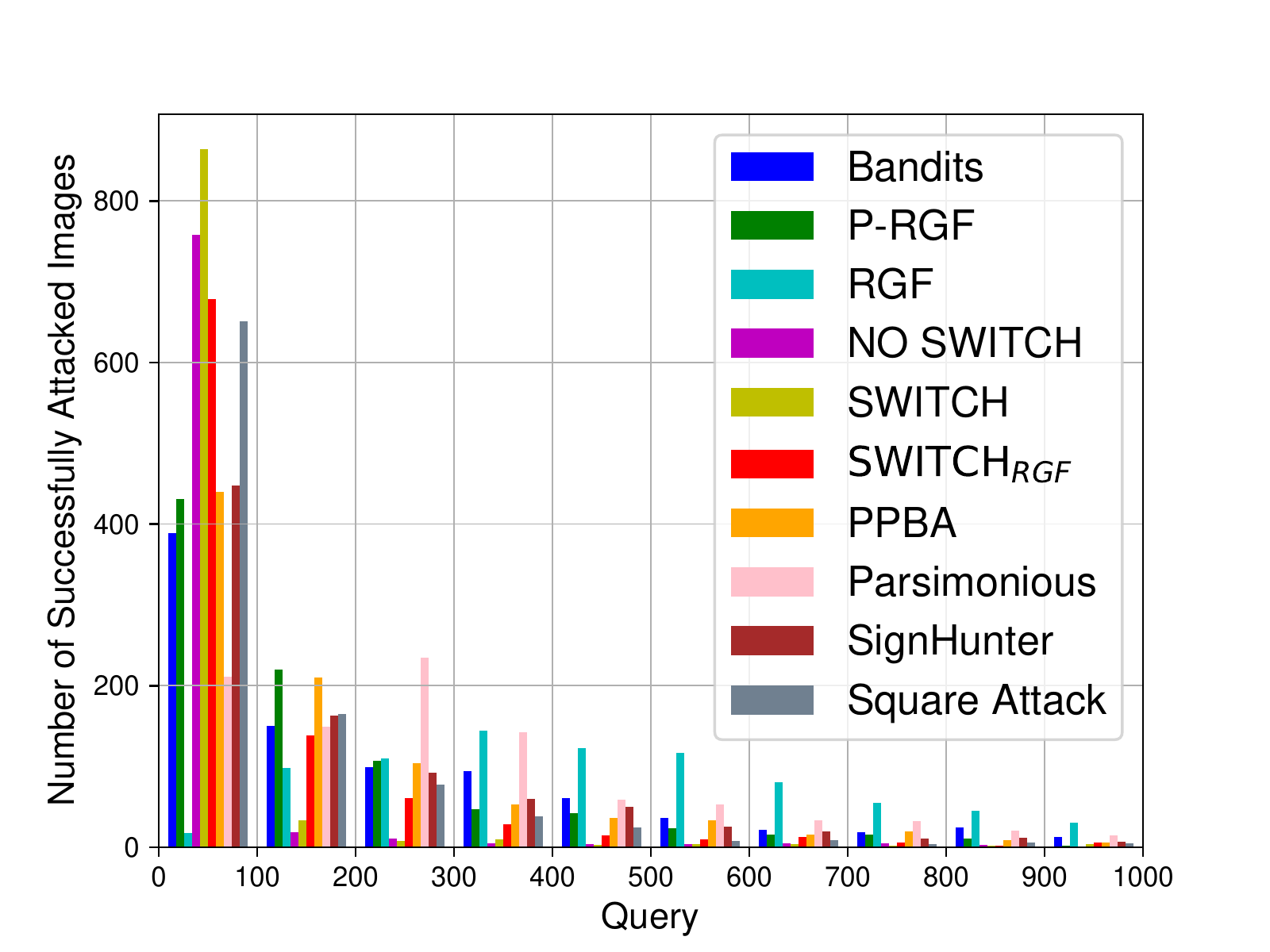}
			\subcaption{untargeted $\ell_\infty$ attack GDAS}
		\end{minipage}
		\begin{minipage}[b]{.23\textwidth}
			\includegraphics[width=\linewidth]{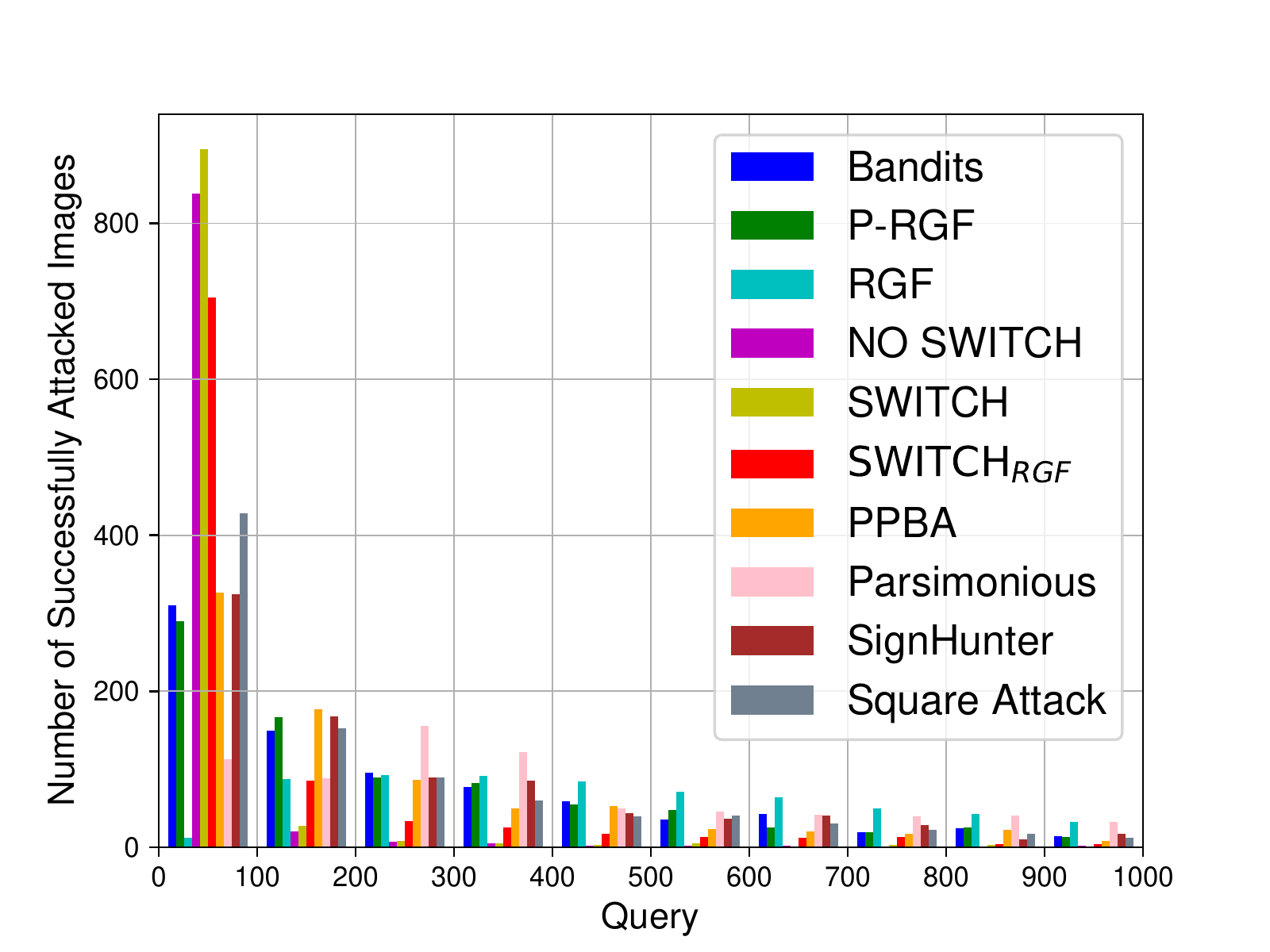}
			\subcaption{untargeted $\ell_\infty$ attack WRN-28}
		\end{minipage}
		\begin{minipage}[b]{.23\textwidth}
			\includegraphics[width=\linewidth]{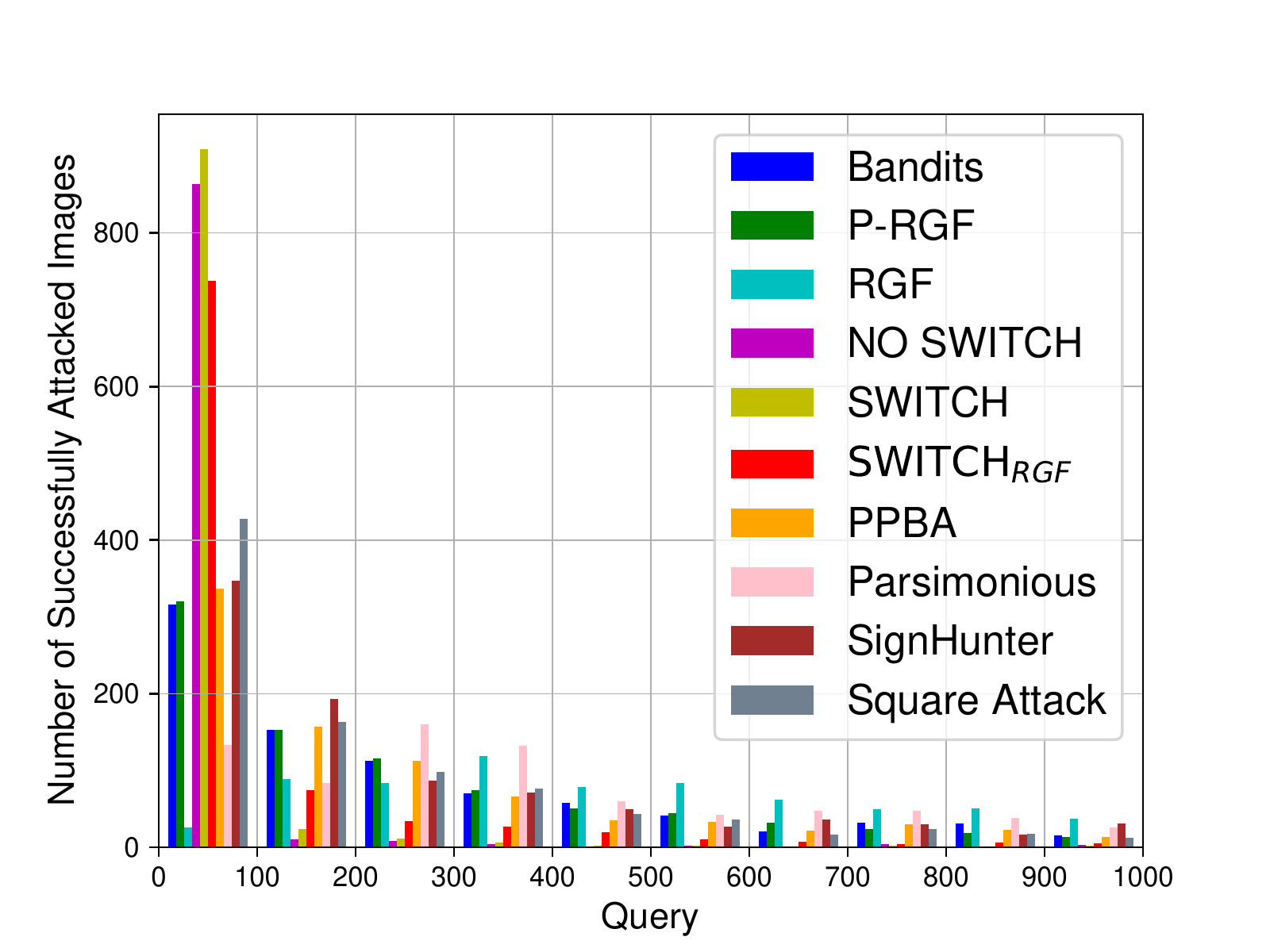}
			\subcaption{untargeted $\ell_\infty$ attack WRN-40}
		\end{minipage}
		\begin{minipage}[b]{.23\textwidth}
			\includegraphics[width=\linewidth]{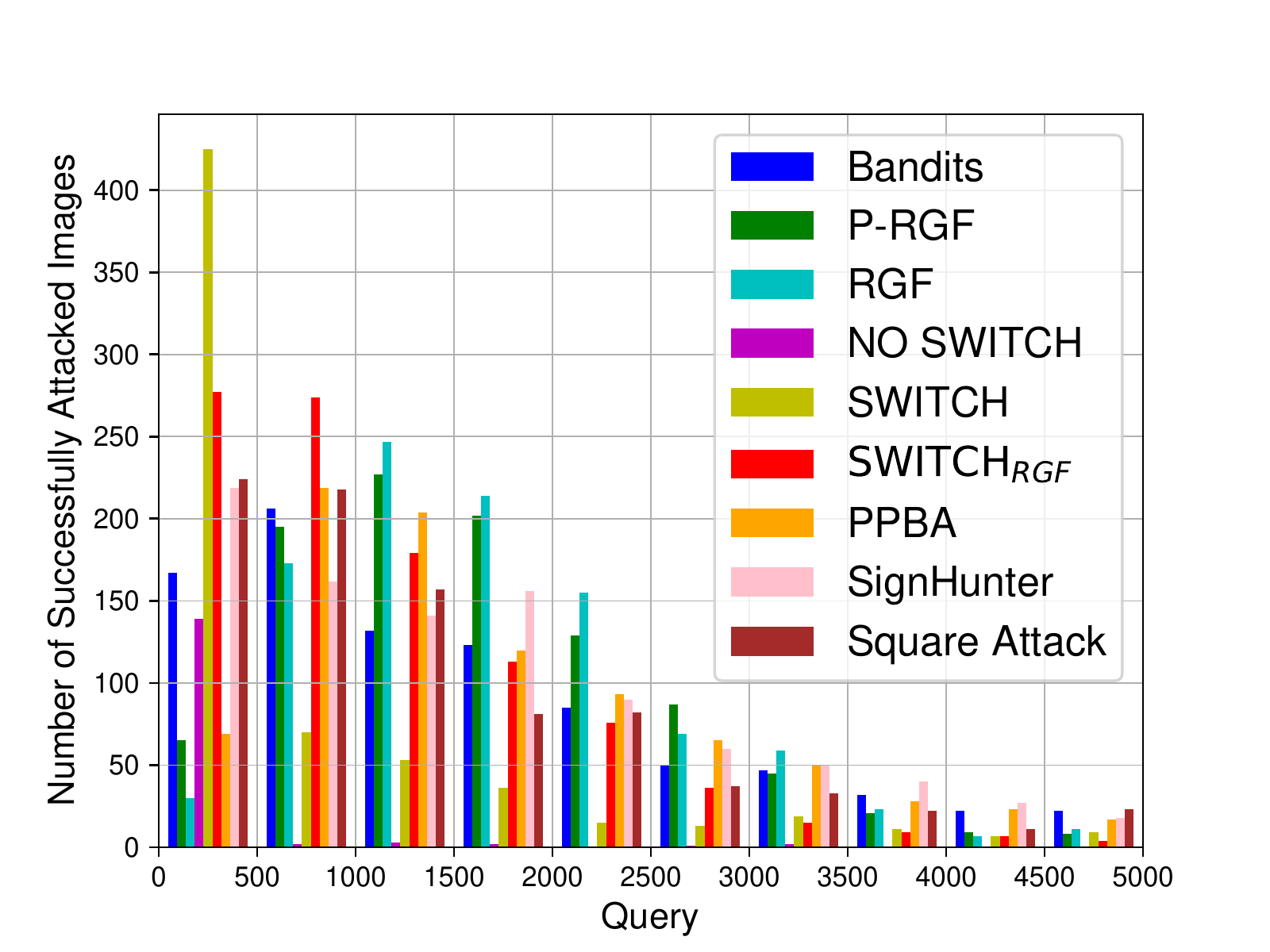}
			\subcaption{targeted $\ell_2$ attack PyramidNet-272}
		\end{minipage}
		\begin{minipage}[b]{.23\textwidth}
			\includegraphics[width=\linewidth]{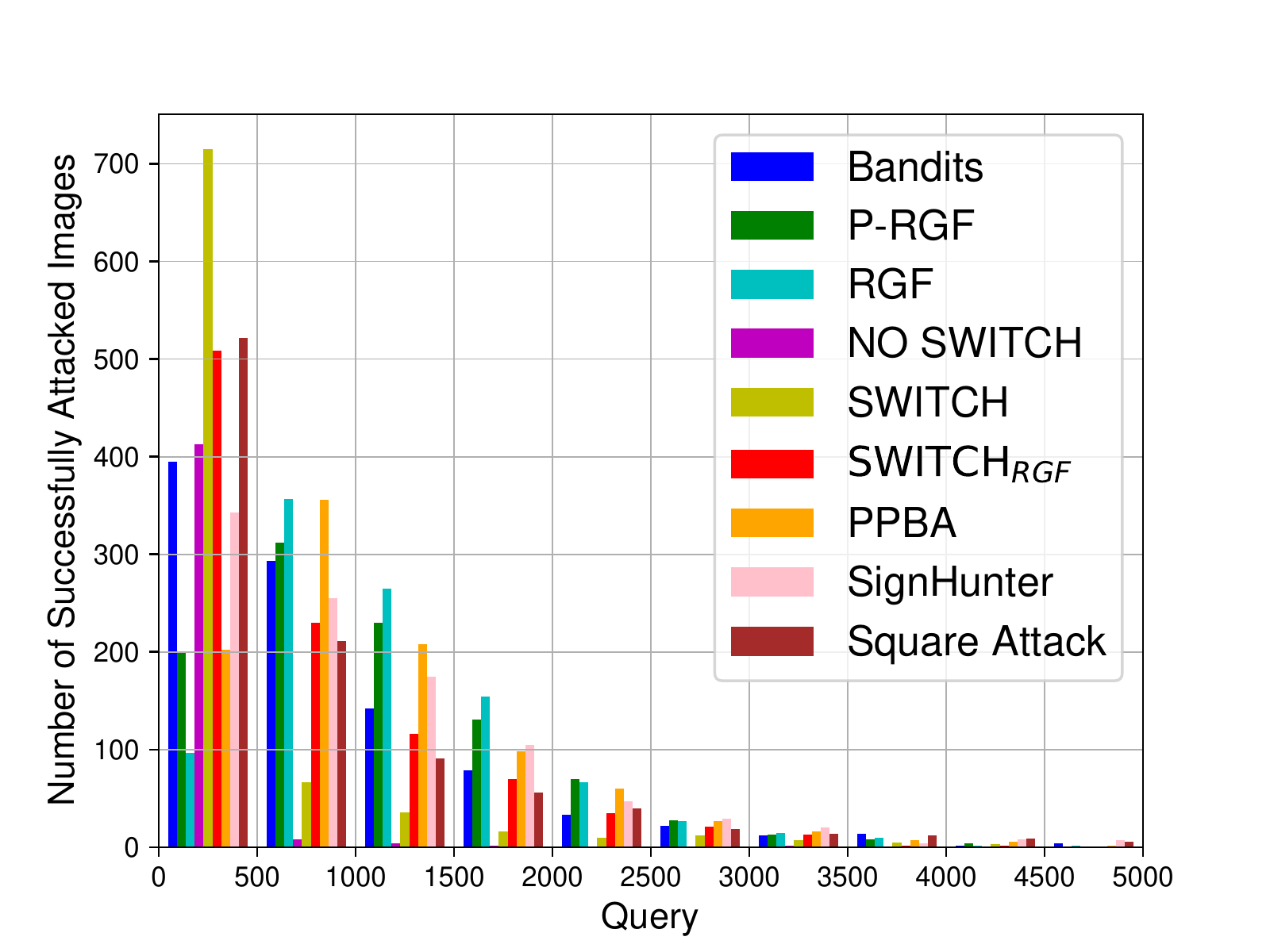}
			\subcaption{targeted $\ell_2$ attack GDAS}
		\end{minipage}
		\begin{minipage}[b]{.23\textwidth}
			\includegraphics[width=\linewidth]{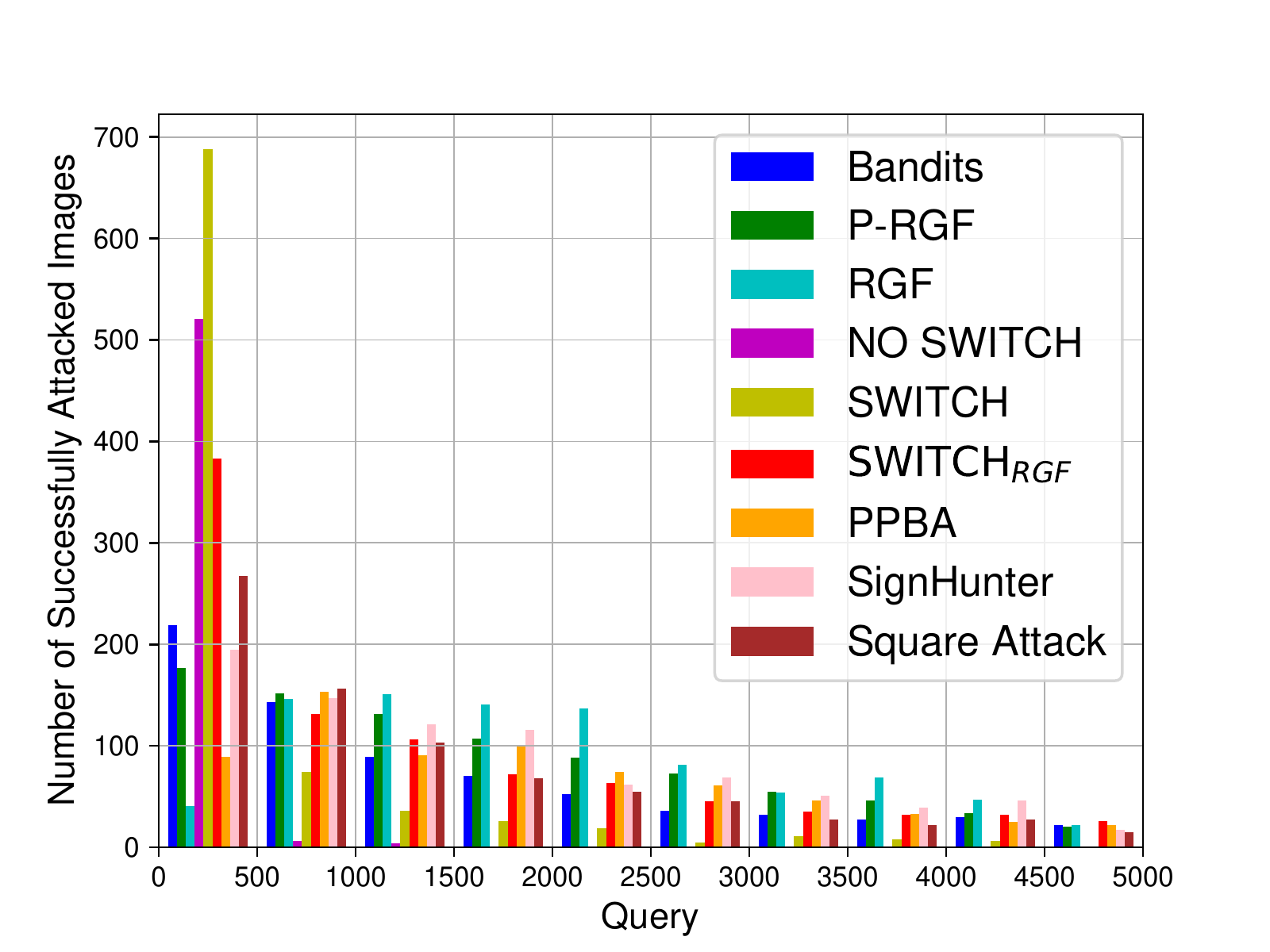}
			\subcaption{targeted $\ell_2$ attack WRN-28}
		\end{minipage}
		\begin{minipage}[b]{.23\textwidth}
			\includegraphics[width=\linewidth]{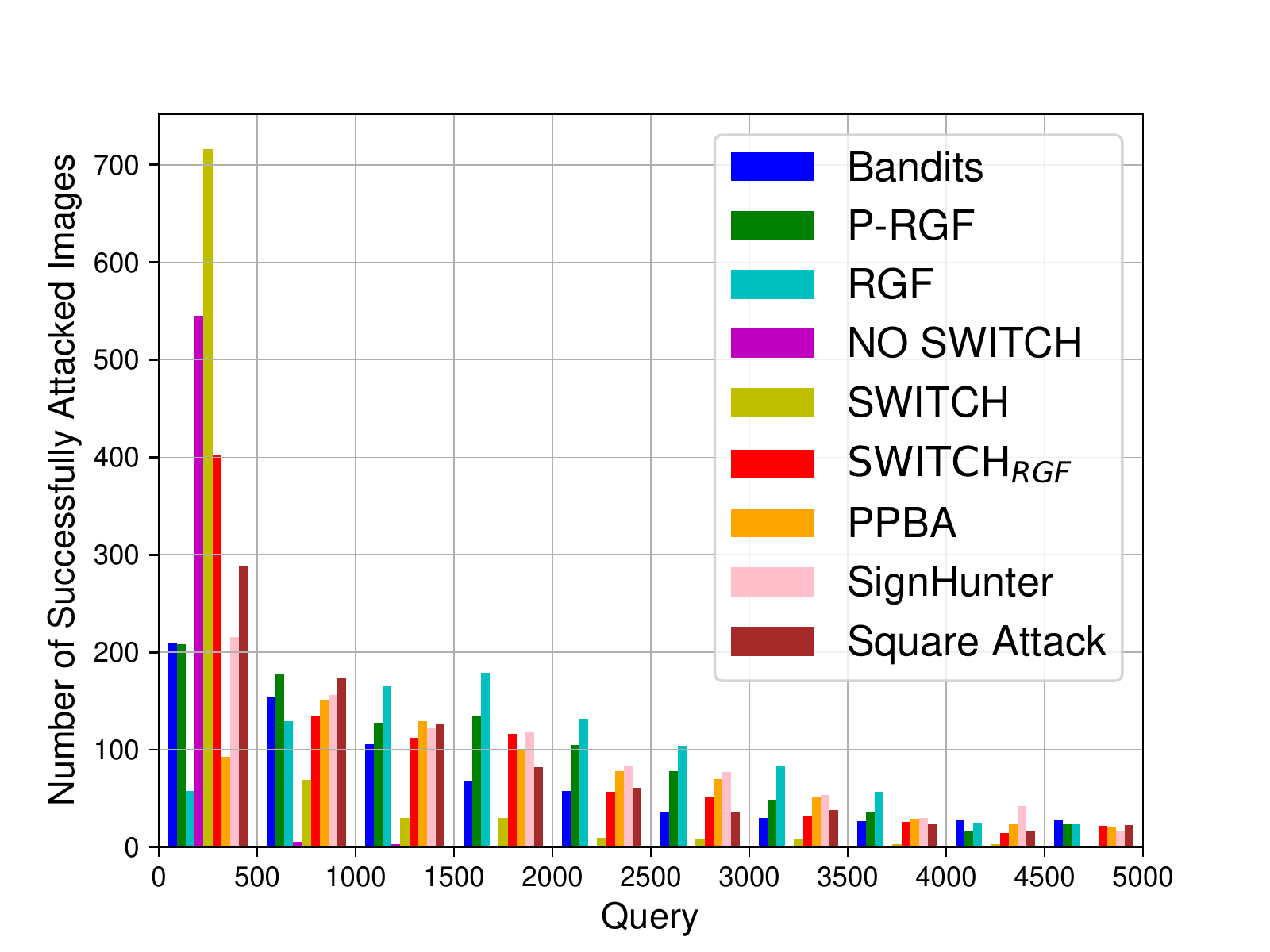}
			\subcaption{targeted $\ell_2$ attack WRN-40}
		\end{minipage}
		\caption{The histogram of query number in the CIFAR-10 dataset.}
		\label{fig:histogram_CIFAR-10}
	\end{figure*}

\begin{figure*}[htbp]
		\setlength{\abovecaptionskip}{0pt}
		\setlength{\belowcaptionskip}{0pt}
		\captionsetup[sub]{font={scriptsize}}
		\centering 
		\begin{minipage}[b]{.23\textwidth}
			\includegraphics[width=\linewidth]{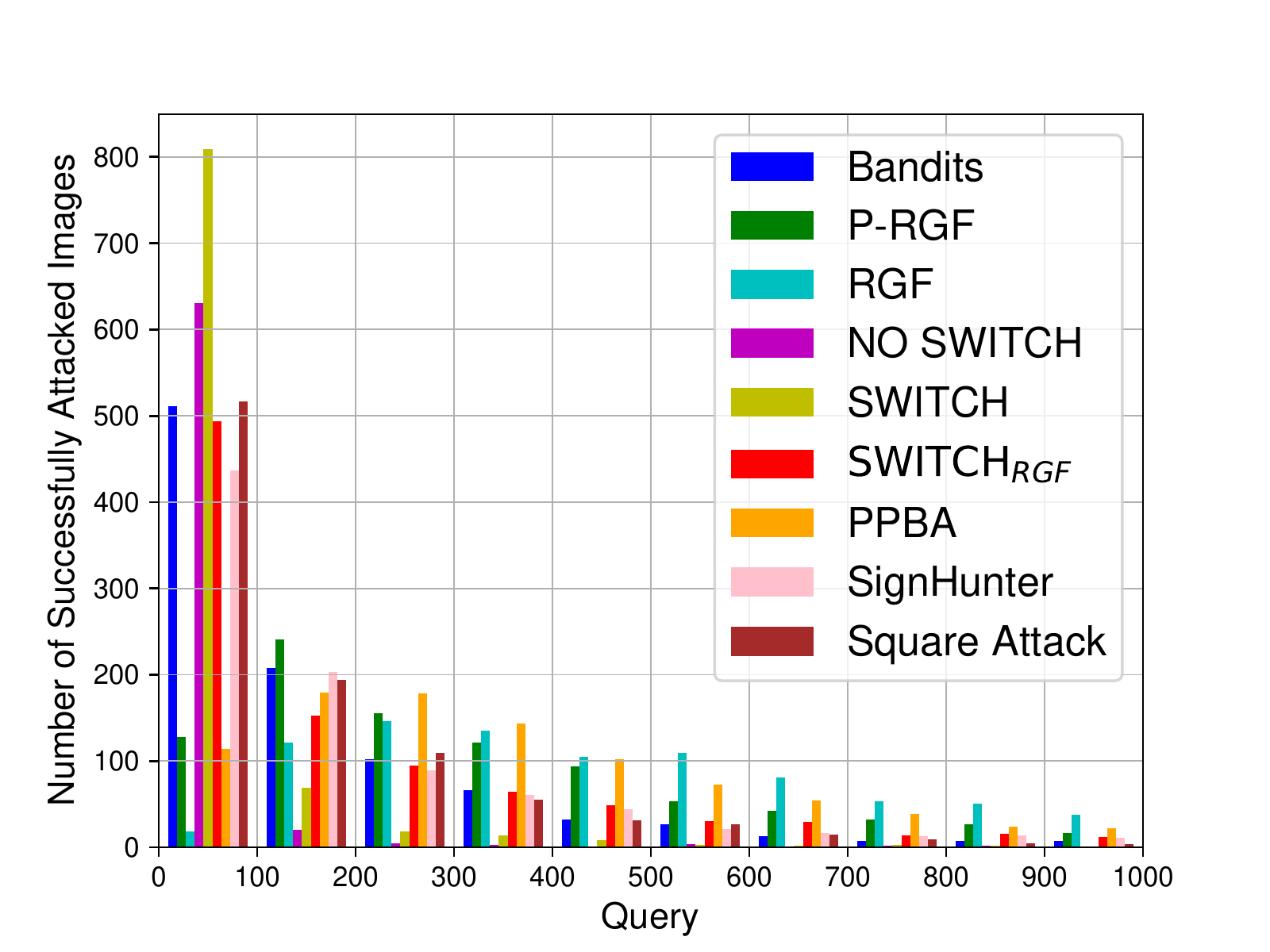}
			\subcaption{untargeted $\ell_2$ attack PyramidNet-272}
		\end{minipage}
		\begin{minipage}[b]{.23\textwidth}
			\includegraphics[width=\linewidth]{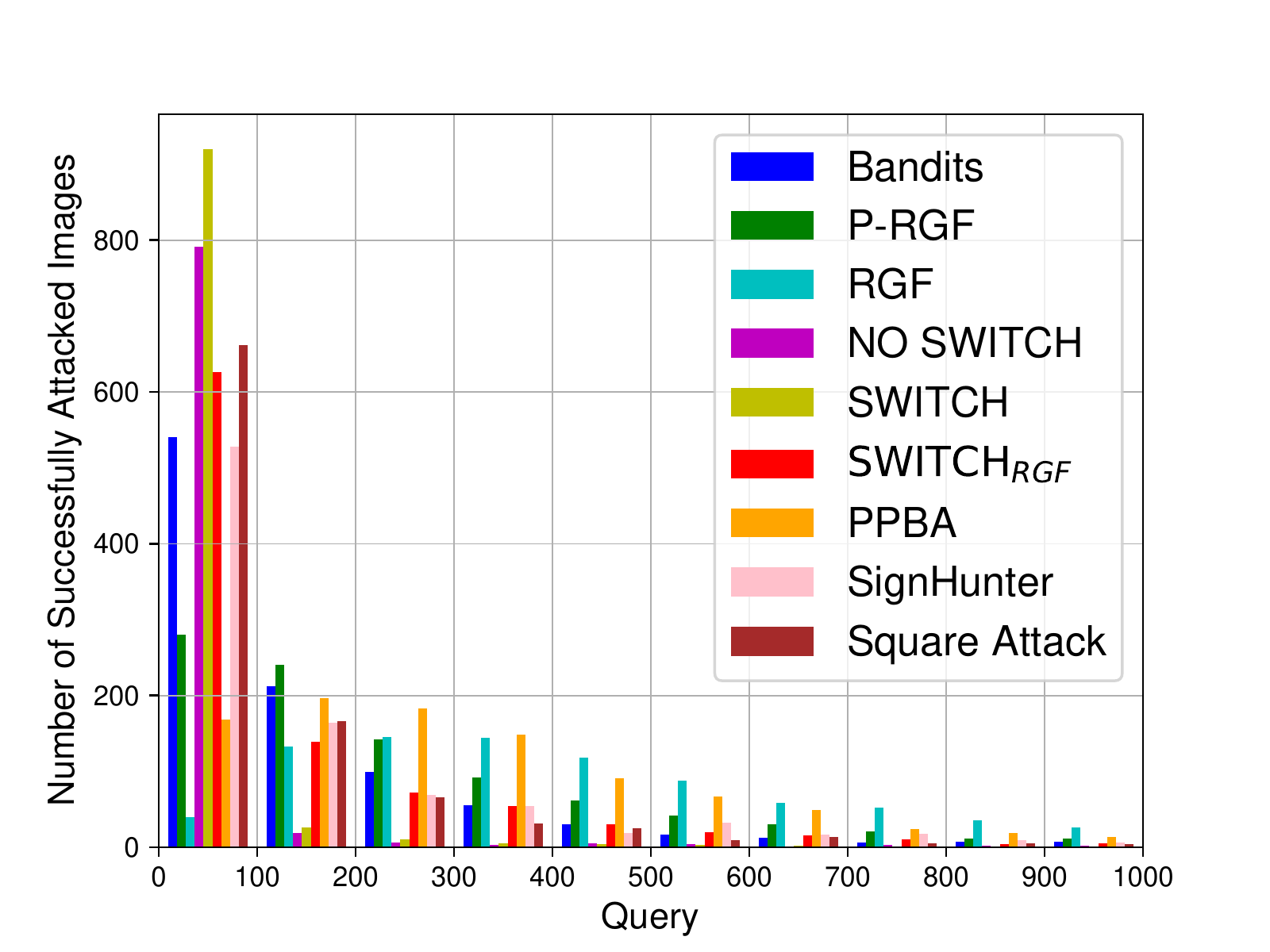}
			\subcaption{untargeted $\ell_2$ attack GDAS}
		\end{minipage}
		\begin{minipage}[b]{.23\textwidth}
			\includegraphics[width=\linewidth]{figures/query_histogram/untargeted_attack/CIFAR-10_l2_untargeted_attack_on_WRN-28-10-drop.pdf}
			\subcaption{untargeted $\ell_2$ attack WRN-28}
		\end{minipage}
		\begin{minipage}[b]{.23\textwidth}
			\includegraphics[width=\linewidth]{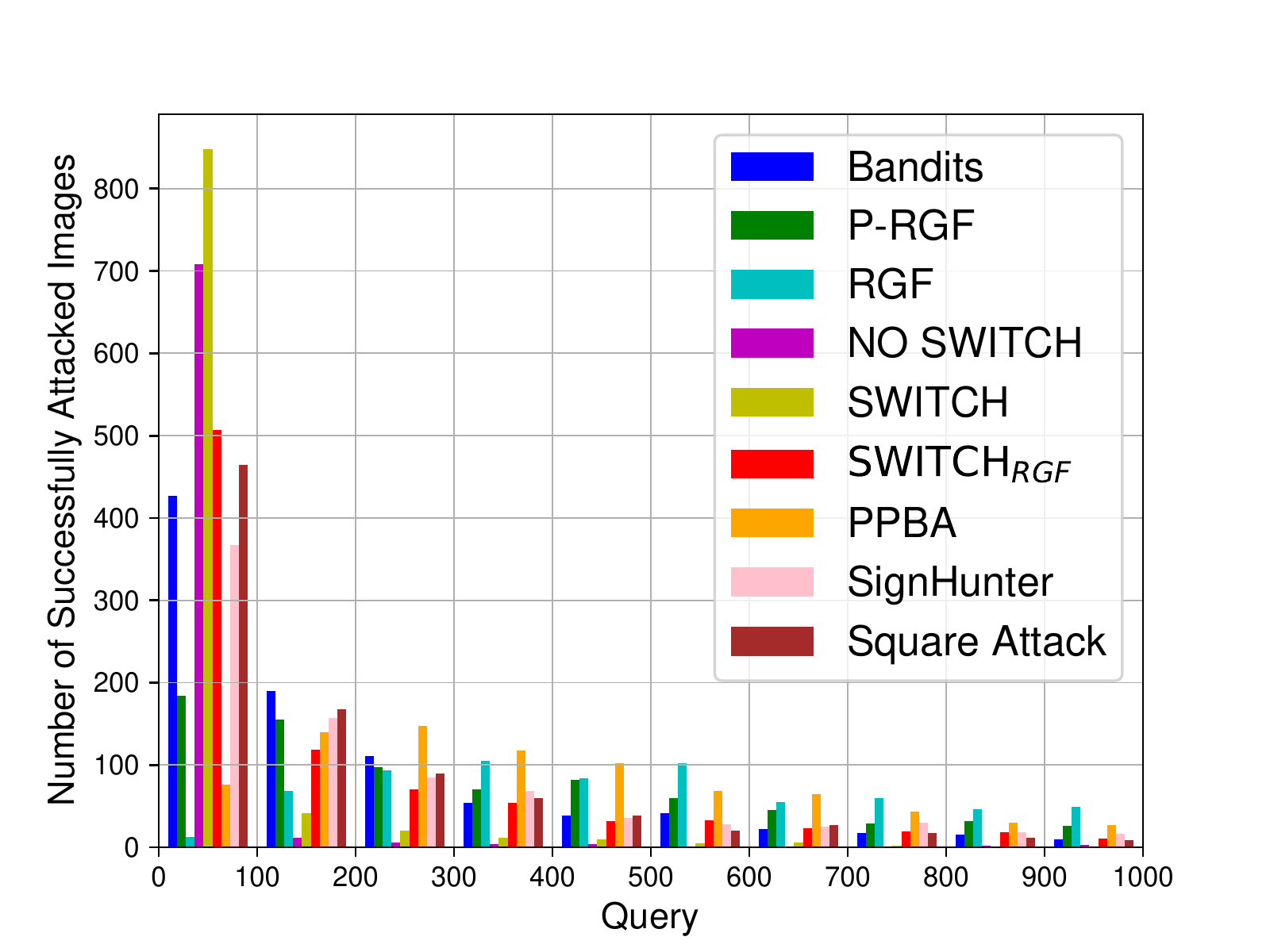}
			\subcaption{untargeted $\ell_2$ attack WRN-40}
		\end{minipage}
		\begin{minipage}[b]{.23\textwidth}
			\includegraphics[width=\linewidth]{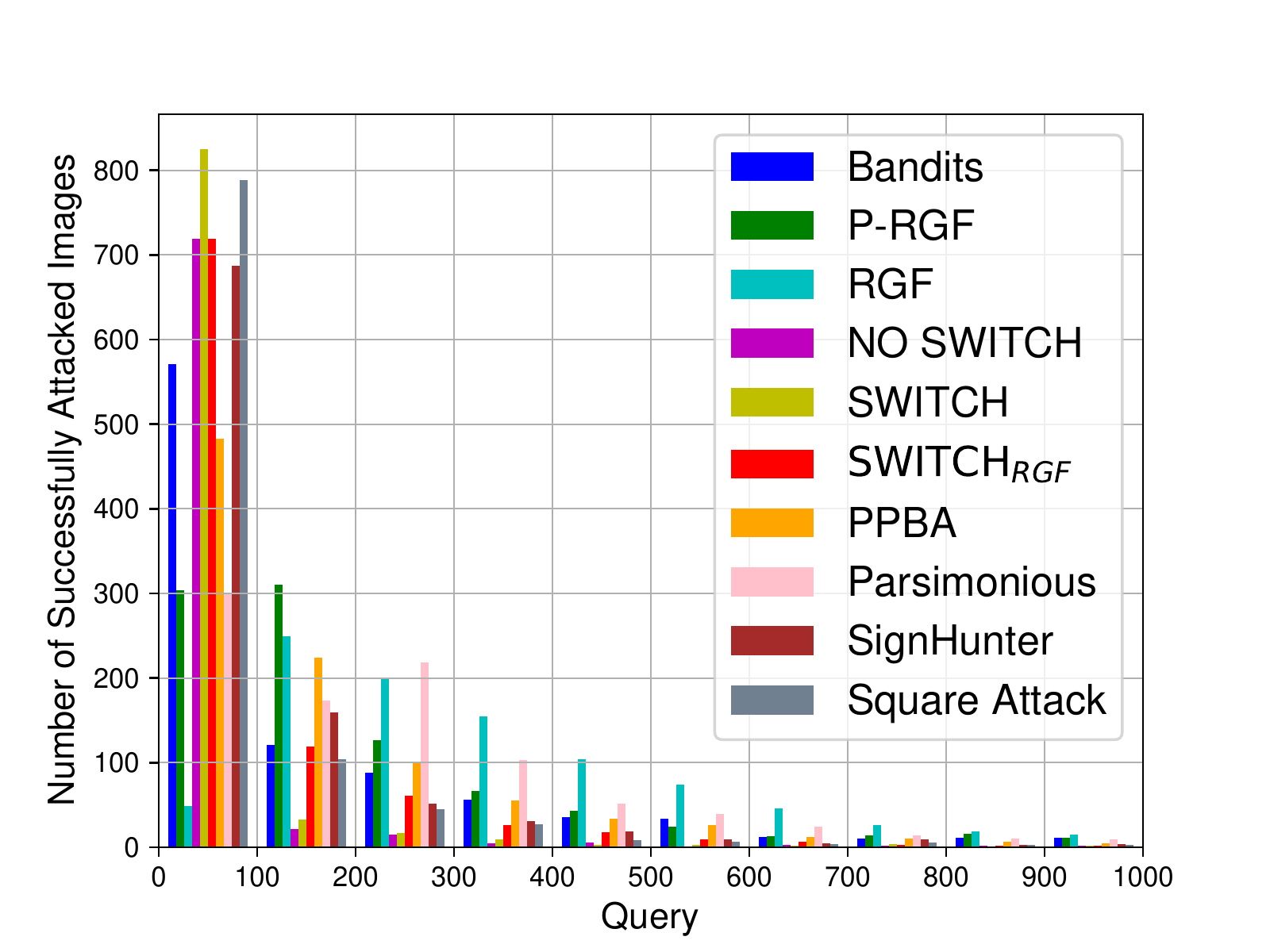}
			\subcaption{untargeted $\ell_\infty$ attack PyramidNet-272}
		\end{minipage}
		\begin{minipage}[b]{.23\textwidth}
			\includegraphics[width=\linewidth]{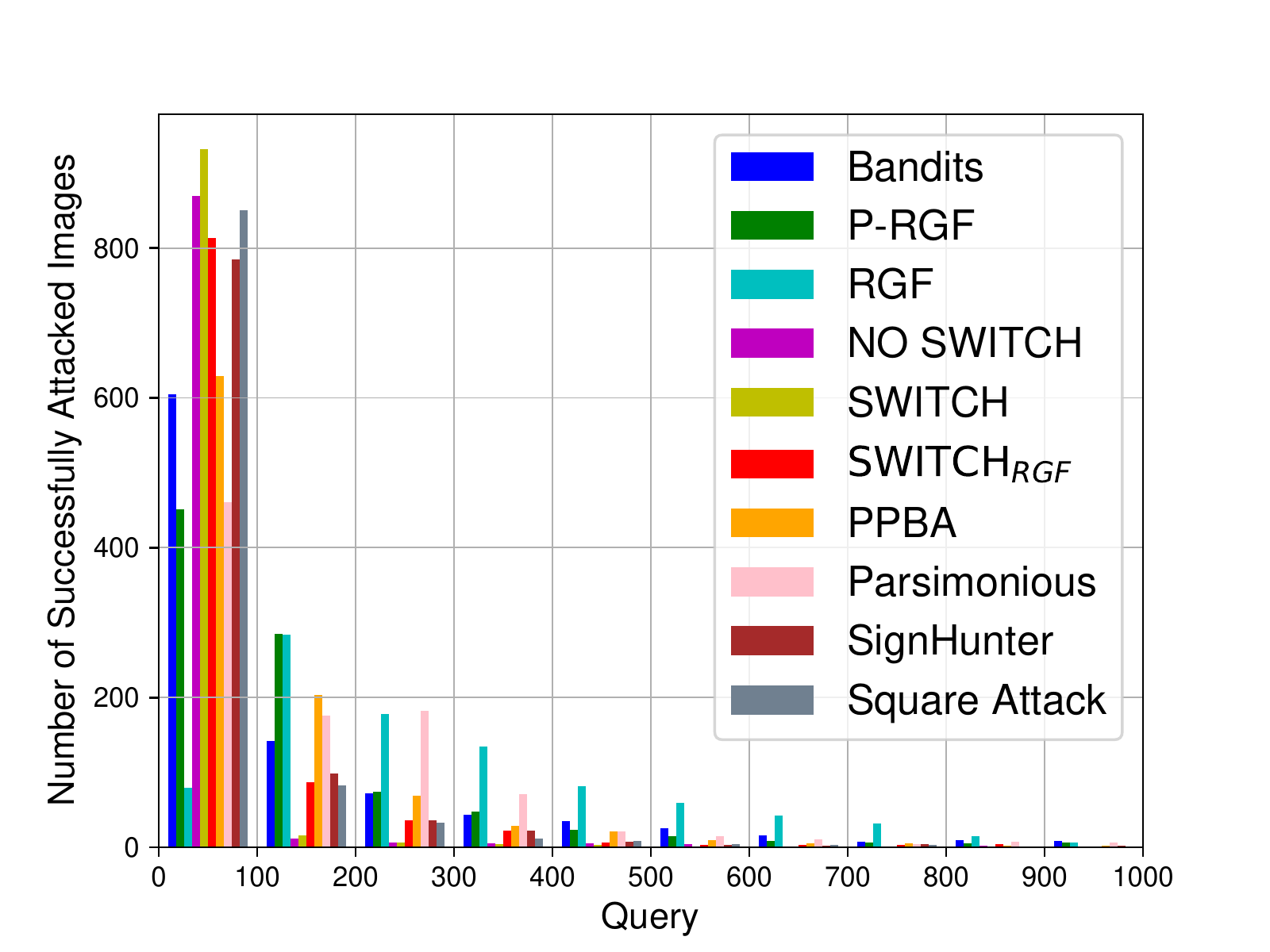}
			\subcaption{untargeted $\ell_\infty$ attack GDAS}
		\end{minipage}
		\begin{minipage}[b]{.23\textwidth}
			\includegraphics[width=\linewidth]{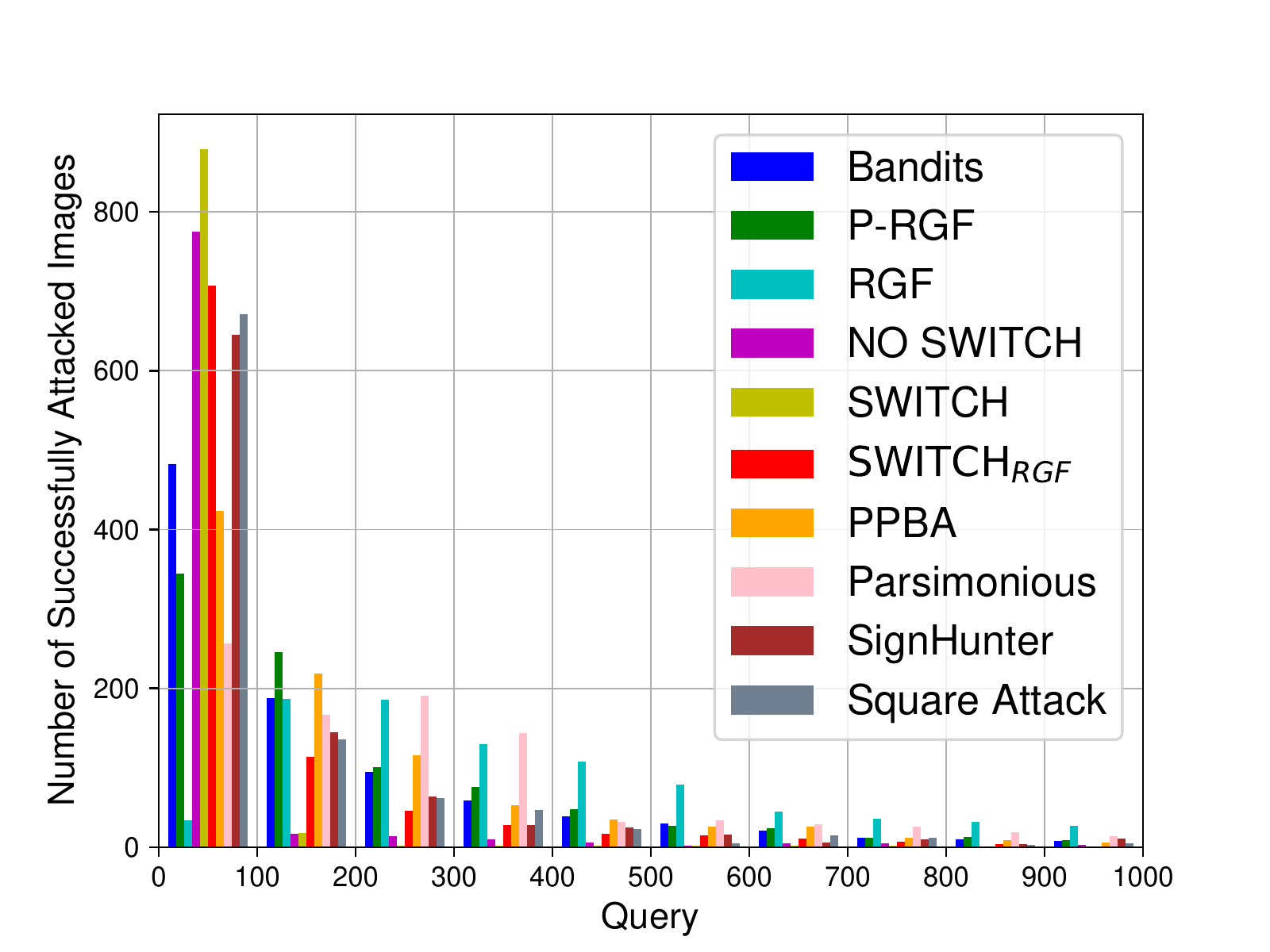}
			\subcaption{untargeted $\ell_\infty$ attack WRN-28}
		\end{minipage}
		\begin{minipage}[b]{.23\textwidth}
			\includegraphics[width=\linewidth]{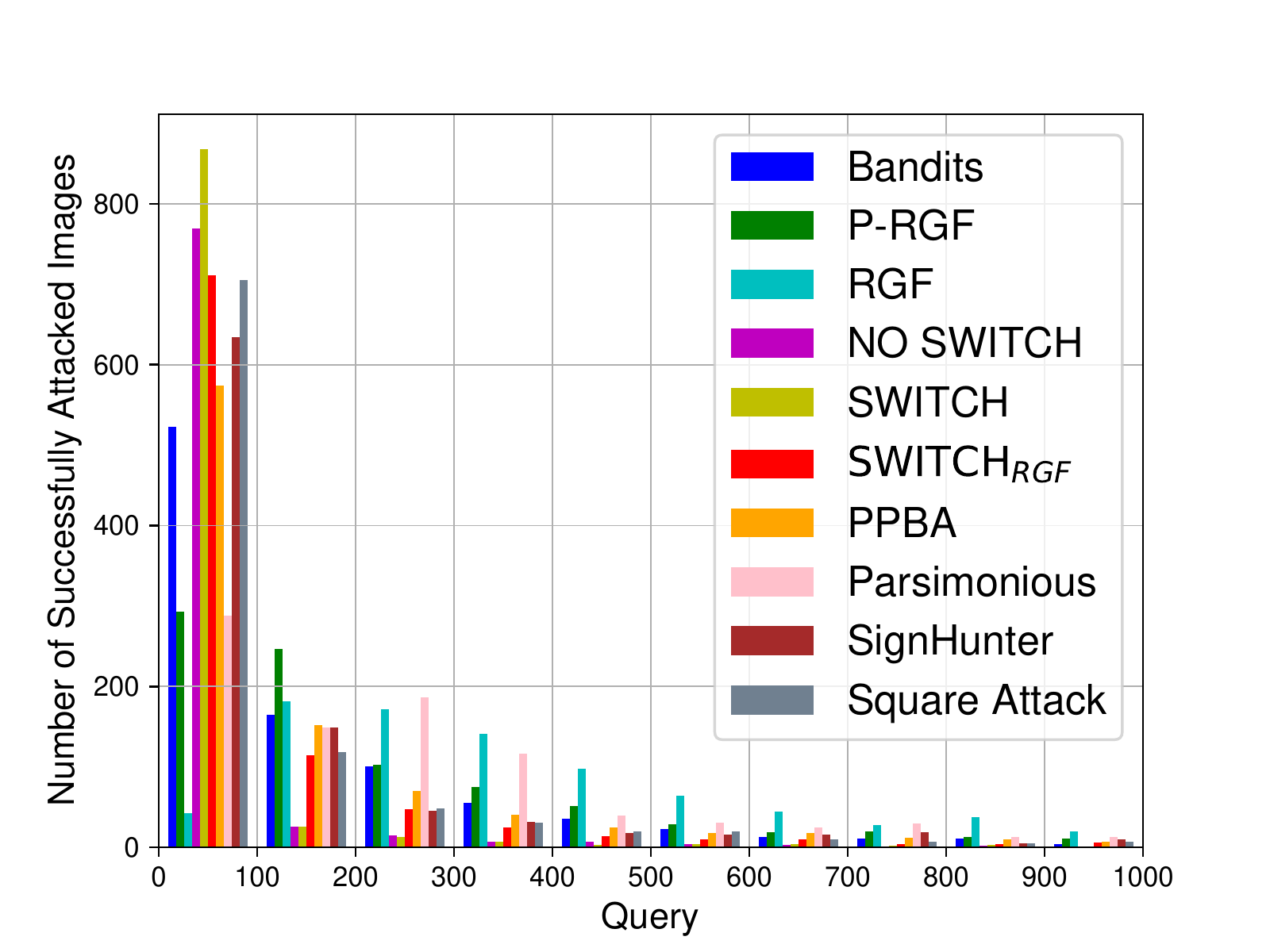}
			\subcaption{untargeted $\ell_\infty$ attack WRN-40}
		\end{minipage}
		\begin{minipage}[b]{.23\textwidth}
			\includegraphics[width=\linewidth]{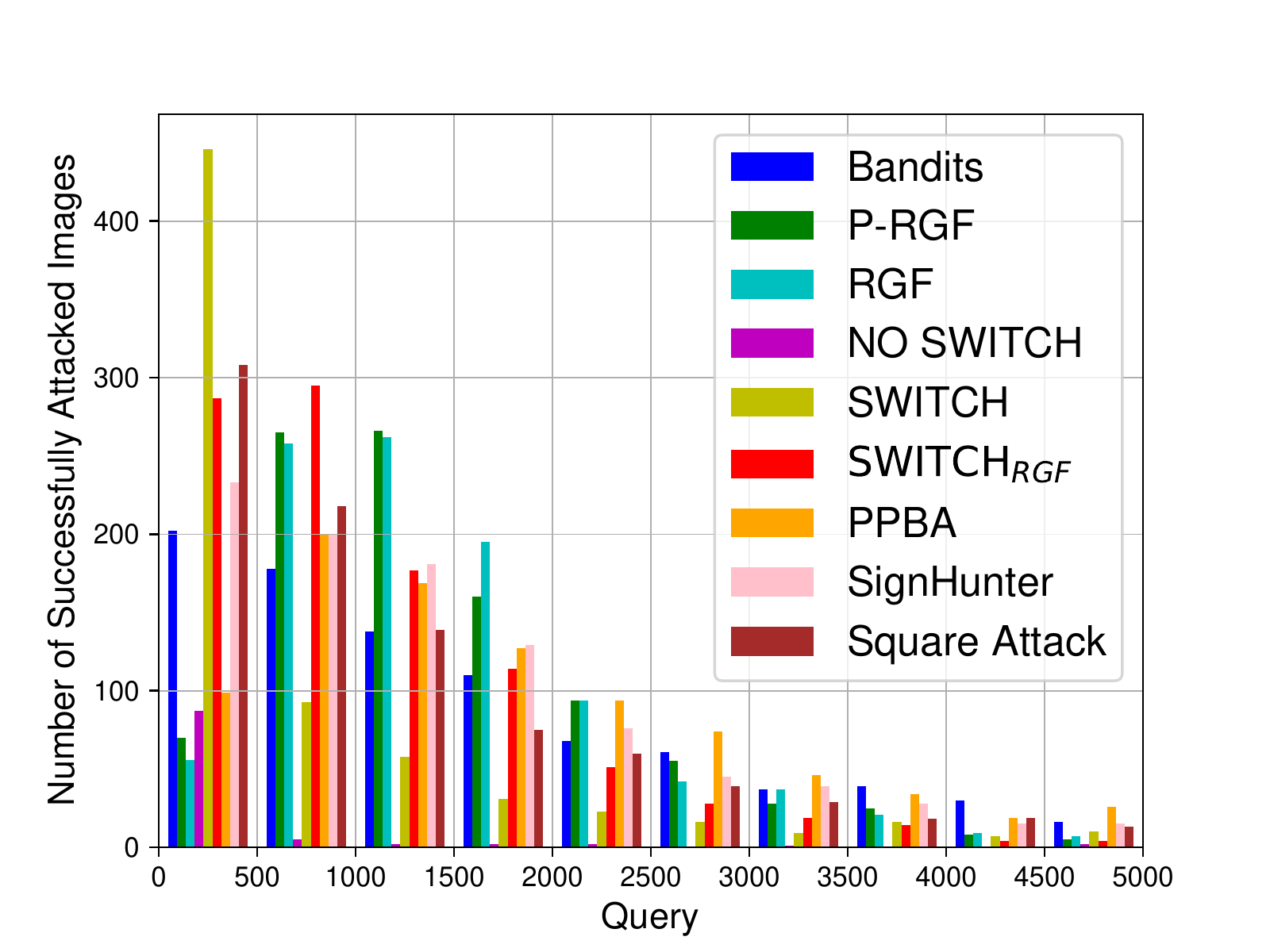}
			\subcaption{targeted $\ell_2$ attack PyramidNet-272}
		\end{minipage}
		\begin{minipage}[b]{.23\textwidth}
			\includegraphics[width=\linewidth]{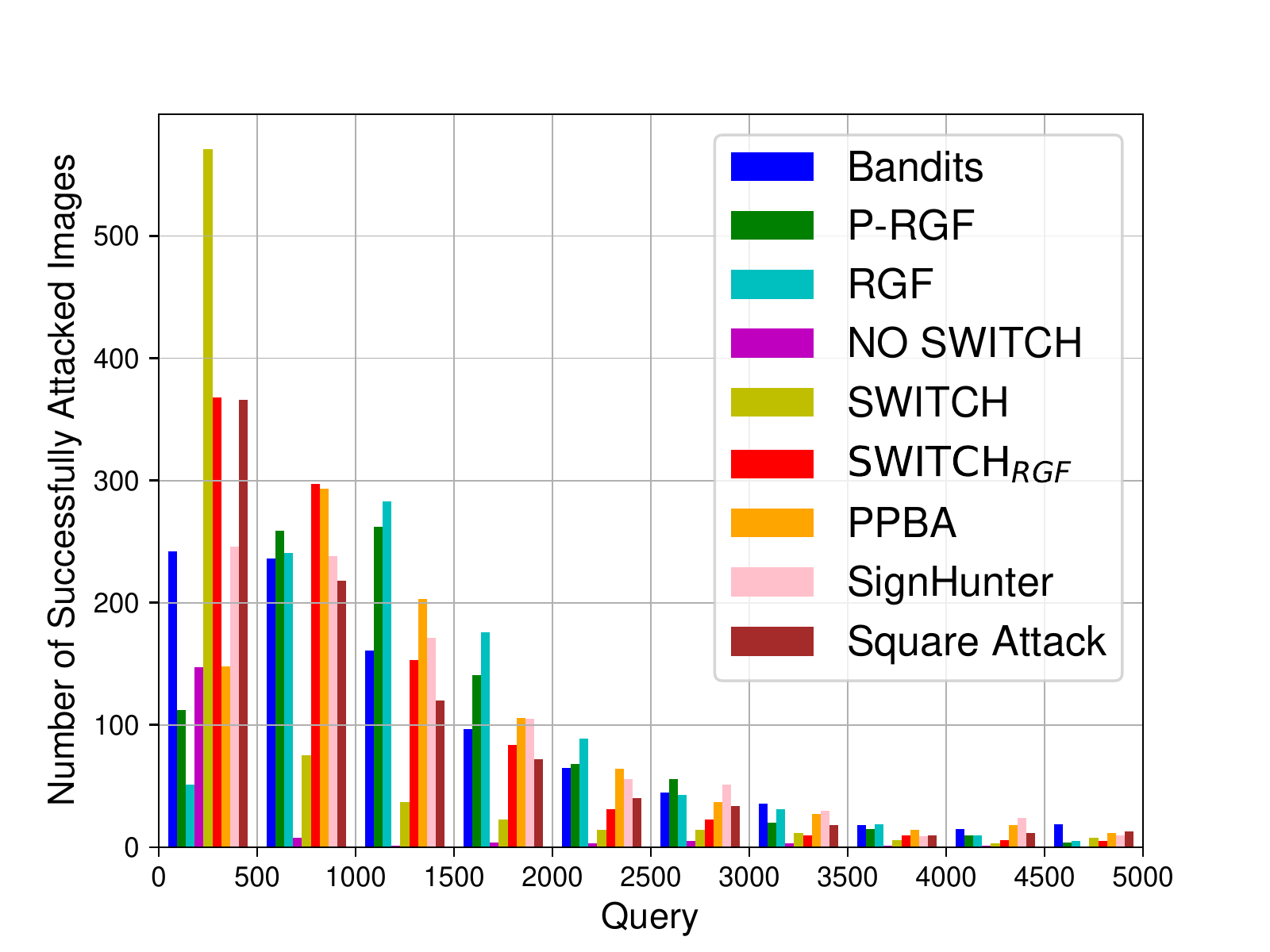}
			\subcaption{targeted $\ell_2$ attack GDAS}
		\end{minipage}
		\begin{minipage}[b]{.23\textwidth}
			\includegraphics[width=\linewidth]{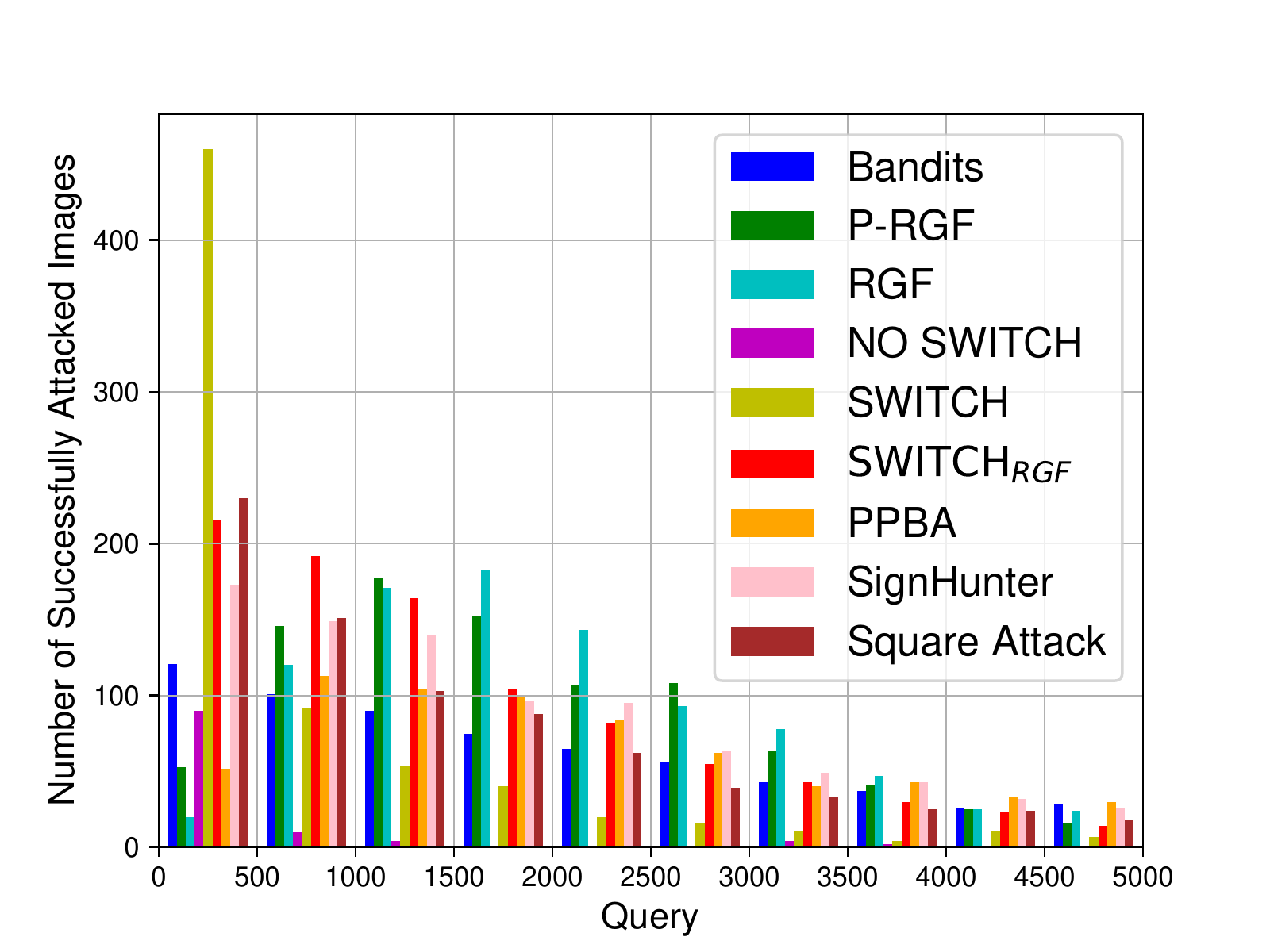}
			\subcaption{targeted $\ell_2$ attack WRN-28}
		\end{minipage}
		\begin{minipage}[b]{.23\textwidth}
			\includegraphics[width=\linewidth]{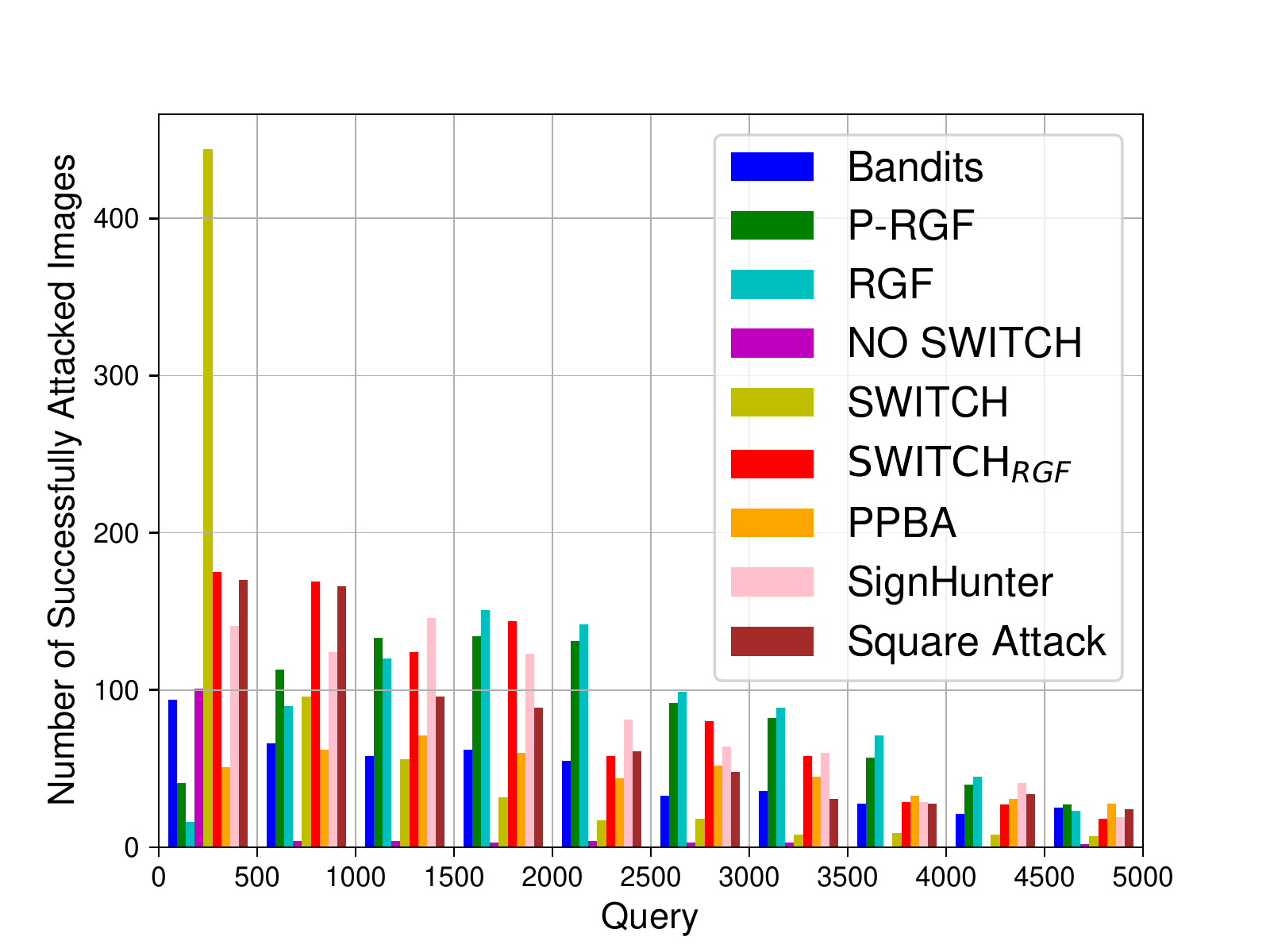}
			\subcaption{targeted $\ell_2$ attack WRN-40}
		\end{minipage}
		\caption{The histogram of query number in the CIFAR-100 dataset.}
		\label{fig:histogram_CIFAR-100}
	\end{figure*}

\begin{figure*}[htbp]
		\setlength{\abovecaptionskip}{0pt}
		\setlength{\belowcaptionskip}{0pt}
		\captionsetup[sub]{font={scriptsize}}
		\centering 
		\begin{minipage}[b]{.3\textwidth}
			\includegraphics[width=\linewidth]{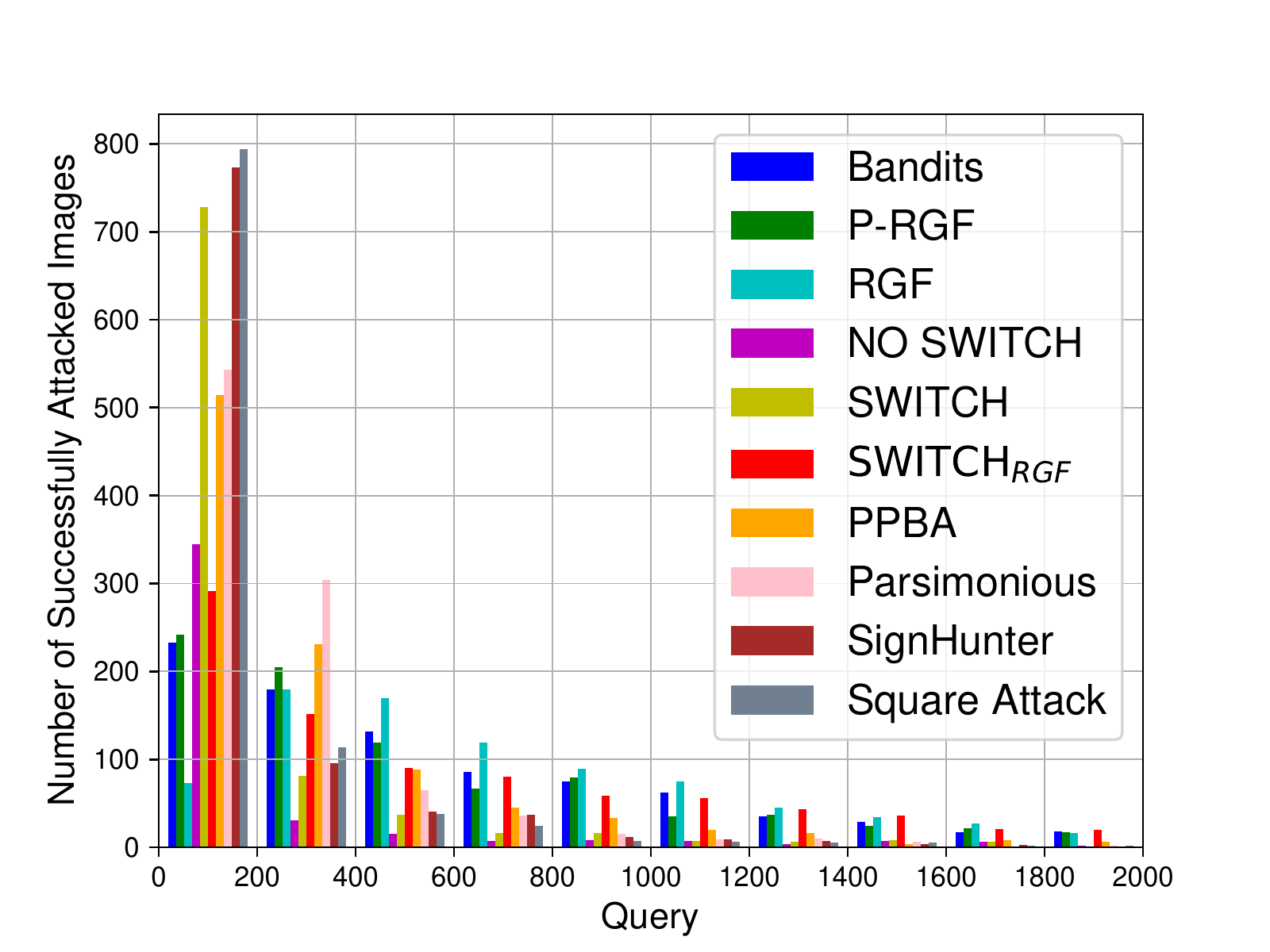}
			\subcaption{untargeted $\ell_\infty$ attack DenseNet-121}
		\end{minipage}
		\begin{minipage}[b]{.3\textwidth}
			\includegraphics[width=\linewidth]{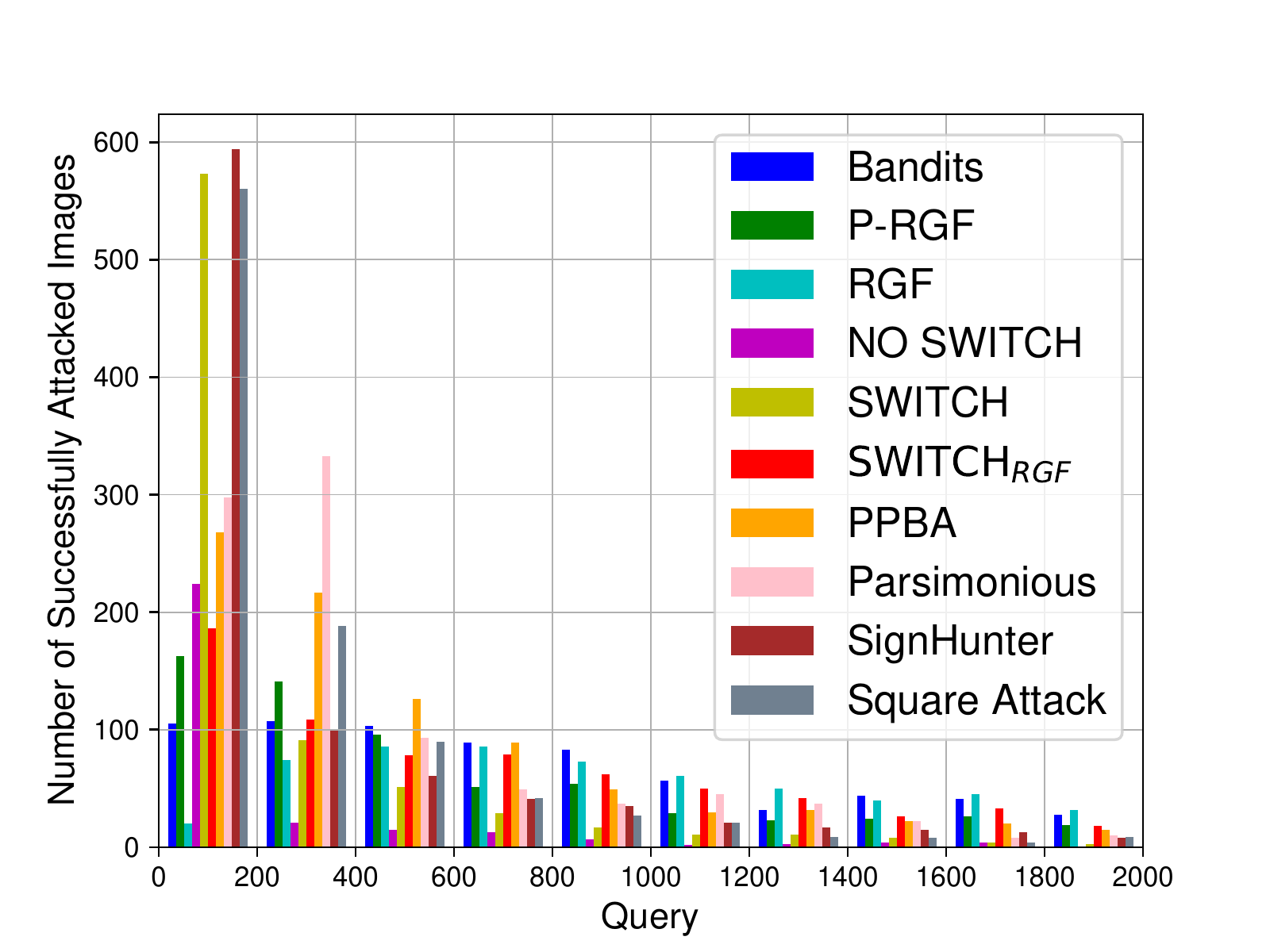}
			\subcaption{untargeted $\ell_\infty$ attack ResNext-101~(32$\times$4d)}
		\end{minipage}
		\begin{minipage}[b]{.3\textwidth}
			\includegraphics[width=\linewidth]{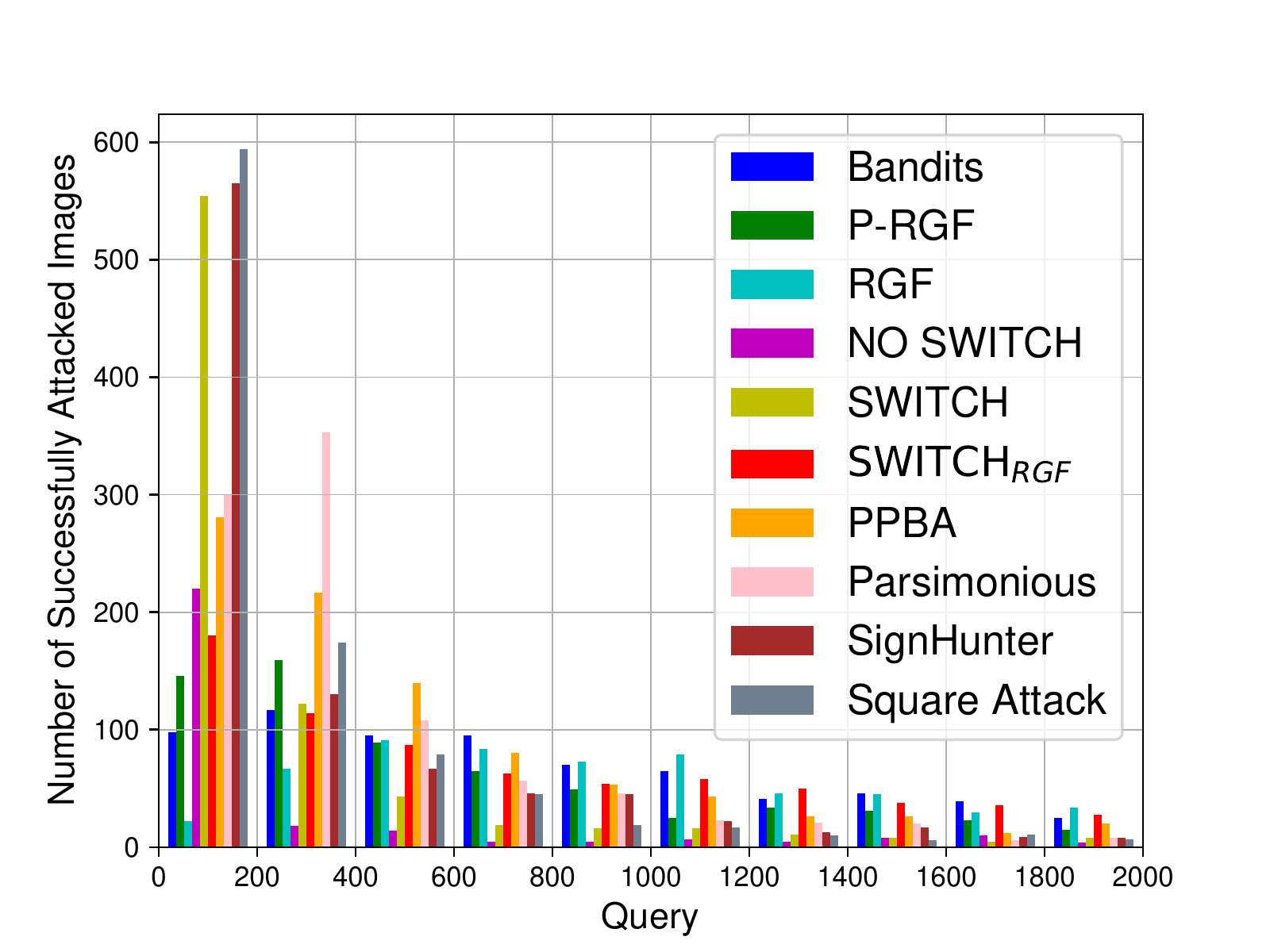}
			\subcaption{untargeted $\ell_\infty$ attack ResNext-101~(64$\times$4d)}
		\end{minipage}
		\begin{minipage}[b]{.3\textwidth}
			\includegraphics[width=\linewidth]{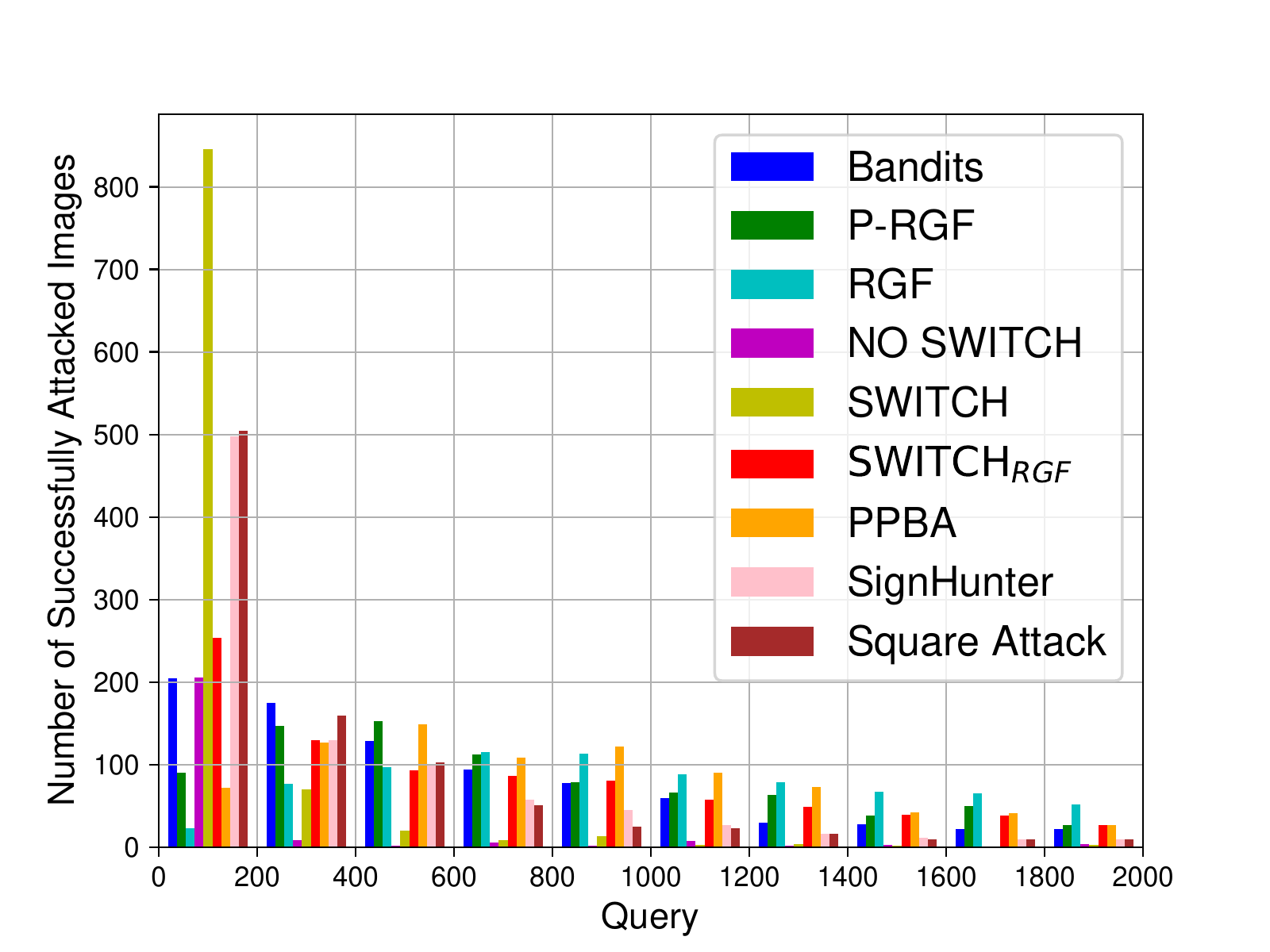}
			\subcaption{untargeted $\ell_2$ attack DenseNet-121}
		\end{minipage}
		\begin{minipage}[b]{.3\textwidth}
			\includegraphics[width=\linewidth]{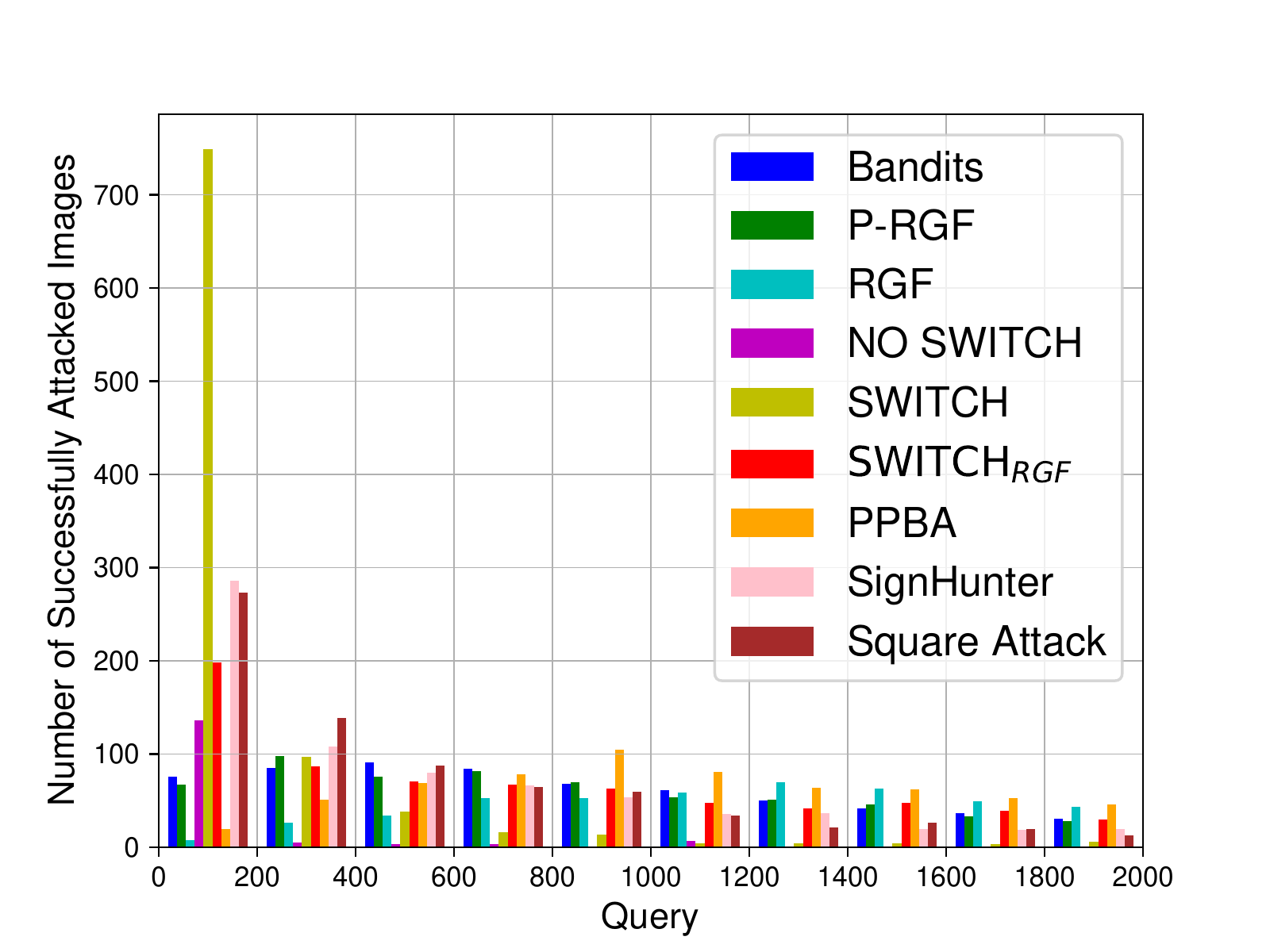}
			\subcaption{untargeted $\ell_2$ attack ResNext-101~(32$\times$4d)}
		\end{minipage}
		\begin{minipage}[b]{.3\textwidth}
			\includegraphics[width=\linewidth]{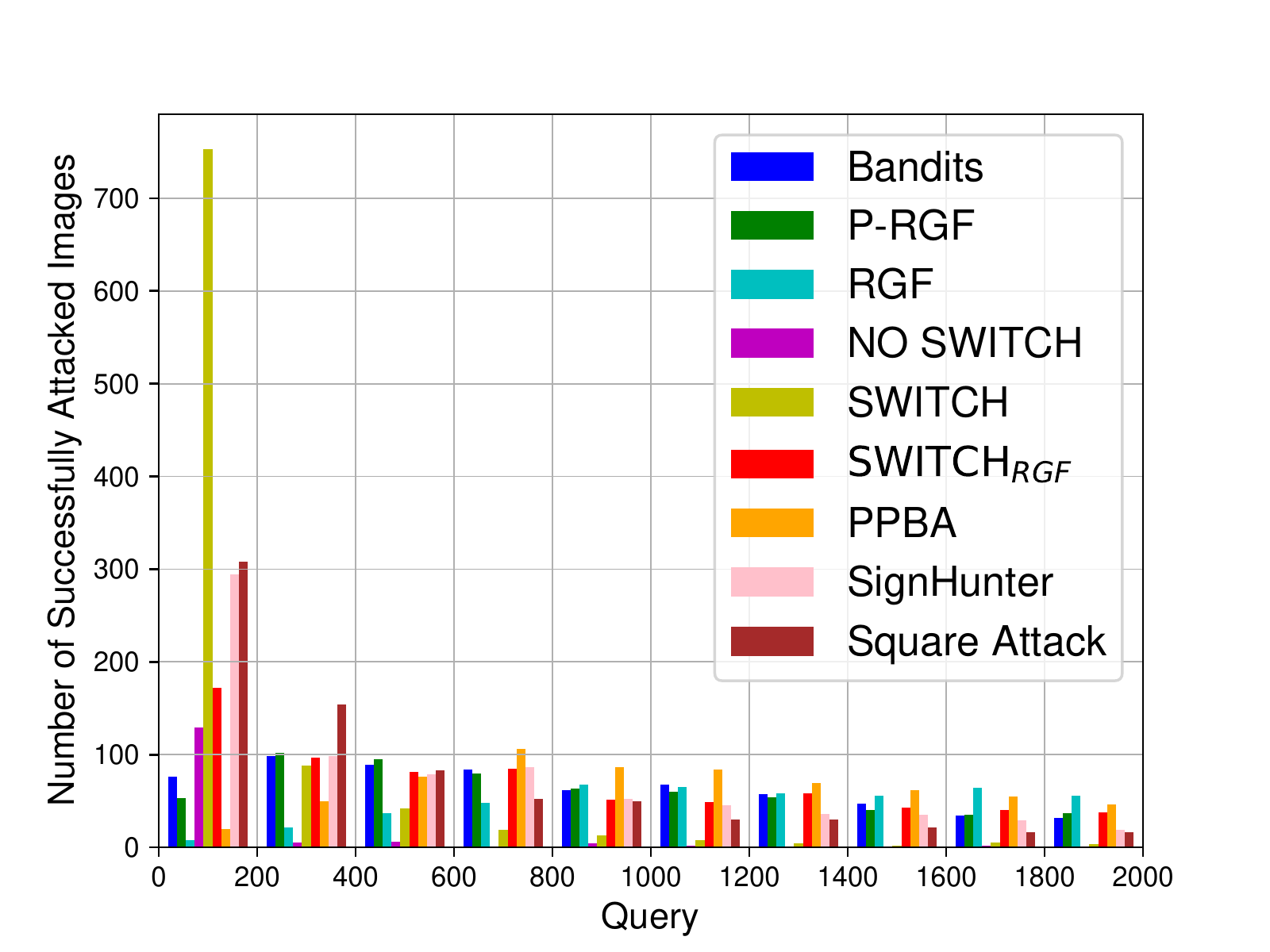}
			\subcaption{untargeted $\ell_2$ attack ResNext-101~(64$\times$4d)}
		\end{minipage}
		\begin{minipage}[b]{.3\textwidth}
			\includegraphics[width=\linewidth]{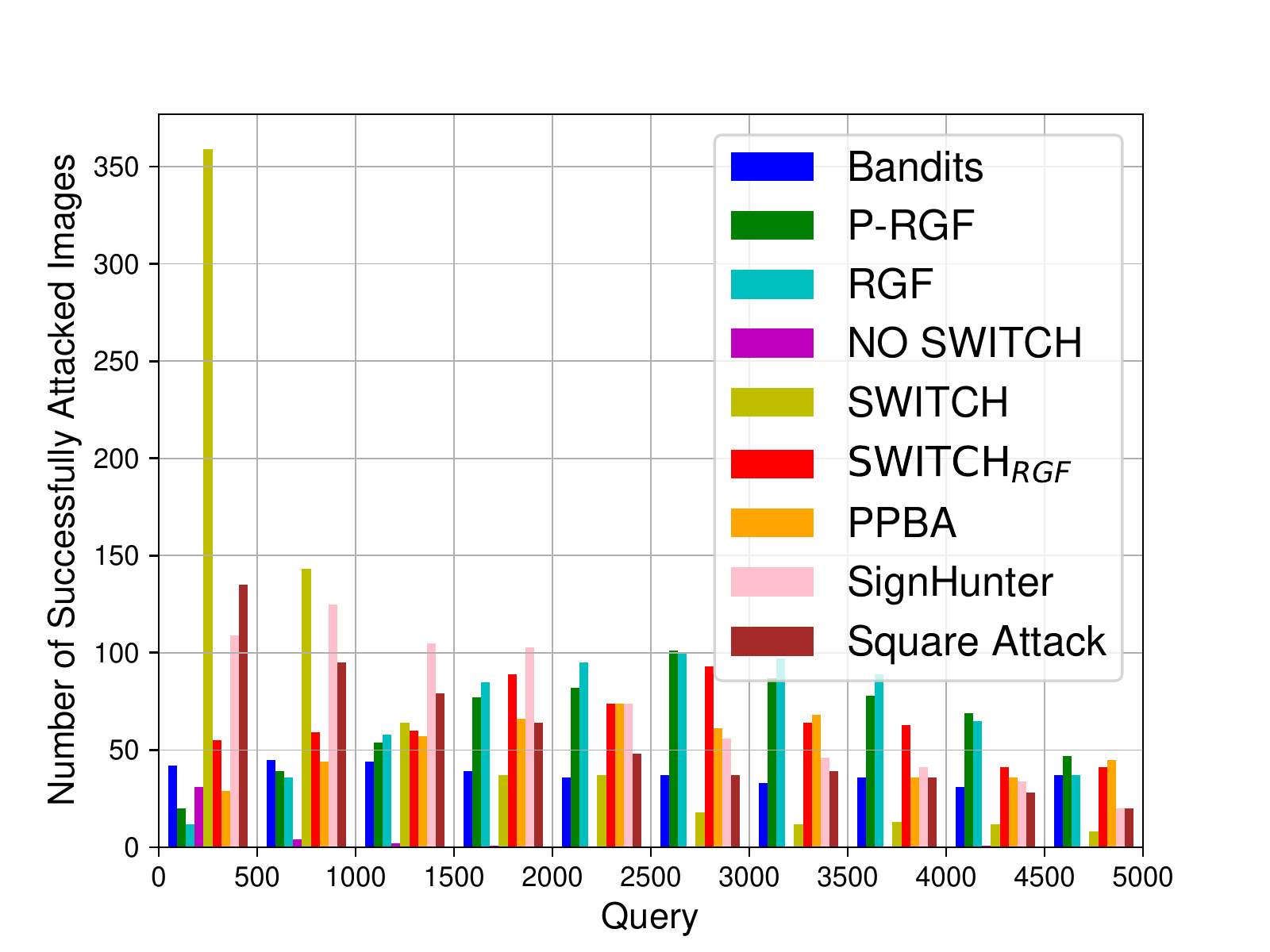}
			\subcaption{targeted $\ell_2$ attack DenseNet-121}
		\end{minipage}
		\begin{minipage}[b]{.3\textwidth}
			\includegraphics[width=\linewidth]{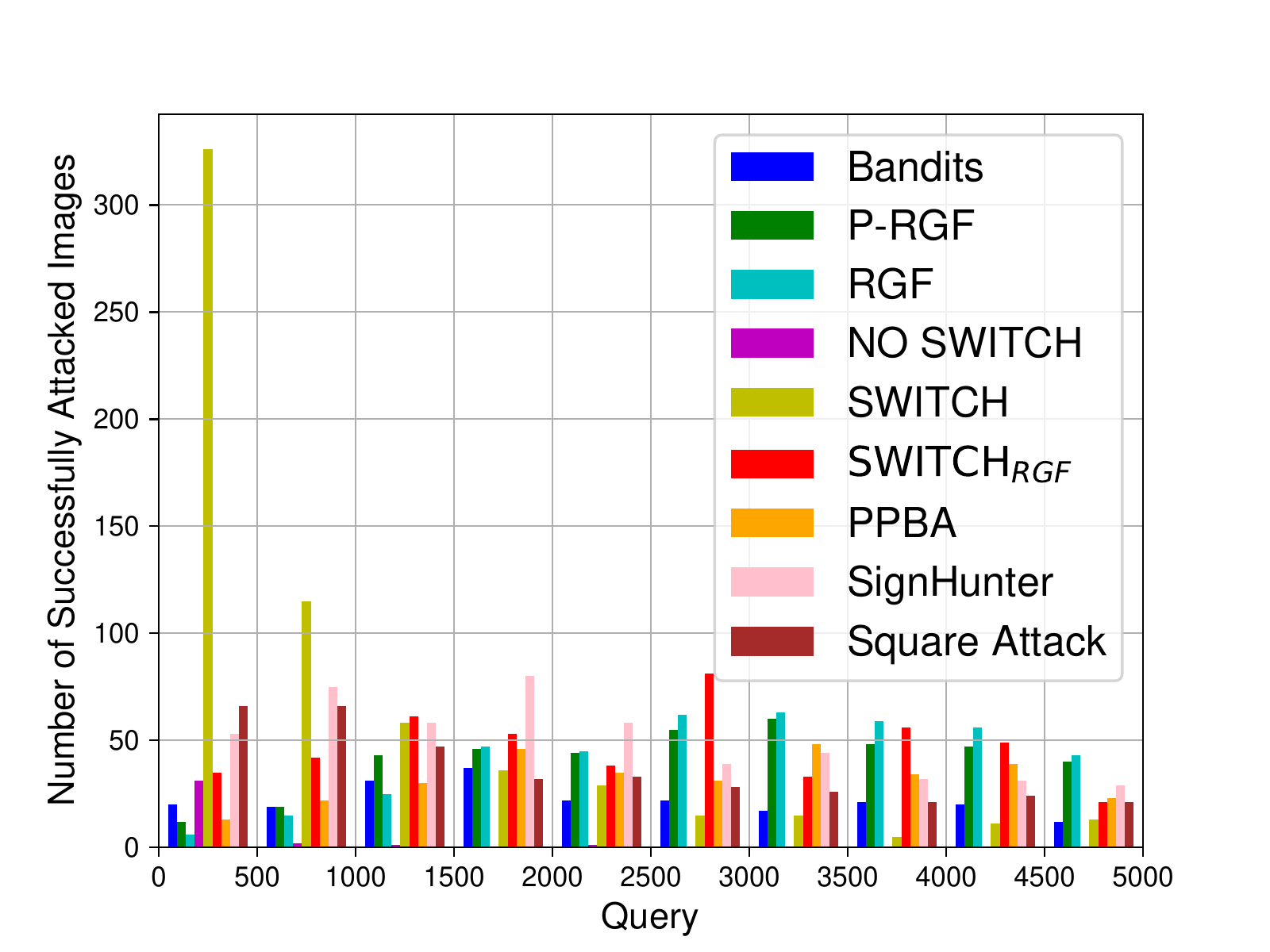}
			\subcaption{targeted $\ell_2$ attack ResNext-101~(32$\times$4d)}
		\end{minipage}
		\begin{minipage}[b]{.3\textwidth}
			\includegraphics[width=\linewidth]{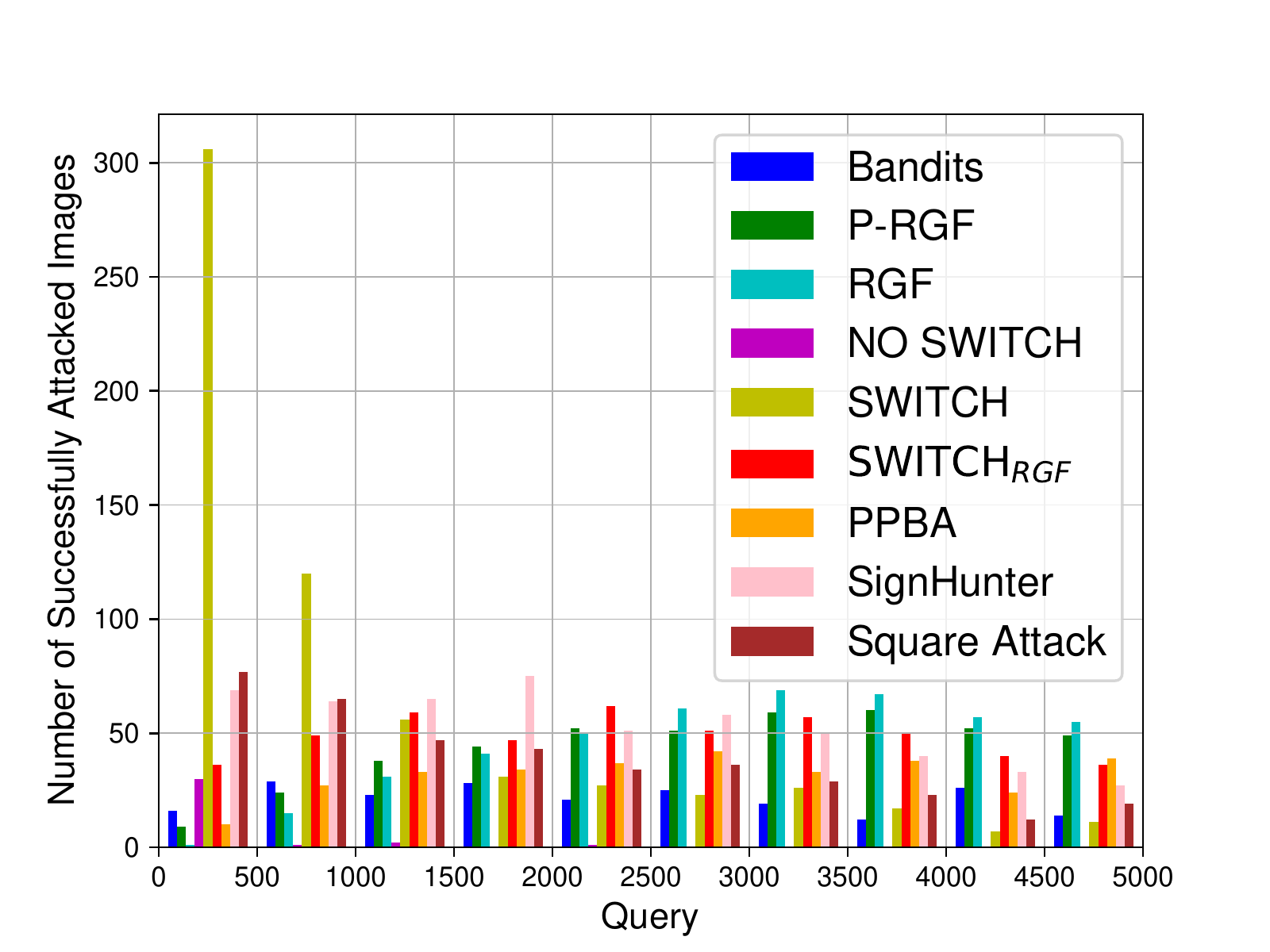}
			\subcaption{targeted $\ell_2$ attack ResNext-101~(64$\times$4d)}
		\end{minipage}
		\caption{The histogram of query number in the TinyImageNet dataset.}
		\label{fig:histogram_TinyImageNet}
	\end{figure*}

	\noindent\textbf{Histogram of Query Numbers.} To observe the distribution of query numbers in more detail, Fig. \ref{fig:histogram_CIFAR-10}, Fig. \ref{fig:histogram_CIFAR-100}, Fig. \ref{fig:histogram_TinyImageNet} show the histograms of query numbers in CIFAR-10, CIFAR-100, TinyImageNet. To draw the histogram, we separate the range of query number into 10 intervals to count the number of samples in each interval separately, and the intervals are separated by the vertical lines of figures. Each bar indicates one attack, and the height of each bar indicates the number of samples with the queries belong to this query interval.
	The results show that the highest red bars (SWITCH) are always located in the low query number's area, thereby confirming that most adversarial examples of SWITCH have the fewest queries.
	\begin{table}[ht]
		\small
		\tabcolsep=0.1cm
		\caption{Experimental results of untargeted attack under $\ell_2$ norm on TinyImageNet dataset. D$_{121}$: DenseNet-121, R$_{32}$: ResNeXt-101~(32$\times$4d), R$_{64}$: ResNeXt-101~(64$\times$4d). We include failed samples in the computation of the average and median query, and the query numbers of failed samples are set to \nn{10000}.}
		\label{tab:TinyImageNet_untargeted_l2_result_include_failed}
		\resizebox{1\linewidth}{!}{
			\begin{tabular}{c|ccc|ccc|ccc}
				\toprule
				\B{Attack} & \multicolumn{3}{c|}{\B{Attack Success Rate}} &  \multicolumn{3}{c|}{\B{Avg. Query}} &  \multicolumn{3}{c}{\B{Median Query}} \\
				&  D$_{121}$ & R$_{32}$ & R$_{64}$  & D$_{121}$ & R$_{32}$ & R$_{64}$ & D$_{121}$ & R$_{32}$ & R$_{64}$ \\ 
				\midrule
				RGF \cite{nesterov2017random} & 99.4\% & 94\% & 96.1\% & 1536 & 3085 & 2868 & 1173 & 2193 & 2094 \\
				P-RGF \cite{cheng2019improving} & 99.5\% & 94.1\% & 94.8\% & 1262 & 2551 & 2402 & 794 & 1402 & 1376 \\
				Bandits \cite{ilyas2018prior} & 98.7\% & 93\% & 94.6\% & 1148 & 2565 & 2333 & 585 & 1333 & 1271 \\
				PPBA \cite{li2020projection} & 99.7\% & 96.9\% & 97.5\% & 1178 & 2252 & 2042 & 853 & 1511 & 1440 \\
				SignHunter \cite{al2020sign} & 97.6\% & 93\% & 94.4\% & 786 & 1900 & 1728 & 204 & 673 & 658 \\
				Square Attack \cite{ACFH2020square} & 98.5\% & 92.2\% & 93.6\% & 707 & 1966 & 1730 & 196 & 600 & 485 \\
				NO SWITCH & 25.6\% & 17.8\% & 16.1\% & 7542 & 8334 & 8470 & 10000 & 10000 & 10000 \\
				SWITCH & 98.6\% & 97.1\% & 96.9\% & 311 & 588 & 599 & 56 & 77 & 79 \\
				SWITCH$_\text{RGF}$ & 99.4\% & 97.7\% & 97.4\% & 1064 & 1891 & 1772 & 663 & 1047 & 1056 \\
				\bottomrule
		\end{tabular}}

	\end{table}
	
	\begin{table}[ht]
		\small
		\tabcolsep=0.1cm
		\caption{Experimental results of untargeted attack under $\ell_\infty$ norm on TinyImageNet dataset. D$_{121}$: DenseNet-121, R$_{32}$: ResNeXt-101~(32$\times$4d), R$_{64}$: ResNeXt-101~(64$\times$4d). We include failed samples in the computation of the average and median query, and the query numbers of failed samples are set to \nn{10000}.}
		\label{tab:TinyImageNet_untargeted_linf_result_include_failed}
		\resizebox{1\linewidth}{!}{
			\begin{tabular}{c|ccc|ccc|ccc}
				\toprule
				\B{Attack} & \multicolumn{3}{c|}{\B{Attack Success Rate}} &  \multicolumn{3}{c|}{\B{Avg. Query}} &  \multicolumn{3}{c}{\B{Median Query}} \\
				&  D$_{121}$ & R$_{32}$ & R$_{64}$  & D$_{121}$ & R$_{32}$ & R$_{64}$ & D$_{121}$ & R$_{32}$ & R$_{64}$ \\ 
				\midrule
				RGF \cite{nesterov2017random} & 96.4\% & 85.3\% & 87.4\% & 1465 & 3251 & 3084 & 715 & 1639 & 1588 \\
				P-RGF \cite{cheng2019improving} & 94.5\% & 83.9\% & 85.9\% & 1384 & 2938 & 2768 & 468 & 968 & 954 \\
				Bandits \cite{ilyas2018prior} & 99.2\% & 94.1\% & 95.3\% & 1036 & 2224 & 2054 & 532 & 1040 & 1075 \\
				PPBA \cite{li2020projection} & 99.6\% & 97.9\% & 98.2\% & 441 & 1116 & 979 & 192 & 425 & 401 \\
				Parsimonious \cite{moonICML19} & 100\% & 99.3\% & 99.3\% & 273 & 708 & 644 & 190 & 301 & 287 \\
				SignHunter \cite{al2020sign} & 99.5\% & 98.6\% & 99\% & 250 & 713 & 621 & 59 & 125 & 156 \\
				Square Attack \cite{ACFH2020square} & 100\% & 99.4\% & 99.6\% & 160 & 454 & 417 & 74 & 160 & 143 \\
				NO SWITCH & 49.3\% & 33.4\% & 33.4\% & 5431 & 6918 & 6898 & 10000 & 10000 & 10000 \\
				SWITCH & 93.1\% & 85.7\% & 85.4\% & 948 & 1880 & 1891 & 73 & 132 & 148 \\
				SWITCH$_\text{RGF}$ & 98.3\% & 92.9\% & 94\% & 1165 & 2204 & 2011 & 534 & 951 & 1002 \\
				\bottomrule
		\end{tabular}}

	\end{table}
	
	\begin{table}[ht]
		\small
		\tabcolsep=0.1cm
		\caption{Experimental results of targeted attack under $\ell_2$ norm on TinyImageNet dataset. D$_{121}$: DenseNet-121, R$_{32}$: ResNeXt-101~(32$\times$4d), R$_{64}$: ResNeXt-101~(64$\times$4d). We include failed samples in the computation of the average and median query, and the query numbers of failed samples are set to \nn{10000}.}
		\label{tab:TinyImageNet_targeted_l2_result_include_failed}
		\resizebox{1\linewidth}{!}{
			\begin{tabular}{c|ccc|ccc|ccc}
				\toprule
				\B{Attack} & \multicolumn{3}{c|}{\B{Attack Success Rate}} &  \multicolumn{3}{c|}{\B{Avg. Query}} &  \multicolumn{3}{c}{\B{Median Query}} \\
				& D$_{121}$ & R$_{32}$ & R$_{64}$  & D$_{121}$ & R$_{32}$ & R$_{64}$ & D$_{121}$ & R$_{32}$ & R$_{64}$ \\ 
				\midrule
				RGF \cite{nesterov2017random} & 87.5\% & 71.4\% & 75.5\% & 4512 & 6309 & 6069 & 3570 & 6248 & 5661 \\
				P-RGF \cite{cheng2019improving} & 84\% & 67.7\% & 71.9\% & 4714 & 6323 & 6121 & 3782 & 6406 & 5838 \\
				Bandits \cite{ilyas2018prior} & 56.4\% & 36.8\% & 36.6\% & 6563 & 7876 & 7988 & 7533 & 10000 & 10000 \\
				PPBA \cite{li2020projection} & 70.1\% & 50.3\% & 51.9\% & 5608 & 7111 & 7140 & 4787 & 9794 & 9322 \\
				SignHunter \cite{al2020sign} & 85.3\% & 67.4\% & 72.2\% & 3731 & 5512 & 5243 & 2395 & 5005 & 4374 \\
				Square Attack \cite{ACFH2020square} & 70.1\% & 49.3\% & 53.9\% & 4809 & 6691 & 6438 & 3568 & 10000 & 8570 \\
				NO SWITCH & 4.3\% & 3.5\% & 3.6\% & 9608 & 9657 & 9662 & 10000 & 10000 & 10000 \\
				SWITCH & 74.7\% & 68.4\% & 68.5\% & 3457 & 4154 & 4200 & 983 & 1548 & 1781 \\
				SWITCH$_\text{RGF}$ & 83\% & 71.1\% & 75.8\% & 4593 & 5813 & 5566 & 3549 & 5456 & 5164 \\
				\bottomrule
		\end{tabular}}

	\end{table}

	\begin{table*}[t]
		\small	
		\tabcolsep=0.1cm
		
		\caption{Experimental results of untargeted attack on CIFAR-10 and CIFAR-100 datasets. We include failed samples in the computation of the average and median query, and the query numbers of failed samples are set to \nn{10000}.}
		\label{tab:CIFAR_untargetd_result_include_failed}
		\resizebox{1\linewidth}{!}{
			\begin{tabular}{c|c|c|cccc|cccc|cccc}
				\toprule
				\B Dataset & \B{Norm} & \B{Attack} & \multicolumn{4}{c|}{\B{Attack Success Rate}} &  \multicolumn{4}{c|}{\B{Avg. Query}} &  \multicolumn{4}{c}{\B{Median Query}} \\
				& & & PyramidNet-272 & GDAS & WRN-28 & WRN-40 & PyramidNet-272 & GDAS & WRN-28 & WRN-40 & PyramidNet-272 & GDAS & WRN-28 & WRN-40 \\ 
				\midrule
				\multirow{21}{*}{CIFAR-10} & \multirow{10}{*}{$\ell_2$} & RGF \cite{nesterov2017random} & 100\% & 98.9\% & 99.4\% & 100\% & 1173 & 989 & 1672 & 1352 & 1020 & 714 & 1173 & 1071 \\
				& & P-RGF \cite{cheng2019improving} & 100\% & 99.8\% & 99.1\% & 99.6\% & 853 & 492 & 1182 & 868 & 666 & 240 & 470 & 432 \\
				& & Bandits \cite{ilyas2018prior} & 100\% & 100\% & 99.4\% & 99.7\% & 692 & 332 & 862 & 693 & 474 & 210 & 336 & 313 \\
				& & PPBA \cite{li2020projection} & 100\% & 99.9\% & 97.1\% & 99.1\% & 1056 & 643 & 1591 & 1199 & 833 & 486 & 801 & 739 \\
				& & SignHunter \cite{al2020sign} & 97\% & 85.8\% & 97.7\% & 99\% & 1070 & 1861 & 1412 & 1128 & 510 & 358 & 614 & 579 \\
				& & Square Attack \cite{ACFH2020square} & 99.8\% & 100\% & 96.6\% & 98.4\% & 786 & 363 & 1178 & 893 & 436 & 160 & 331 & 313 \\
				& & NO SWITCH & 39.8\% & 77.1\% & 80.5\% & 82.3\% & 6077 & 2337 & 1982 & 1811 & 10000 & 16 & 12 & 11 \\
				& & SWITCH & 90\% & 95.5\% & 97.3\% & 97.1\% & 1442 & 566 & 407 & 371 & 83 & 19 & 15 & 13 \\
				& & SWITCH$_\text{RGF}$& 100\% & 100\% & 99.4\% & 100\% & 555 & 326 & 695 & 516 & 369 & 107 & 61 & 28 \\
				\cmidrule(rl){2-15} & \multirow{11}{*}{$\ell_\infty$} & RGF \cite{nesterov2017random} & 98.8\% & 93.8\% & 98.7\% & 99\% & 1051 & 1225 & 1309 & 1042 & 669 & 510 & 663 & 612 \\
				& & P-RGF \cite{cheng2019improving} & 97.3\% & 97.9\% & 97.7\% & 98\% & 992 & 540 & 918 & 754 & 420 & 132 & 239 & 232 \\
				& & Bandits \cite{ilyas2018prior} & 99.6\% & 100\% & 99.4\% & 99.9\% & 1051 & 391 & 667 & 551 & 564 & 166 & 226 & 228 \\
				& & PPBA \cite{li2020projection} \cite{li2020projection} & 96.6\% & 99.5\% & 96.8\% & 97.6\% & 1171 & 382 & 997 & 801 & 361 & 123 & 199 & 203 \\
				& & Parsimonious \cite{moonICML19} & 100\% & 100\% & 100\% & 100\% & 701 & 345 & 891 & 738 & 523 & 230 & 424 & 375 \\
				& & SignHunter \cite{al2020sign} & 99.4\% & 91.8\% & 100\% & 100\% & 436 & 1053 & 506 & 415 & 191 & 122 & 205 & 188 \\
				& & Square Attack \cite{ACFH2020square} & 100\% & 100\% & 99.9\% & 100\% & 332 & 126 & 412 & 342 & 182 & 54 & 145 & 140 \\
				& & NO SWITCH & 48.9\% & 84.3\% & 89.4\% & 91.8\% & 5311 & 1696 & 1119 & 933 & 10000 & 5 & 3 & 3 \\
				& & SWITCH & 77.2\% & 96\% & 97.2\% & 98\% & 2685 & 528 & 397 & 267 & 89 & 5 & 3 & 3 \\
				& & SWITCH$_\text{RGF}$ & 92.8\% & 98.2\% & 97.4\% & 98.2\% & 1207 & 321 & 509 & 394 & 209 & 5 & 3 & 3 \\
				\midrule
				\multirow{21}{*}{CIFAR-100} & \multirow{10}{*}{$\ell_2$} & RGF \cite{nesterov2017random} & 100\% & 99.8\% & 99.6\% & 99.5\% & 565 & 573 & 913 & 1031 & 459 & 408 & 663 & 612 \\
				& & P-RGF \cite{cheng2019improving} & 100\% & 99.8\% & 99.6\% & 99.3\% & 407 & 301 & 603 & 830 & 280 & 180 & 302 & 356 \\
				& & Bandits \cite{ilyas2018prior} & 100\% & 100\% & 99.9\% & 99.6\% & 188 & 170 & 322 & 365 & 96 & 90 & 148 & 136 \\
				& & PPBA \cite{li2020projection} & 100\% & 100\% & 99.8\% & 99.4\% & 439 & 362 & 642 & 771 & 323 & 280 & 412 & 423 \\
				& & SignHunter \cite{al2020sign} & 97.5\% & 94.1\% & 99.1\% & 99.1\% & 539 & 797 & 650 & 672 & 138 & 95 & 187 & 194 \\
				& & Square Attack \cite{ACFH2020square} & 100\% & 100\% & 99.5\% & 99.2\% & 217 & 139 & 375 & 450 & 98 & 55 & 123 & 122 \\
				& & NO SWITCH & 68.5\% & 84.2\% & 73.6\% & 74.7\% & 3225 & 1657 & 2684 & 2577 & 20 & 11 & 16 & 14 \\
				& & SWITCH & 95.7\% & 98.5\% & 96.2\% & 96.1\% & 548 & 229 & 519 & 514 & 20 & 11 & 16 & 17 \\
				& & SWITCH$_\text{RGF}$ & 100\% & 100\% & 99.9\% & 99.7\% & 241 & 160 & 316 & 424 & 107 & 15 & 66 & 87 \\
				\cmidrule(rl){2-15} & \multirow{11}{*}{$\ell_\infty$} & RGF \cite{nesterov2017random} & 99.8\% & 98.8\% & 99\% & 99\% & 409 & 542 & 647 & 721 & 256 & 255 & 357 & 357 \\
				& & P-RGF \cite{cheng2019improving} & 99.3\% & 98.7\% & 97.6\% & 97.8\% & 377 & 369 & 585 & 690 & 153 & 116 & 156 & 182 \\
				& & Bandits \cite{ilyas2018prior} & 100\% & 100\% & 99.8\% & 99.8\% & 266 & 209 & 282 & 280 & 68 & 57 & 108 & 92 \\
				& & PPBA \cite{li2020projection} \cite{li2020projection} & 99.6\% & 99.9\% & 98.7\% & 98.6\% & 290 & 178 & 476 & 417 & 106 & 73 & 129 & 80 \\
				& & Parsimonious \cite{moonICML19} & 100\% & 100\% & 100\% & 99.9\% & 287 & 185 & 383 & 432 & 204 & 103 & 215 & 213 \\
				& & SignHunter \cite{al2020sign} & 99\% & 97.3\% & 99.8\% & 100\% & 225 & 385 & 230 & 255 & 53 & 39 & 59 & 60 \\
				& & Square Attack \cite{ACFH2020square} & 100\% & 100\% & 100\% & 99.8\% & 76 & 57 & 150 & 182 & 17 & 19 & 46 & 37 \\
				& & NO SWITCH & 81.8\% & 92.1\% & 86.4\% & 87.2\% & 2020 & 866 & 1470 & 1442 & 5 & 3 & 4 & 4 \\
				& & SWITCH & 93.4\% & 97.5\% & 93.5\% & 94.3\% & 793 & 311 & 747 & 652 & 5 & 3 & 4 & 4 \\
				& & SWITCH$_\text{RGF}$ & 99.3\% & 98.5\% & 98.4\% & 97.7\% & 206 & 212 & 324 & 374 & 6 & 3 & 5 & 4 \\
				\bottomrule
		\end{tabular}}

	\end{table*}
	
	\begin{table*}[t]
		\small
		\tabcolsep=0.1cm
		\caption{Experimental results of targeted attack under $\ell_2$ norm constraint on CIFAR-10 and CIFAR-100 datasets. We include failed samples in the computation of the average and median query, and the query numbers of failed samples are set to \nn{10000}.}
		\label{tab:CIFAR_targetd_l2_result_include_failed}
		\resizebox{1\linewidth}{!}{
			\begin{tabular}{c|c|cccc|cccc|cccc}
				\toprule
				\B Dataset & \B Attack & \multicolumn{4}{c|}{\B{Attack Success Rate}} &  \multicolumn{4}{c|}{\B{Avg. Query}} &  \multicolumn{4}{c}{\B{Median Query}} \\
				& &  PyramidNet-272 & GDAS & WRN-28 & WRN-40 & PyramidNet-272 & GDAS & WRN-28 & WRN-40 & PyramidNet-272 & GDAS & WRN-28 & WRN-40 \\ 
				\midrule
				\multirow{9}{*}{CIFAR-10} & RGF \cite{nesterov2017random} & 100\% & 100\% & 97.4\% & 99.8\% & 1795 & 1249 & 2643 & 2191 & 1632 & 1071 & 2091 & 1938 \\
				& P-RGF \cite{cheng2019improving} & 99.9\% & 100\% & 96.9\% & 99.7\% & 1706 & 1149 & 2369 & 1762 & 1548 & 966 & 1668 & 1444 \\
				& Bandits \cite{ilyas2018prior} & 96.3\% & 100\% & 81.5\% & 84.2\% & 2264 & 905 & 3554 & 3335 & 1478 & 650 & 1824 & 1673 \\
				& PPBA \cite{li2020projection} & 96\% & 99.8\% & 82.1\% & 88.1\% & 2380 & 1198 & 3943 & 3493 & 1535 & 887 & 2474 & 2190 \\
				& SignHunter \cite{al2020sign} & 98.9\% & 99.9\% & 92.7\% & 97.3\% & 1772 & 1056 & 2543 & 2083 & 1427 & 793 & 1590 & 1536 \\
				& Square Attack \cite{ACFH2020square} & 95.1\% & 99.5\% & 88.8\% & 93.5\% & 2107 & 931 & 2877 & 2239 & 1162 & 449 & 1355 & 1140 \\
				& NO SWITCH & 14.8\% & 43.1\% & 53.2\% & 56.1\% & 8545 & 5735 & 4722 & 4455 & 10000 & 10000 & 144 & 88 \\
				& SWITCH & 71.4\% & 89.2\% & 89.6\% & 90.7\% & 3752 & 1550 & 1562 & 1432 & 1042 & 121 & 116 & 97 \\
				& SWITCH$_\text{RGF}$ & 100\% & 100\% & 97.6\% & 99.8\% & 1146 & 746 & 1769 & 1343 & 894 & 482 & 900 & 842 \\
				\midrule
				\multirow{9}{*}{CIFAR-100} & RGF \cite{nesterov2017random} & 99.9\% & 99.6\% & 98.2\% & 96.2\% & 1550 & 1573 & 2513 & 3053 & 1326 & 1275 & 2040 & 2397 \\
				& P-RGF \cite{cheng2019improving} & 99.9\% & 99.6\% & 97.7\% & 96.1\% & 1537 & 1472 & 2474 & 2952 & 1288 & 1185 & 1874 & 2278 \\
				& Bandits \cite{ilyas2018prior} & 97.5\% & 98.4\% & 82.3\% & 62.3\% & 2263 & 1751 & 4245 & 5761 & 1469 & 1106 & 2989 & 5658 \\
				& PPBA \cite{li2020projection} & 95.4\% & 96.4\% & 82.3\% & 63.8\% & 2447 & 1911 & 4249 & 5840 & 1643 & 1189 & 2914 & 5606 \\
				& SignHunter \cite{al2020sign} & 98.4\% & 96.4\% & 91.6\% & 89.4\% & 1632 & 1734 & 2613 & 2920 & 1190 & 1107 & 1684 & 1866 \\
				& Square Attack \cite{ACFH2020square} & 96.6\% & 94.2\% & 87.4\% & 85.1\% & 1763 & 1788 & 3088 & 3397 & 925 & 777 & 1649 & 1898 \\
				& NO SWITCH & 9.6\% & 14.6\% & 11.1\% & 11.6\% & 9114 & 8628 & 8977 & 8909 & 10000 & 10000 & 10000 & 10000 \\
				& SWITCH & 73.6\% & 78\% & 74.2\% & 73.5\% & 3375 & 2782 & 3284 & 3444 & 751 & 347 & 662 & 790 \\
				& SWITCH$_\text{RGF}$ & 99.9\% & 99.8\% & 97.8\% & 96.4\% & 1110 & 1011 & 1926 & 2367 & 839 & 699 & 1288 & 1587 \\
				\bottomrule
		\end{tabular}}
		
	\end{table*}

\end{document}